%% file: emnlp2020.tex
\documentclass[11pt,a4paper]{article}
\usepackage[hyperref]{emnlp2020}
\usepackage{times}
\usepackage{latexsym}

\usepackage{multicol}
\usepackage{graphicx}
\graphicspath{{./Figures/}{./Figures/Pad/}{./Figures/Wiki_Attack/}}
\usepackage{amsmath}
\usepackage{changepage}

\usepackage{subcaption}
\usepackage{float}
\usepackage{cleveref}
\usepackage{booktabs}
\usepackage{makecell}
\usepackage{algorithm,algorithmic}
\usepackage{todonotes}

\usepackage{float}

\makeatletter
\newcommand\avsuminner[2]{%
  {\sbox0{$\m@th#1\sum$}%
  \vphantom{\usebox0}%
  \ooalign{%
     \hidewidth
     \smash{\vrule height\dimexpr\ht0+1pt\relax depth\dimexpr\dp0+1pt\relax}%
     \hidewidth\cr
     $\m@th#1\sum$\cr
  }%
  }%
}
\makeatother

\crefformat{section}{\S#2#1#3} %
\crefformat{subsection}{\S#2#1#3}
\crefformat{subsubsection}{\S#2#1#3}

\usepackage{microtype}

\aclfinalcopy %

\def\intitle{Why and when should you pool?\\Analyzing Pooling in Recurrent Architectures}
\title{\intitle}

\author{Pratyush Maini$^\dagger$, Keshav Kolluru$^\dagger$, Danish Pruthi$^\ddagger$, Mausam$^\dagger$ \\
$^\dagger$Indian Institute of Technology, Delhi, India \\
$^\ddagger$Carnegie Mellon University, Pittsburgh, USA \\
\texttt{\{pratyush.maini}, \texttt{keshav.kolluru\}@gmail.com}, \\ 
\texttt{ddanish@cs.cmu.edu}, \texttt{mausam@cse.iitd.ac.in}
}

\date{}

\newcommand\maxout{{\textsc{MaxPool}}}
\newcommand\attmax{{\textsc{MaxAtt}}}

\newcommand\att{{\textsc{Att}}}
\newcommand\meanout{{\textsc{MeanPool}}}
\newcommand\last{{BiLSTM}}
\newcommand\lastf{{BiLSTM}}

\begin{document}
\maketitle
\begin{abstract}
\input{sections/abstract.tex}

\end{abstract}

\section{Introduction}
\label{sec:intro}
\input{sections/intro.tex}

\section{Related Work}
\label{sec:related}
\input{sections/related.tex}

\section{Methods}
\label{sec:pooling}
\input{sections/methods.tex}

\section{Datasets \& Experimental Setup}
\label{sec:datasets}

\input{sections/datasets.tex}

\section{Gradient Propagation}
\label{sec:gradients}
\input{sections/gradients.tex}

\section{Positional Biases}
\label{sec:positional-biases}
\input{sections/positional-biases.tex}

\section{Discussion {\&} Conclusion}
\label{sec:discussion}
\input{sections/discussion.tex}

\section*{Acknowledgemets}
We thank Sankalan Pal Chowdhury, Mansi Gupta, Gantavya Bhatt, Atishya Jain, Kundan Krishna and Vishal Sharma for
their insightful comments and help with the paper. 
Mausam is supported by IBM AI Horizons Network for grant, an IBM SUR award, grants
by Google, Bloomberg and 1MG, Jai Gupta Chair Fellowship and a Visvesvaraya faculty award by Govt. of India. We thank
IIT Delhi HPC facility for computational resources.

\bibliography{anthology,emnlp2020}
\bibliographystyle{acl_natbib}

\clearpage
\appendix

\section*{Supplementary Material}

\input{sections/appendix.tex}

\end{document}

%% file: sections/abstract.tex
Pooling-based recurrent neural architectures
consistently outperform
their counterparts
without pooling on sequence classification tasks.
However, the reasons for their
enhanced performance
are largely unexamined.
In this work, we explore %
three commonly
used pooling techniques 
(mean-pooling, max-pooling, and attention\footnote{\label{footnote:attention}Attention aggregates representations via a weighted sum, thus we consider it under the umbrella of pooling in this paper.}), 
and propose \emph{max-attention}, a novel variant 
that captures
interactions among predictive tokens in a sentence. 
Using novel experiments, we demonstrate that pooling architectures 
substantially differ 
from their non-pooling equivalents 
in their learning ability and positional biases: (i) pooling facilitates better gradient flow than BiLSTMs in initial training epochs, and (ii) BiLSTMs are biased towards tokens at the beginning and end of the input, whereas pooling alleviates this bias.
Consequently, we find that pooling 
yields
large gains in 
low resource scenarios, and instances when salient words lie towards the middle of the input.
Across several text classification tasks,
we find max-attention to frequently outperform 
other pooling techniques.\footnote{Code and data is made available at \href{https://github.com/dair-iitd/PoolingAnalysis}{https://github.com/dair-iitd/PoolingAnalysis}.}

%% file: sections/intro.tex
Pooling mechanisms are 
ubiquitous components
in Recurrent Neural Networks (RNNs) 
used for natural language tasks. %
Pooling operations consolidate
hidden representations from RNNs
into a single sentence representation.
Various pooling techniques, like mean-pooling, max-pooling, 
and attention, have been shown to improve
the performance
of RNNs on text classification tasks~\cite{Lai&al15,conneau&al17}.
Despite widespread adoption, precisely how and when pooling 
benefits the models is largely under-explored.

In this work, 
we perform an in-depth analysis comparing 
popular pooling methods, and proposed max-attention,
with standard BiLSTMs for several text classification tasks. 
We identify two key factors 
that explain the benefits of pooling techniques: 
learnability, and positional invariance. 

First, 
we analyze the flow of gradients for different classification tasks 
to assess the learning ability of BiLSTMs (\S\ref{sec:gradients}). 
We observe that the gradients corresponding to hidden representations in the middle of the sequence vanish during the initial epochs.
On training for more examples, these gradients slowly recover, 
suggesting that the gates of standard BiLSTMs require many examples to learn. 
In contrast, we find the gradient norms in pooling-based architectures to be 
free from this problem. 
Pooling enables a fraction of the gradients to
directly reach any hidden 
state instead of having to backpropagate through a long series of recurrent cells. %
Thus we hypothesize, and subsequently confirm, that pooling 
is particularly beneficial for tasks with long input sequences.%

Second, we explore the positional biases of 
BiLSTMs, with and without pooling  (\S\ref{sec:positional-biases}).
Across several classification tasks, using various novel experimental setups, 
we expose that BiLSTMs are less responsive to 
tokens towards the middle of the sequence, when compared to tokens at the beginning or the end of the sequence. 
However, we find that this bias is largely absent in pooling-based architectures, 
indicating their ability to respond to salient tokens regardless of their position.

Third,
we propose 
max-attention, a novel pooling technique, 
which combines 
the advantages of 
max-pooling and attention (\S\ref{subsec:max-attention}).
Max-attention uses the max-pooled representation as its query vector
to compute the attention weights for each hidden state. %
Max-pooled representations are extensively used in the literature to capture 
prominent tokens (or objects) in a sentence (or an image)~\cite{zhang2015sensitivity,boureau2010theoretical}.
Therefore, using them as a query vector 
effectively captures interactions among salient portions in the input.
Max-attention is simple to use, and yields performance gains over other pooling methods
on several classification setups.

%% file: sections/related.tex
\paragraph{Pooling:} 
A wide body of work compares the performance of different pooling techniques in object recognition tasks~\cite{boureau2010learning, boureau2010theoretical, boureau2011ask}
and finds max-pooling to generally outperform mean-pooling.
However, pooling in natural language
tasks is relatively understudied.
For some text classification tasks,
pooled recurrent architectures~\cite{Lai&al15,zhang2015sensitivity, johnson&zhang16,jacovi2018understanding, yang2016hierarchical},
outperform CNNs 
and BiLSTMs.
Additionally, for textual entailment tasks, ~\citet{conneau&al17} find that max-pooled representations better capture 
salient words in a sentence.
Our work extends the analysis 
and examines several pooling techniques,
including attention, 
for BiLSTMs applied to natural language tasks.
While past approaches assess the ability of pooling in capturing linguistic phenomena, to the best of our knowledge, 
we are the first to systematically study the training
advantages of various pooling techniques. 

\paragraph{Attention:}
First proposed as a way to align target tokens to the source tokens in translation~\cite{attention_first_paper},
the core idea behind attention---learning a weighted sum of 
the 
hidden states---has been widely adopted.
As attention aggregates hidden representations, %
we consider
it 
under the umbrella of pooling.
Recently, ~\citet{pruthi2020learning} conjecture that attention offers benefits during
training;
our work 
explains, and provides empirical evidence to support the speculation.

\paragraph{Gradient Propagation:}
Vanilla RNNs %
are 
known
to suffer from the problem of vanishing and exploding gradients~\cite{hochreiter&al91,bengio&al94}. 
In response,
~\citet{hochreiter&al97} invented LSTMs, 
which provide a direct connection passage through all the cells in order to remember new inputs without forgetting prior history.
However, 
recent work suggests that LSTMs 
do not solve this problem completely~\cite{arjovsky&al15,chandar&al19}.
Our work quantitatively investigates this phenomenon, 
exposing scenarios where the effect is pronounced, and demonstrating how pooling techniques
mitigate the problem, leading to better sample efficiency, and generalization.

%% file: sections/methods.tex
\begin{figure*}[t]
\centering
\includegraphics[width=\textwidth]{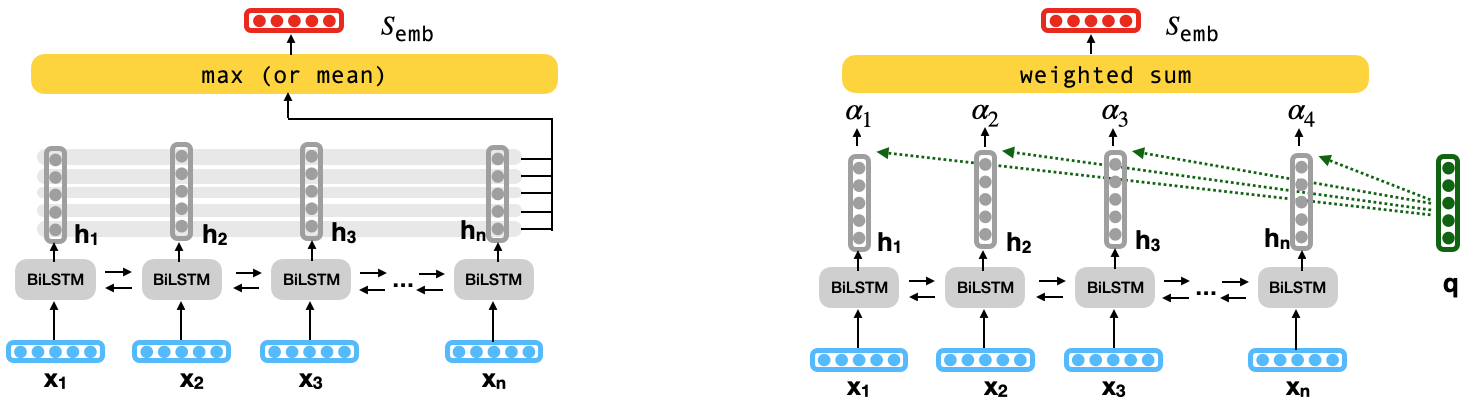}
\caption{A pictorial overview of the pooling techniques. Left: element-wise mean and max pooling operations aggregate hidden representations. Right: attention scores ($\alpha$) are computed using the similarity between hidden representations (${h})$ and query vector (${q}$), which are subsequently used to weight hidden representations. Our proposed max-attention uses the sentence embedding from max-pooling as a query to attend over hidden states.}
\label{fig:PoolingSummary}
\end{figure*}
\subsection{Background and Notation}
Let $s = \{x_1, x_2, \ldots, x_n\}$ be an input sentence, 
where $x_t$ is a representation of the input word at position $t$. 
A recurrent neural network such as an LSTM produces a hidden state $h_t$, 
and a cell state $c_t$ for each input word $x_t$, 
where $h_{t}, c_{t} = \phi(h_{t-1}, c_{t-1}, x_{t})$. 
Standard BiLSTMs concatenate the first hidden state of the backward LSTM, and the last hidden state of the forward LSTM for the final sentence representation: 
$s_{\text{emb}} = [\overrightarrow{h_n}, \overleftarrow{h_1}]$. 
The sentence embedding ($s_{\text{emb}}$) is further 
fed to a downstream text classifier. 
For training BiLSTMs, multiple works have emphasized the importance of initializing the bias for forget gates to a high value (between 1-2) to prevent the model from forgetting information before it learns what to forget~\cite{gers&al99,westhuizen2018unreasonable}. Hence, in our analysis, 
we experiment with both a high and low value of bias for the forget gate. For the non-pooled \lastf{}, we initialize the forget gate bias to 1, unless specified.
For brevity, from hereon we would use $h_t$ to mean $[\overrightarrow{h_t}, \overleftarrow{h_t}]$. Below, we formally discuss popular pooling techniques:

\paragraph{Max-pooling:}
For a max-pooled BiLSTM (\maxout{}), the sentence embedding $s_{\text{emb}}$, is:
\begin{equation*}
s_{\text{emb}}^{i} = \max_{t \in (1,n)}(h_{t}^{i}) 
\end{equation*}
where $h_{t}^{i}$ represents the $i^{\text{th}}$ dimension of the hidden state corresponding to the word at position $t$. This implies that while backpropagating the loss, we find a direct pathway to the $t^{\text{th}}$ hidden state as:
\begin{equation*}
\frac{\partial s_{\text{emb}}^{i}}{\partial h_{t}^{i}}=
        \begin{cases}
        1, & \text{if} ~t = \rm{arg}\!\max_{t \in (1,n)}h_{t}^{i}\\
        \frac{\partial h_{k}^{i}}{\partial h_{t}^{i}}, & \text{if} ~k = \rm{arg}\!\max_{t \in (1,n)}h_{t}^{i}, k \neq t
        \end{cases}
\end{equation*}
\noindent Similarly, in \textbf{mean-pooling} (\meanout{}), $s_{\text{emb}}$ is an average over all the hidden states.\footnote{Refer to Appendix~\ref{app:Mean_eqns} for the mathematical formulation.}

\paragraph{Attention:} Attention (\att{}) works by calculating a non-negative weight for each 
hidden state
that together sum to 1. Hidden representations are 
then multiplied with these weights and summed, resulting in 
a fixed-length vector~\cite{attention_first_paper,luong-etal-2015-effective}: %
\begin{align*}
\alpha_{t} = \frac{\exp({h_{t}^{\top}q})}{\sum_{j=1}^n\exp(h_{j}^{\top}q)}; \quad
s_{\text{emb}} = \sum_{t=1}^n\alpha_{t} h_{t}
\end{align*}
where \textit{q} is a learnable query vector.
Several variations like hierarchical attention \cite{hierarchical-attention}, self-attention \cite{attention_classification} have been proposed for text classification. However, 
the above formulation (referred in literature as ``Luong attention'') is most widely used in text classification tasks~\cite{jain2019attention, pinter2019attention,pruthi2020learning}.%

\subsection{Max-attention}
\label{subsec:max-attention}
We introduce a novel pooling variant called max-attention (\attmax{})
to capture inter-word dependencies. It uses the max-pooled hidden representation as the query vector for attention. Formally:
\begin{equation*}
\begin{aligned}
q^{i} &= \max_{t \in (1,n)}(h_{t}^{i}); 
&\hat{h_{t}} &= h_{t}/\|h_{t}\|\\ 
\alpha_{t} &= \frac{\exp(\hat{h_{t}}^{\top}q)}{\sum_{j=1}^n\exp(\hat{h_{j}}^{\top}q)}; 
&s_{\text{emb}} &= \sum_{t=1}^n \alpha_{t}h_{t} 
\end{aligned}
\end{equation*}
It is worth noting that the learnable query vector in Luong attention
is the same for the entire corpus, 
whereas in max-attention 
each sentence has a unique locally-informed query.
Previous literature extensively uses max-pooling to capture the 
prominent tokens (or objects) in a sentence (or image). 
Hence, using max-pooled representation as a query 
for attention 
allows for a second round of aggregation among important hidden representations.
\subsection{Transformers}
We briefly experiment with 
transformer architectures~\cite{vaswani2017attention}, and observe that
purely attention-based architectures 
perform poorly on text-classification without significant pre-training. 
Further, the memory footprint for transformers is $O(n^2)$ vs $O(n)$ for LSTMs.
Thus, for long examples in our experiments ($\sim4000$ words), %
XL-Net~\cite{yang2019xlnet} runs out of memory even for a batch size of 1 on a 32GB GPU.

We observe that text classification using representations corresponding to \texttt{[CLS]} token in pretrained transformers (such as RoBERTa \cite{liu&al19}) yields near state-of-art performance. Alternate ways to pool feature representations result in a marginal difference in performance ($\sim0.2\%$ on IMDB sentiment analysis). Pooling does not benefit transformers as they do not suffer from vanishing gradients and positional biases which pooling helps to mitigate in LSTMs (\S~\ref{sec:gradients},\S~\ref{sec:positional-biases}). Therefore, we limit the scope of this work to recurrent architectures.

%% file: sections/datasets.tex
We experiment with four different text classification tasks: 
(1) The \textbf{IMDb} dataset~\cite{IMDB} contains movie reviews and their associated sentiment label;
(2) \textbf{Yahoo! Answers}~\cite{Zhang&al15} dataset comprises $1.4$ million question and answer pairs, spread across $10$ topics, where the task is to predict the topic of the answer, using the answer text; 
(3) \textbf{Amazon} reviews~\cite{ni-etal-2019-justifying} contain product reviews from the Amazon website, filtered by their category. We construct a $20$-class classification task using these reviews\footnote{Appendix~\ref{app:datasets} contains further details about the dataset.};
(4) \textbf{Yelp} Reviews~\cite{Zhang&al15} is another sentiment polarity classification task.

For these datasets, we only use the text and labels, ignoring any auxiliary information (like title or location). 
We select subsets of the datasets with sequences having greater than $100$ words to better understand the impact of vanishing gradients and positional bias in recurrent architectures. 
A summary of statistics is presented in Table~\ref{table:corpus_statistics}. 
\begin{table}[htb]
        \small \centering
        \begin{tabular}{@{}lccccc@{}}
        \toprule
        Dataset & Classes & \makecell{Avg. \\ Length} & \makecell{Max \\ Length} & \makecell{Train \\ Size} & \makecell{Test \\ Size} \\ 
        \midrule
        IMDb   & 2       & 240.4       & 2470       & 20K        & 9.8K \\
        Yahoo! Answers  & 10      & 206.2       & 998        & 25K        & 4.8K \\
        Amazon Reviews  &20  &185.6 &500   &25K      &12.5K \\
        Yelp Reviews   & 2       & 202.4    & 1000 & 25K    & 9.5K\\
        \bottomrule
        \end{tabular}
        \caption{Corpus statistics for classification tasks.}
        \label{table:corpus_statistics}
\end{table}

In all the experiments, we use a single-layered \last{} with hidden dimension size of $256$ and embedding dimension size of $100$ (initialized with GloVe vectors~\cite{pennington&al14} trained on a 6 billion word corpus). %
The sentence embeddings generated by the BiLSTM are passed to a final classification layer to obtain per-class probability distributions.
We train our models using Adam optimizer~\cite{kingma2014adam}, with a learning rate of $2 \times 10^{-3}$. The batch size is set to 32 for all the experiments. We train for $20$ epochs and select the model with the best validation accuracy. All experiments are repeated over 5 random seeds
using a single GPU (Tesla K40).\footnote{Further details to aid reproducibility are in the Appendix~\ref{app:reproducibility}.}

%% file: sections/gradients.tex
\label{sec:vanishing-gradients}
\begin{figure*}[htp]
		\includegraphics[width=0.32\textwidth]{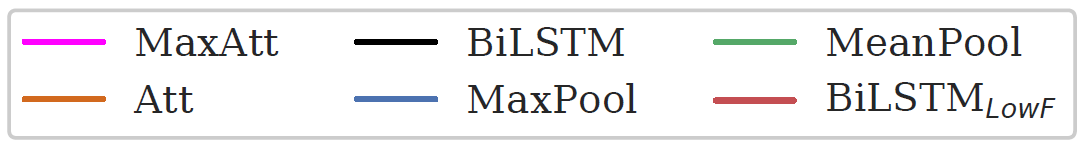}
		\hfill
		\includegraphics[width=0.64\textwidth]{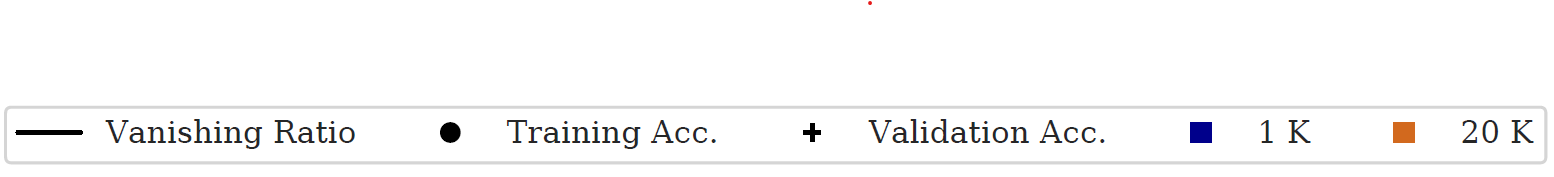}\\
        \subcaptionbox{Gradient Norms\label{fig:vanishing_graph}}{\includegraphics[width=0.32\textwidth]{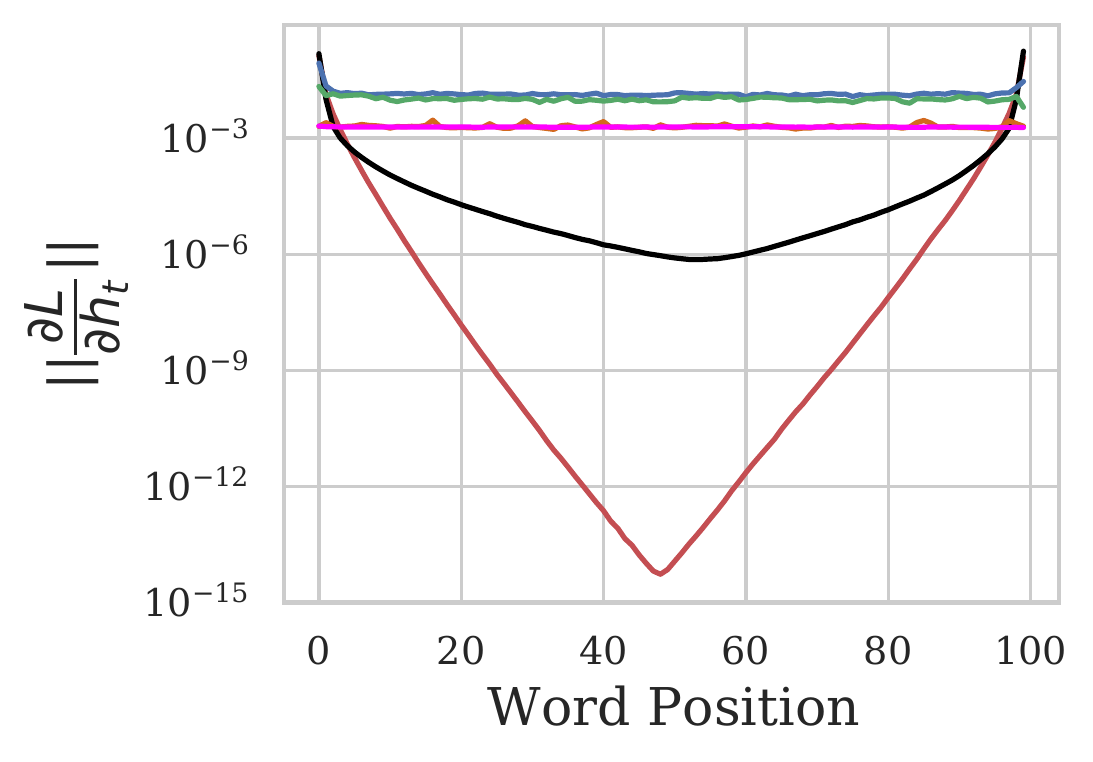}}
        \subcaptionbox{\lastf\label{fig:last_ratios}}{\includegraphics[width=0.34\textwidth]{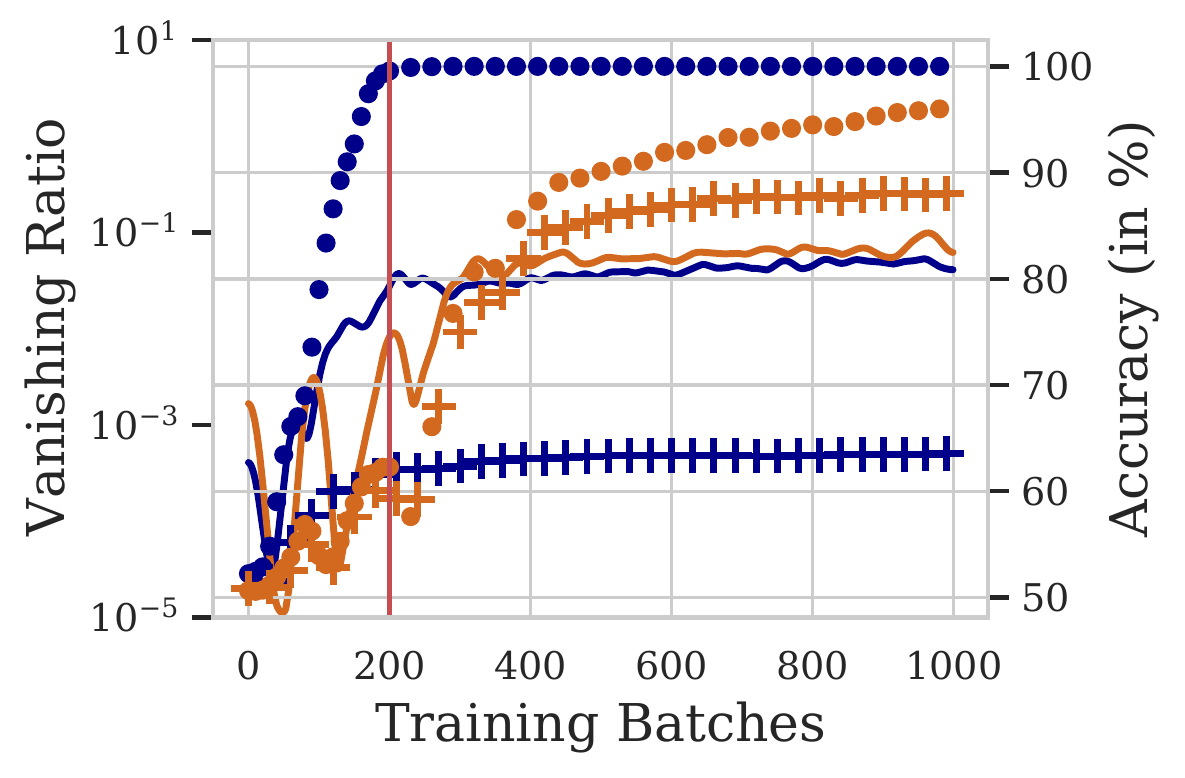}}
         \subcaptionbox{\attmax\label{fig:maxatt_ratios}}{\includegraphics[width=0.34\textwidth]{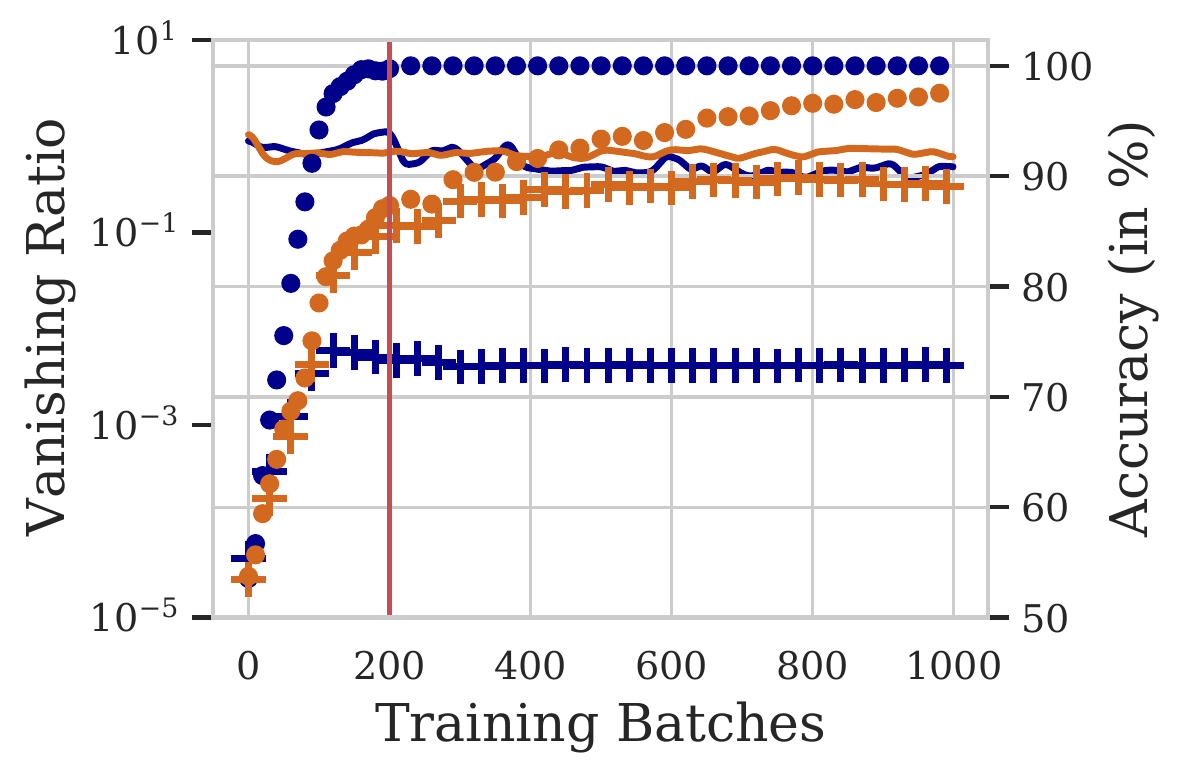}}
         
        \caption{(a): The gradient norm $(\lVert\frac{\partial L}{\partial h_{t}}\rVert)$ across different word positions. \last{}\textsubscript{LowF} suffers from extreme vanishing of  gradients, with the gradient norm in the middle nearly $10^{-15}$ times that at the ends. In contrast, pooling methods result in gradients of nearly the same value, irrespective of the word position. (b), (c): The vanishing ratio $(\lVert\frac{\partial L}{\partial h_{\text{mid}}}\rVert$$/$$\lVert\frac{\partial L}{\partial h_{\text{end}}}\rVert)$ over training batches for \last{} and \attmax{}, using $1$K, $20$K unique training examples from the IMDb dataset. The respective training and validation accuracies are also depicted. }
        \label{fig:vanishing_ratios}
\end{figure*}

In this section, we study the flow of gradients in different architectures and training regimes.
Pooling techniques
used in conjunction with BiLSTMs 
provide a direct gradient pathway to intermediate hidden states. 
However for BiLSTMs without pooling, 
it is crucial that the parameters for the input, output, and forget gates are appropriately learned 
so that the loss backpropagates across long input sequences, without the gradients vanishing. %
\paragraph{Experimental Setup:} 
In order to quantify the extent to which the gradients vanish across different word positions, we compute the gradient of the loss function w.r.t the hidden state at every word position $t$, and study their $\ell_2$ norm ($\lVert\frac{\partial L}{\partial h_{t}}\rVert$). 
To aggregate the gradients across multiple training examples (of different lengths), we 
linearly interpolate the distribution of gradient values for each example to a fixed length between $1$ and $100$.
The gradient values at each (normalized) position are averaged across all the training examples. We plot these values (on a $\log$ scale) after training on the first $500$ IMDb reviews to study the effect of gradient vanishing at the beginning of training (Figure~\ref{fig:vanishing_graph}). 

To understand how the distribution of gradients (across word positions) changes with the number of training batches, we compute the ratio of the gradient norm corresponding to the word at the middle and word at the end: 
$\lVert\frac{\partial L}{\partial h_{\text{mid}}}\rVert$ $/$ $\lVert\frac{\partial L}{\partial h_{\text{end}}}\rVert$.\footnote{Implementation detail: we choose the left end, as some sequences in a batch might be padded with zeros on the right.}
We call this the \emph{vanishing ratio} and use it as a measure to quantify the extent of vanishing (where lower values indicate severe vanishing). %
Each training batch on the x-axis in Figures~\ref{fig:last_ratios},~\ref{fig:maxatt_ratios} corresponds to $64$ training examples.

\paragraph{Results}
It is evident from Figure~\ref{fig:vanishing_graph}
that the gradients vanish significantly for \lastf{}, with $\lVert\frac{\partial L}{\partial h_{t}}\rVert$ falling to the order of $10^{-6}$ as we approach the middle positions in the sequence. This effect is even more pronounced for the case of BiLSTM$_{\text{LowF}}$, which uses the Xavier initialization~\cite{xavier&al10} for the bias of the forget-gate.  
The plot suggests that specific initialization of the gates with best practices (such as setting the bias of forget-gate to a high value) helps to reduce the extent of the issue, but the problem still persists. 
In contrast, none of the pooling techniques face this issue, 
resulting in an almost straight line.

Additionally, from Figure~\ref{fig:last_ratios} we note that the problem of vanishing gradients is more pronounced at the beginning of training, when the gates are still untrained. The problem continues to persist, albeit to a lesser degree, until later in the training process. This specifically limits the performance of \lastf{} in resource-constrained settings, with fewer training examples.
For instance, in the $1$K training data setting, \lastf{} has an extremely low value of vanishing ratio ($\sim10^{-3}$) at the $200^{\text{th}}$ training batch (denoted by red vertical line in the plot), when it achieves nearly 100\% accuracy on the training data.\footnote{Refer to Appendix~\ref{app:vanishing} for plots of other pooling techniques.} 

\begin{table}
	\small
	\centering
	\scalebox{1.05}{
	\begin{tabular}{@{}lrrrr|rrrrr@{}}
	\toprule
	& \multicolumn{4}{c}{Vanishing ratio}  & \multicolumn{5}{c}{Validation acc.}    \\ %
	\midrule
	\multicolumn{1}{c}{} & 
	\multicolumn{1}{c}{1K} & \multicolumn{1}{c}{5K} & \multicolumn{1}{c}{20K} & 
	\multicolumn{1}{c}{} & \multicolumn{1}{c}{} & 
	\multicolumn{1}{c}{1K} & \multicolumn{1}{c}{5K} & \multicolumn{1}{c}{20K} & 
	
	\\ \cmidrule(r){2-5} \cmidrule(r){7-10} %
	
	\lastf{}   
    & 5$\times10^{-3}$ & 0.03 & 0.06 &&
    & 64.9 & 82.8 & 88.4 \\
	\meanout{} 
	& 2.5 & 0.56 & 1.32 &&
    & 78.4 & 82.6 & 88.5 \\
	\maxout{}  
    & 0.40 & 0.42 & 0.53 &&
     & 78.0 & 84.7 & 89.6 \\
    \att{}     
    & 3.87 & 1.04 & 1.19 &&
    & 77.1 & 84.6 & 90.0 \\
	\attmax{}  
    & 0.69 & 0.69 & 0.64 &&
    & 78.1 & 86.0 & 90.2 \\
	\bottomrule
	\end{tabular}}
	  \caption{Values of vanishing ratio
	  as computed 
	  when different models achieve 95\% training accuracy, along with the best validation accuracy for that run.}
	  \label{table:vanish_table}
\end{table}

Consequently, the \lastf{} model (prematurely) achieves a high training accuracy, solely based on the starting and ending few words, well before the gates can learn to allow the gradients to pass through (and mitigate the vanishing gradients problem).
Further reduction in vanishing ratio is unable to improve validation accuracy, due to saturation in training.
To examine this more closely, we tabulate the vanishing ratios at the point where the model reaches 95\% accuracy on the training data in Table~\ref{table:vanish_table}. A low value at this point indicates that the gradients are still skewed towards the ends, even as the model begins to overfit on the training data.
The vanishing ratio is low for \last{}, especially in low-data settings. This results in a 13-14\% lower test accuracy in the $1$K data setting, compared to other pooling techniques. %
We conclude that the phenomenon of vanishing gradients results in poorer performance of BiLSTMs. 
Encouragingly, pooling methods do not exhibit low vanishing ratios, right from the beginning of training, leading to performance gains as demonstrated in the next section.

%% file: sections/positional-biases.tex
\begin{figure*}[!htb]
	\centering
		\includegraphics[width=0.6\textwidth]{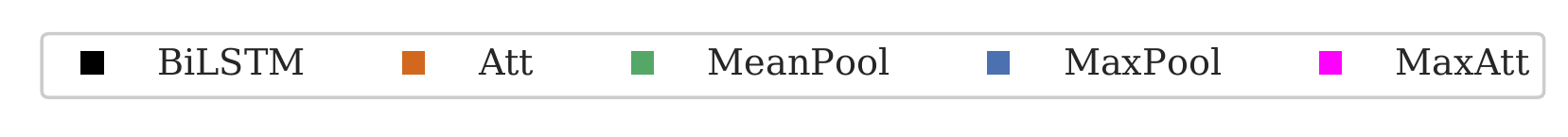}\\
		\subcaptionbox{Left \label{fig:IMDB:WIKI_L}}{\includegraphics[width=0.32\textwidth]{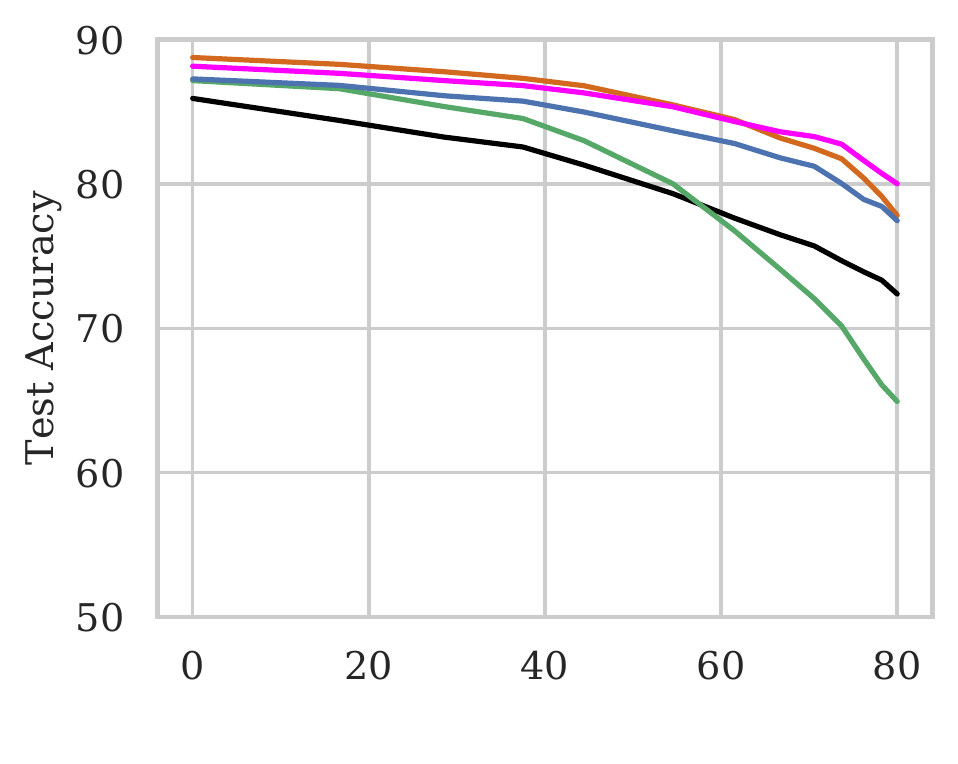}}
		\hfill
		\subcaptionbox{Mid \label{fig:IMDB:WIKI_M}}{\includegraphics[width=0.3\textwidth]{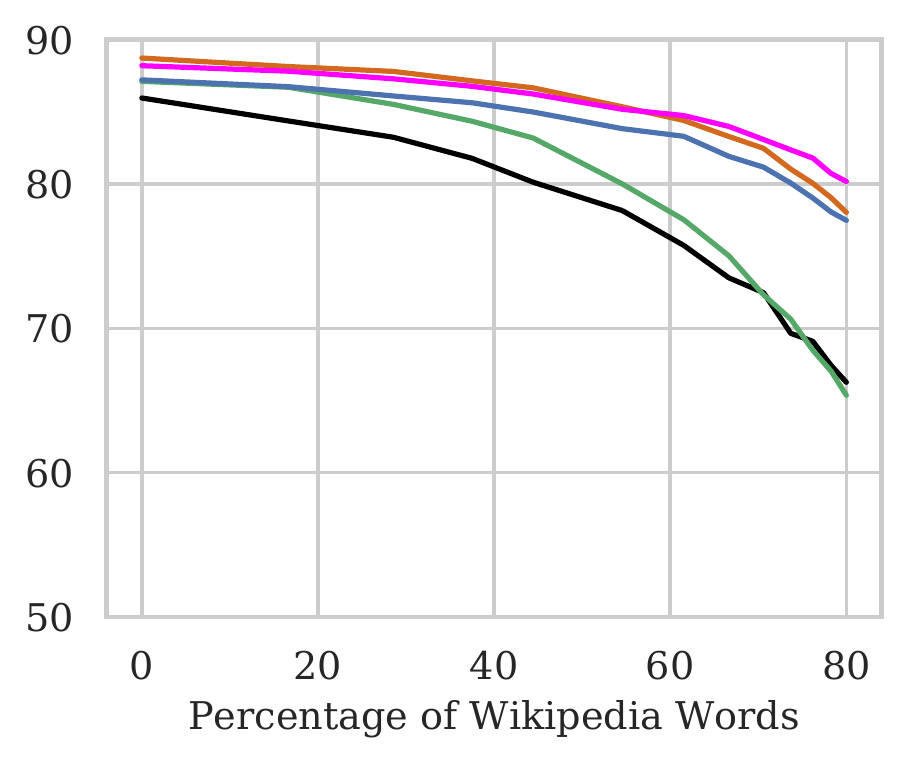}}
		\hfill
		\subcaptionbox{Right \label{fig:IMDB:WIKI_r}}{\includegraphics[width=0.3\textwidth]{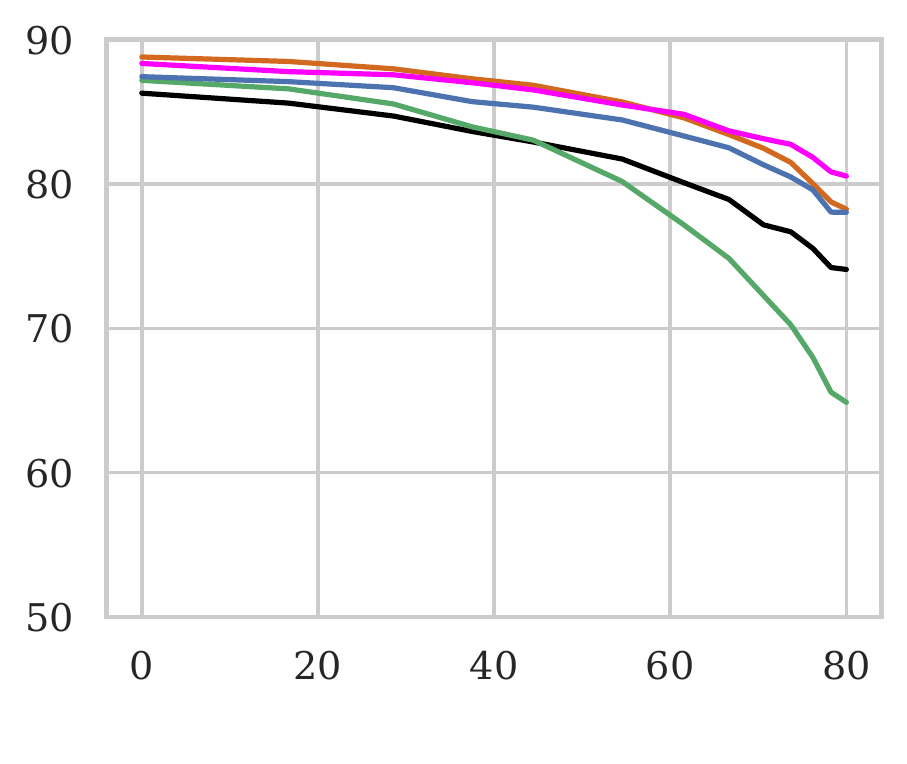}}
		\hfill
	\caption{For models trained on 10K examples, varying amounts of random Wikipedia sentences are appended to the original IMDb reviews \emph{at test time}. Original review is preserved on the (a) left; (b) middle; and (c) right of the modified input. Performance degrades significantly for \lastf{} and \meanout{}, whereas \att{}, \maxout{} and \attmax{} are more resilient.} 
	\label{fig:IMDB:wiki-attack-5K}
\end{figure*}
\input{tables/table1_updated.tex}
Analyzing the gradient propagation in BiLSTMs suggests 
that standard recurrent networks are biased towards 
the end tokens, 
as the overall contribution of distant hidden states
is extremely low in the gradient of the loss. 
This implies that the weights of various parameters in an 
LSTM cell (all cells of an LSTM have tied weights) are 
hardly influenced by the middle words of the sentence. 
In this light, we aim to 
evaluate positional biases
of recurrent architectures with different pooling techniques. %
\subsection{Evaluating Natural Positional Biases}
\label{subsec:effect_rand_noise}
\begin{figure}[t]
\centering
\includegraphics[width=\columnwidth]{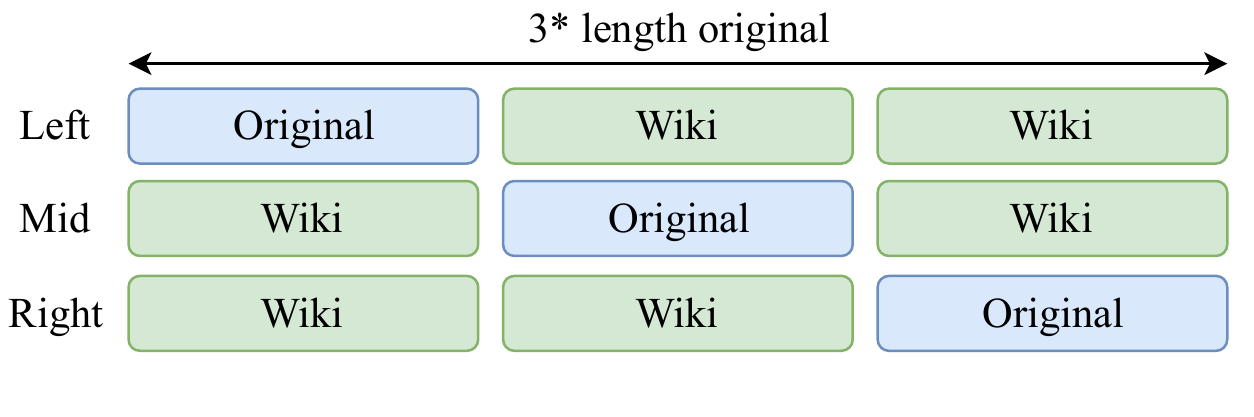} 
\caption{Explaining Wikipedia sentence addition.}
\label{fig:Wikipedia_Explain}
\end{figure}

\emph{Can organically trained recurrent models skip over unimportant words 
on either ends of the sentence?}
\paragraph{Experimental Setup:} We append 
randomly chosen Wikipedia sentences 
to the input examples of two text classification tasks, based on IMDb and Amazon Reviews, \emph{only at test time}, 
keeping the training datasets unchanged. 
Wikipedia sentences are declarative statements of fact, and should not influence the sentiment of movie reviews, and given the diverse nature of the Wikipedia sentences it is unlikely that 
they would interfere with the few categories (i.e. the labels) of Amazon product reviews. 
Therefore, 
it is not unreasonable to expect the models
to be robust to such random noise,
even though they were not trained for the same. 
We perform this experiment in three configurations, 
such that original input
is preserved on the (a) left, (b) middle, and (c) right of the modified input. 
For these configurations, we vary the length of added
Wikipedia text in proportion to the length of the original sentence.  Figure~\ref{fig:Wikipedia_Explain} illustrates the setup when 66\% of the total words come from Wikipedia.
\paragraph{Results:} The effect of
adding random words can be seen in Figure~\ref{fig:IMDB:wiki-attack-5K}. 
We draw two conclusions: %
(1) Adding random sentences on both ends is more detrimental to the performance of \lastf{} as compared to the scenario where the input is appended to only one end.\footnote{One practical implication of this finding is that adversaries can easily attack middle portions of the input text.}
This corroborates our previous findings that these models
largely account for information at the ends for their predictions.
(2) We speculate that paying equal importance to all hidden states prevents \meanout{} from distilling out important information effectively, making it more susceptible to random noise addition.
On the contrary, both max-pooling and attention based architectures like \maxout{}, \att{} and \attmax{} are significantly more robust in all the settings. This indicates that max-pooling and attention can help account for salient words and ignore unrelated ones, regardless of their position.
Lastly, we provide concurring results on the Amazon dataset, and examine the robustness of different models given lesser training data in Appendix~\ref{app:wiki_attack:full}.

\subsection{Training to Skip Unimportant Words}\label{subsec:coarse-grained}
\emph{How well can different models be trained to skip unrelated words?}
\paragraph{Experimental setup:} 
We create new training datasets by
appending random 
Wikipedia sentences
to the original input examples of the datasets described in \S~\ref{sec:datasets}, such that 66\% of the text of each new training example comes from Wikipedia sentences
(see Figure~\ref{fig:Wikipedia_Explain}). 
We experiment with a varying number of training examples, %
however, the test set remains the same for fair comparisons.

\paragraph{Results}
The results are presented in Table \ref{table:wiki-summary}.
First, we note that 
\lastf{} severely suffers when random sentences are appended at both ends. 
In fact, the accuracy of \lastf{} in mid settings drops to 50\%, 12\%, 5\%, 50\% on IMDb, 
Yahoo, Amazon, Yelp datasets respectively, which is equal to the majority class baseline.
However, the performance drop (while large) is not as drastic when 
sentences are added to only one 
end of the text. %
We speculate that this is because a \lastf{} is composed of a  forward and a backward LSTM, and when random sentences are appended to the left, the backward LSTM is able to 
capture information about the original sentence on the right and vice versa. 

Second, 
while accuracies of all pooling techniques begin to converge given sufficient data, the differences in low training data regime are substantial. 
Further, the poor performance of \lastf{} re-validates the findings of \S~\ref{sec:vanishing-gradients}, 
where we hypothesize that the model's training saturates before 
the gradients can learn to reach the middle tokens.\footnote{Results on more dataset sizes, and the `left' setting are in Appendix~\ref{app:wiki-full_eval}. Conclusions drawn from the `right' setting are in line with the observations from the `left'.} 

Third, when the number of classes is large (as in Yahoo and Amazon datasets), we observe a significant performance difference between \att{} and \attmax{}. We speculate that as the number of labels increase, a single global query vector (as in \att{}) is inadequate to identify important words relevant to each label, whereas a sentence dependent query (as in \attmax{)} mitigates this concern.

\paragraph{Evaluation on Short Sentences}
Finally, we re-evaluate this experiment on (new) datasets with short sentences ($<100$ words). Results for the standard and `mid' settings are presented in Table~\ref{table:wiki-short-mini}.
Unlike long sequences, where BiLSTM model was no  
better than majority classifier (see Table~\ref{table:wiki-summary}),
with shorter sequences, 
the \lastf{} model performs better.
This result supports our hypothesis that the effect of vanishing gradients is prominent in longer sequences.\footnote{Refer to Appendix~\ref{app:wiki-short_sent} for full evaluation.}
Overall, among all the scenarios discussed in \cref{table:wiki-summary,table:wiki-short-mini}, on comparing all pooling methods (and \last{}) on the basis of their mean test accuracy, \textbf{\attmax{} is the best performing model in about 80\% cases, \att{} in 18\% cases.}

\input{tables/table_short_mini.tex}
\input{sections/NWI.tex}

%% file: tables/table1_updated.tex
\begin{table*}[t]

\small
\centering
\scalebox{1.0}{
\begin{tabular}{@{}lrrrr|rrrrr|rrrr@{}}

\toprule
& \multicolumn{4}{c}{IMDb}  & \multicolumn{5}{c}{IMDb (mid) + Wiki}    & \multicolumn{4}{c}{IMDb (right) + Wiki} \\
\midrule
\multicolumn{1}{c}{} & 
\multicolumn{1}{c}{1K} & \multicolumn{1}{c}{2K} & \multicolumn{1}{c}{10K} & 
\multicolumn{1}{c}{} & \multicolumn{1}{c}{} & 
\multicolumn{1}{c}{1K} & \multicolumn{1}{c}{2K} & \multicolumn{1}{c}{10K} & 
\multicolumn{1}{l}{} & \multicolumn{1}{c}{} & 
\multicolumn{1}{c}{1K} & \multicolumn{1}{c}{2K} & \multicolumn{1}{c}{10K} 

\\ \cmidrule(r){2-5} \cmidrule(r){7-10} \cmidrule(r){12-14}

\lastf{}   	& 64.7 \tiny $\pm$ 2.3 		& 75.0 \tiny $\pm$ 0.4 & 86.6 \tiny $\pm$ 0.8 &  &
			& 49.6 \tiny $\pm$ 0.7      & 49.9 \tiny $\pm$ 0.5 & 50.3 \tiny $\pm$ 0.3 &  &
			& 53.5 \tiny $\pm$ 2.5      & 64.7 \tiny $\pm$ 2.8 & 85.9 \tiny $\pm$ 0.5 \\
\meanout{} 	& 73.0 \tiny $\pm$ 3.0 		& 81.7 \tiny $\pm$ 0.7 & 87.1 \tiny $\pm$ 0.6 &  &
			& 69.8 \tiny $\pm$ 2.1      & 76.2 \tiny $\pm$ 1.0 & 84.1 \tiny $\pm$ 0.7 &  &
			& 70.0 \tiny $\pm$ 1.1      & 76.8 \tiny $\pm$ 1.0 & 84.8 \tiny $\pm$ 0.9 \\
\maxout{}  	& 69.0 \tiny $\pm$ 3.9 		& 80.1 \tiny $\pm$ 0.5 & 87.8 \tiny $\pm$ 0.6 &  &
			& 64.5 \tiny $\pm$ 1.8      & 77.2 \tiny $\pm$ 2.0 & 86.0 \tiny $\pm$ 0.8 &  &
			& 65.9 \tiny $\pm$ 4.6      & 77.8 \tiny $\pm$ 0.9 & \textbf{87.2} \tiny $\pm$ 0.6 \\
\att{}     	& 75.7 \tiny $\pm$ 2.6 		& \textbf{82.8} \tiny $\pm$ 0.8 & \textbf{89.0} \tiny $\pm$ 0.3 &  &
			& 75.0 \tiny $\pm$ 0.8      & 79.4 \tiny $\pm$ 0.8 & 86.7 \tiny $\pm$ 1.4 &  &
			& 74.7 \tiny $\pm$ 1.4      & 80.2 \tiny $\pm$ 1.8 & 87.1 \tiny $\pm$ 1.0 \\
\attmax{}  	& \textbf{75.9} \tiny $\pm$ 2.2 		& 82.5 \tiny $\pm$ 0.4 & 88.5 \tiny $\pm$ 0.5 &  &
			& \textbf{75.4} \tiny $\pm$ 2.4      & \textbf{80.9} \tiny $\pm$ 1.8 & \textbf{86.8} \tiny $\pm$ 0.5 &  &
			& \textbf{77.9} \tiny $\pm$ 0.9      & \textbf{81.9} \tiny $\pm$ 0.5 & \textbf{87.2} \tiny $\pm$ 0.5 \\ 

\toprule
& \multicolumn{4}{c}{Yahoo}  & \multicolumn{5}{c}{Yahoo (mid) + Wiki}    & \multicolumn{4}{c}{Yahoo (right) + Wiki} \\
\midrule
\multicolumn{1}{c}{} & 
\multicolumn{1}{c}{1K} & \multicolumn{1}{c}{2K} & \multicolumn{1}{c}{10K} & 
\multicolumn{1}{c}{} & \multicolumn{1}{c}{} & 
\multicolumn{1}{c}{1K} & \multicolumn{1}{c}{2K} & \multicolumn{1}{c}{10K} & 
\multicolumn{1}{l}{} & \multicolumn{1}{c}{} & 
\multicolumn{1}{c}{1K} & \multicolumn{1}{c}{2K} & \multicolumn{1}{c}{10K} 

\\ \cmidrule(r){2-5} \cmidrule(r){7-10} \cmidrule(r){12-14}

\lastf{}   & 38.3 \tiny$\pm$ 4.8 		& 51.4 \tiny$\pm$ 2.1 & 63.5 \tiny$\pm$ 0.6 &  &
  		   & 12.7 \tiny$\pm$ 1.1        & 12.7 \tiny$\pm$ 1.1 & 11.4 \tiny$\pm$ 0.8 &  &
  		   & 18.8 \tiny$\pm$ 2.5        & 37.3 \tiny$\pm$ 0.9 & 60.1 \tiny$\pm$ 1.5 \\
\meanout{} & 48.2 \tiny$\pm$ 2.3 		& 56.6 \tiny$\pm$ 0.5 & 64.7 \tiny$\pm$ 0.6 &  &
  		   & 31.9 \tiny$\pm$ 2.3        & 43.1 \tiny$\pm$ 2.0 & 58.5 \tiny$\pm$ 0.6 &  &
  		   & 33.9 \tiny$\pm$ 2.1        & 43.2 \tiny$\pm$ 1.0 & 58.6 \tiny$\pm$ 0.4 \\
\maxout{}  & 50.2 \tiny$\pm$ 2.1 		& 56.3 \tiny$\pm$ 1.8 & 63.9 \tiny$\pm$ 1.1 &  &
  		   & 33.0 \tiny$\pm$ 1.0        & 40.1 \tiny$\pm$ 1.4 & 58.4 \tiny$\pm$ 1.2 &  &
  		   & 33.1 \tiny$\pm$ 2.5        & 41.2 \tiny$\pm$ 0.9 & 60.9 \tiny$\pm$ 1.0 \\
\att{}     & 47.3 \tiny$\pm$ 2.2 		& 54.2 \tiny$\pm$ 1.1 & \textbf{65.1} \tiny$\pm$ 1.5 &  &
  		   & 39.4 \tiny$\pm$ 0.5        & 45.1 \tiny$\pm$ 1.8 & 61.5 \tiny$\pm$ 1.7 &  &
  		   & 37.9 \tiny$\pm$ 1.4        & 47.6 \tiny$\pm$ 2.3 & 62.2 \tiny$\pm$ 0.9 \\
\attmax{}  & \textbf{51.8} \tiny$\pm$ 1.1 		& \textbf{57.0} \tiny$\pm$ 1.1 & \textbf{65.1} \tiny$\pm$ 1.1 &  &
  		   & \textbf{39.6} \tiny$\pm$ 0.9        & \textbf{48.5} \tiny$\pm$ 0.6 & \textbf{62.2} \tiny$\pm$ 1.6 &  &
  		   & \textbf{40.3} \tiny$\pm$ 1.5        & \textbf{50.1} \tiny$\pm$ 1.6 & \textbf{63.1} \tiny$\pm$ 0.7 \\

\toprule
& \multicolumn{4}{c}{Amazon}  & \multicolumn{5}{c}{Amazon (mid) + Wiki}    & \multicolumn{4}{c}{Amazon (right) + Wiki} \\
\midrule
\multicolumn{1}{c}{} & 
\multicolumn{1}{c}{1K} & \multicolumn{1}{c}{2K} & \multicolumn{1}{c}{10K} & 
\multicolumn{1}{c}{} & \multicolumn{1}{c}{} & 
\multicolumn{1}{c}{1K} & \multicolumn{1}{c}{2K} & \multicolumn{1}{c}{10K} & 
\multicolumn{1}{l}{} & \multicolumn{1}{c}{} & 
\multicolumn{1}{c}{1K} & \multicolumn{1}{c}{2K} & \multicolumn{1}{c}{10K} 

\\ \cmidrule(r){2-5} \cmidrule(r){7-10} \cmidrule(r){12-14}

\lastf{}   & 38.5 \tiny$\pm$ 4.2 		& 52.7 \tiny$\pm$ 7.7 & 76.2 \tiny$\pm$ 0.7 &  &
  		   & 5.3 \tiny$\pm$ 0.3         & 5.4 \tiny$\pm$ 0.3  & 5.1 \tiny$\pm$ 0.4  &  &
  		   & 7.9 \tiny$\pm$ 0.6         & 27.9 \tiny$\pm$ 9.9 & 70.8 \tiny$\pm$ 1.5 \\
\meanout{} & 44.8 \tiny$\pm$ 9.8 		& 55.6 \tiny$\pm$ 6.4 & 76.9 \tiny$\pm$ 0.4 &  &
  		   & 34.4 \tiny$\pm$ 3.5        & 52.7 \tiny$\pm$ 3.5 & 70.3 \tiny$\pm$ 1.7 &  &
  		   & 33.3 \tiny$\pm$ 1.0        & 48.2 \tiny$\pm$ 3.4 & 71.9 \tiny$\pm$ 0.8 \\
\maxout{}  & 49.6 \tiny$\pm$ 3.9 		& 61.6 \tiny$\pm$ 2.6 &\textbf{79.1} \tiny$\pm$ 0.4 &  &
  		   & 17.0 \tiny$\pm$ 0.7        & 34.5 \tiny$\pm$ 2.0 & 72.8 \tiny$\pm$ 0.6 &  &
  		   & 17.0 \tiny$\pm$ 1.7        & 36.5 \tiny$\pm$ 3.0 & 72.4 \tiny$\pm$ 0.3 \\
\att{}     & 54.1 \tiny$\pm$ 5.2 		& 61.2 \tiny$\pm$ 2.9 & 77.0 \tiny$\pm$ 0.3 &  &
  		   & 48.0 \tiny$\pm$ 1.7        & 59.1 \tiny$\pm$ 1.8 & \textbf{75.3} \tiny$\pm$ 0.5 &  &
  		   & 48.9 \tiny$\pm$ 1.5        & 58.9 \tiny$\pm$ 1.3 & \textbf{75.7} \tiny$\pm$ 0.3 \\
\attmax{}  &\textbf{58.2} \tiny$\pm$ 3.8 		&\textbf{65.6} \tiny$\pm$ 0.9 & 77.3 \tiny$\pm$ 0.2 &  &
  		   &\textbf{57.7} \tiny$\pm$ 0.5        &\textbf{63.0} \tiny$\pm$ 0.8 & 74.8 \tiny$\pm$ 0.5 &  &
  		   &\textbf{57.8} \tiny$\pm$ 0.8        &\textbf{63.7} \tiny$\pm$ 0.8 & 75.3 \tiny$\pm$ 0.3 \\

\toprule
& \multicolumn{4}{c}{Yelp}  & \multicolumn{5}{c}{Yelp (mid) + Wiki}    & \multicolumn{4}{c}{Yelp (right) + Wiki} \\
\midrule
\multicolumn{1}{c}{} & 
\multicolumn{1}{c}{1K} & \multicolumn{1}{c}{2K} & \multicolumn{1}{c}{10K} & 
\multicolumn{1}{c}{} & \multicolumn{1}{c}{} & 
\multicolumn{1}{c}{1K} & \multicolumn{1}{c}{2K} & \multicolumn{1}{c}{10K} & 
\multicolumn{1}{l}{} & \multicolumn{1}{c}{} & 
\multicolumn{1}{c}{1K} & \multicolumn{1}{c}{2K} & \multicolumn{1}{c}{10K} 

\\ \cmidrule(r){2-5} \cmidrule(r){7-10} \cmidrule(r){12-14}

\lastf{}		&80.7 \tiny$\pm$ 4.1		&84.9 \tiny$\pm$ 8.0		&93.1 \tiny$\pm$ 0.1
		&&&50.2 \tiny$\pm$ 0.4		&51.1 \tiny$\pm$ 0.9		&51.4 \tiny$\pm$ 0.7
		&&&59.4 \tiny$\pm$ 3.7		&79.6 \tiny$\pm$ 6.2		&92.7 \tiny$\pm$ 0.4\\
\meanout{}		&\textbf{87.1} \tiny$\pm$ 1.2		&\textbf{87.9} \tiny$\pm$ 1.7		&93.4 \tiny$\pm$ 0.3
		&&&79.2 \tiny$\pm$ 1.1		&86.7 \tiny$\pm$ 1.0		&92.7 \tiny$\pm$ 0.2
		&&&79.4 \tiny$\pm$ 0.9		&87.1 \tiny$\pm$ 0.6		&92.3 \tiny$\pm$ 0.4\\
\maxout{}		&84.4 \tiny$\pm$ 2.0		&86.4 \tiny$\pm$ 5.1		&93.4 \tiny$\pm$ 0.2
		&&&81.1 \tiny$\pm$ 1.5		&85.6 \tiny$\pm$ 0.6		&92.5 \tiny$\pm$ 0.4
		&&&80.6 \tiny$\pm$ 0.8		&86.7 \tiny$\pm$ 0.9		&\textbf{93.2} \tiny$\pm$ 0.2\\
\att{}		&82.5 \tiny$\pm$ 3.7		&85.6 \tiny$\pm$ 6.5		&\textbf{93.7} \tiny$\pm$ 0.2
		&&&84.4 \tiny$\pm$ 1.0		&89.3 \tiny$\pm$ 1.0		&92.5 \tiny$\pm$ 0.6
		&&&\textbf{84.8} \tiny$\pm$ 0.7		&89.1 \tiny$\pm$ 0.9		&92.8 \tiny$\pm$ 0.4\\
\attmax{}		&81.3 \tiny$\pm$ 5.1		&86.0 \tiny$\pm$ 6.3		&\textbf{93.7} \tiny$\pm$ 0.3
		&&&\textbf{85.1} \tiny$\pm$ 0.8		&\textbf{89.4} \tiny$\pm$ 0.5		&\textbf{92.9} \tiny$\pm$ 0.3
		&&&84.1 \tiny$\pm$ 2.5		&\textbf{89.5} \tiny$\pm$ 0.7		&93.0 \tiny$\pm$ 0.4\\

\bottomrule
\end{tabular}}
  \caption{Mean test accuracy ($\pm$ std) (in \%) across 5 random seeds. In low-resource settings, \attmax{} consistently outpeforms other pooling variants. The performance of different pooling methods converges with increase in data.}
  \label{table:wiki-summary}
\end{table*}

%% file: tables/table_short_mini.tex
\begin{table}[H]
\centering
\small
\scalebox{1.0}{
\begin{tabular}{@{}lrrr|rrr@{}}

\toprule
\multicolumn{1}{c}{} & \multicolumn{6}{c}{Datasets with Short Sentences}\\ 

\toprule
\multicolumn{1}{c}{} & \multicolumn{3}{c}{Yahoo}  & \multicolumn{3}{c}{Yahoo (mid) + Wiki}\\ 
\midrule
\multicolumn{1}{c}{} & 
\multicolumn{1}{c}{1K} & \multicolumn{1}{c}{10K} & 
\multicolumn{1}{c}{} & \multicolumn{1}{c}{} & 
\multicolumn{1}{c}{1K}  & \multicolumn{1}{c}{10K}    
\\
\cmidrule(r){2-4} \cmidrule(r){6-7} 

\lastf{}		&20.5 \tiny$\pm$ 2.9	&42.4 \tiny$\pm$ 0.2	
            &&  &9.9    \tiny$\pm$ 0.7	&24.2 \tiny$\pm$ 0.9 \\
\meanout{}		&23.1 \tiny$\pm$ 1.8	&43.0 \tiny$\pm$ 0.3	
            &&  &14.9 \tiny$\pm$ 2.2	&32.8 \tiny$\pm$ 0.8 \\
\maxout{}		&23.0 \tiny$\pm$ 2.8	&\textbf{43.3} \tiny$\pm$ 0.4	
            &&  &14.1 \tiny$\pm$ 2.6	&33.8 \tiny$\pm$ 1.2 \\
\att{}			&24.3 \tiny$\pm$ 1.1	&43.1 \tiny$\pm$ 0.2	
            &&  &16.9 \tiny$\pm$ 3.0	&37.6 \tiny$\pm$ 0.5 \\
\attmax{}		&\textbf{25.1} \tiny$\pm$ 2.2	&\textbf{43.3} \tiny$\pm$ 0.3	
            &&  &\textbf{18.2} \tiny$\pm$ 2.4	&\textbf{37.8} \tiny$\pm$ 0.8 \\

\toprule
\multicolumn{1}{c}{} & \multicolumn{3}{c}{Amazon}  & \multicolumn{3}{c}{Amazon (Mid) + Wiki}\\ 
\midrule
\multicolumn{1}{c}{} & 
\multicolumn{1}{c}{1K} & \multicolumn{1}{c}{10K} & 
\multicolumn{1}{c}{} & \multicolumn{1}{c}{} & 
\multicolumn{1}{c}{1K}  & \multicolumn{1}{c}{10K}    
\\
\cmidrule(r){2-4} \cmidrule(r){6-7} 

\lastf{}	&26.6 \tiny$\pm$ 4.4	&54.0 \tiny$\pm$ 2.6	
		&& 	&5.6 \tiny$\pm$ 0.4		&37.9 \tiny$\pm$ 0.9 \\
\meanout{}	&29.4 \tiny$\pm$ 4.0	&54.4 \tiny$\pm$ 2.6	
		&& 	&10.8 \tiny$\pm$ 1.9	&46.5 \tiny$\pm$ 0.5 \\
\maxout{}	&33.5 \tiny$\pm$ 4.5	&55.9 \tiny$\pm$ 2.0	
		&& 	&10.6 \tiny$\pm$ 1.8	&47.0 \tiny$\pm$ 0.9 \\
\att{}		&36.4 \tiny$\pm$ 3.7	&55.6 \tiny$\pm$ 0.6	
		&& 	&17.4 \tiny$\pm$ 3.2	&\textbf{49.7} \tiny$\pm$ 0.3 \\
\attmax{}	&\textbf{37.4} \tiny$\pm$ 3.8	&\textbf{56.2} \tiny$\pm$ 0.8	
		&& 	&\textbf{17.8} \tiny$\pm$ 4.6	&\textbf{49.7} \tiny$\pm$ 0.5 \\

\bottomrule
\end{tabular}}
  \caption{Mean test accuracy ($\pm$ std) (in \%) on standard, `mid' settings across 5 random seeds on Yahoo, Amazon datasets with \textbf{short} sentences ($<100$ words).}
  \label{table:wiki-short-mini}
\end{table}

%% file: sections/NWI.tex
\subsection{Fine-grained Positional Biases}
\label{subsec:fine-grained}
\emph{How does the position of a word affect its contribution to a model's prediction?}
\begin{figure}[htb]
	\centering
	\includegraphics[width=\columnwidth]{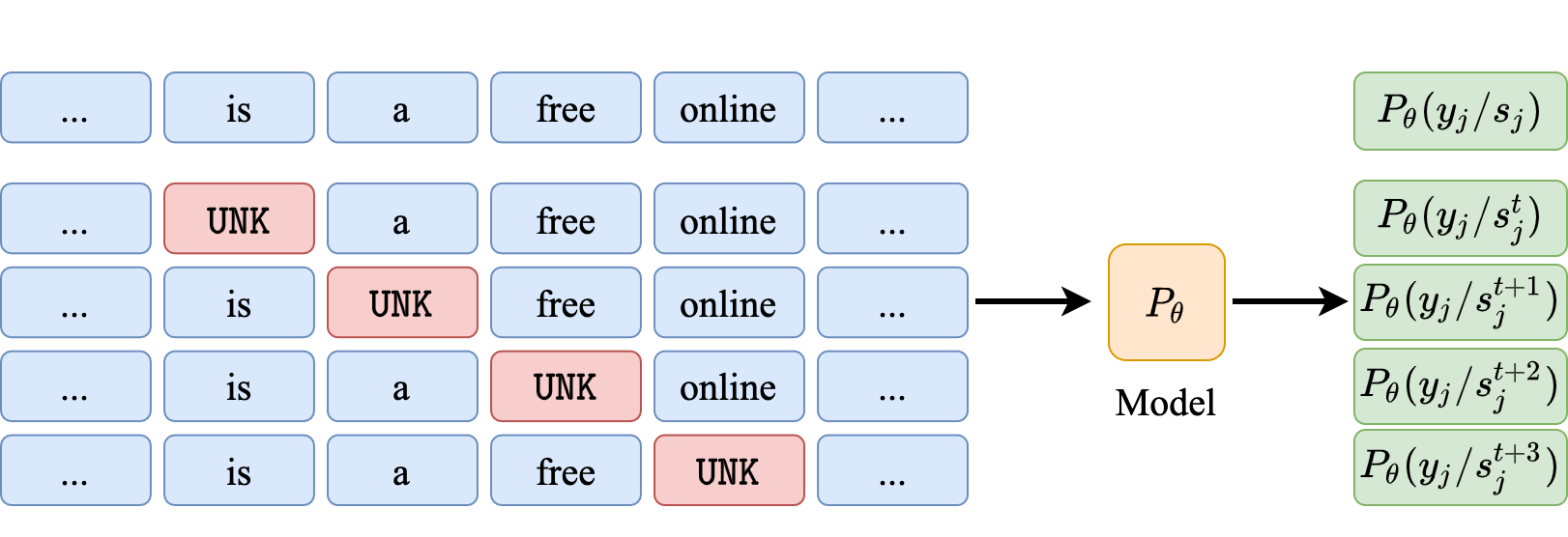} 
	\caption{Explaining NWI evaluation.}
	\label{fig:NWI_explain}
\end{figure}
\begin{figure*}[htb]
\centering
    \includegraphics[width=0.6\textwidth]{Figures/Wiki_Attack/legend.pdf}\\
    \subcaptionbox{Standard \label{fig:YAHOO:NWI_none}}{\includegraphics[width=0.26\textwidth]{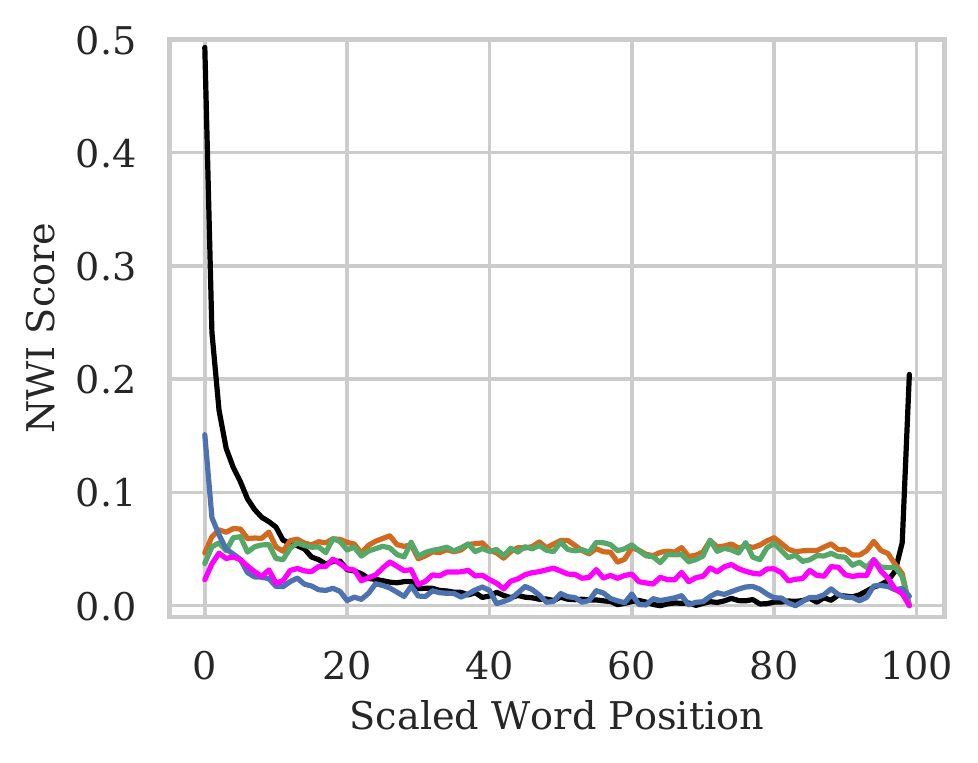}}
	\hfill
	\subcaptionbox{Left \label{fig:YAHOO:NWI_left}}{\includegraphics[width=0.24\textwidth]{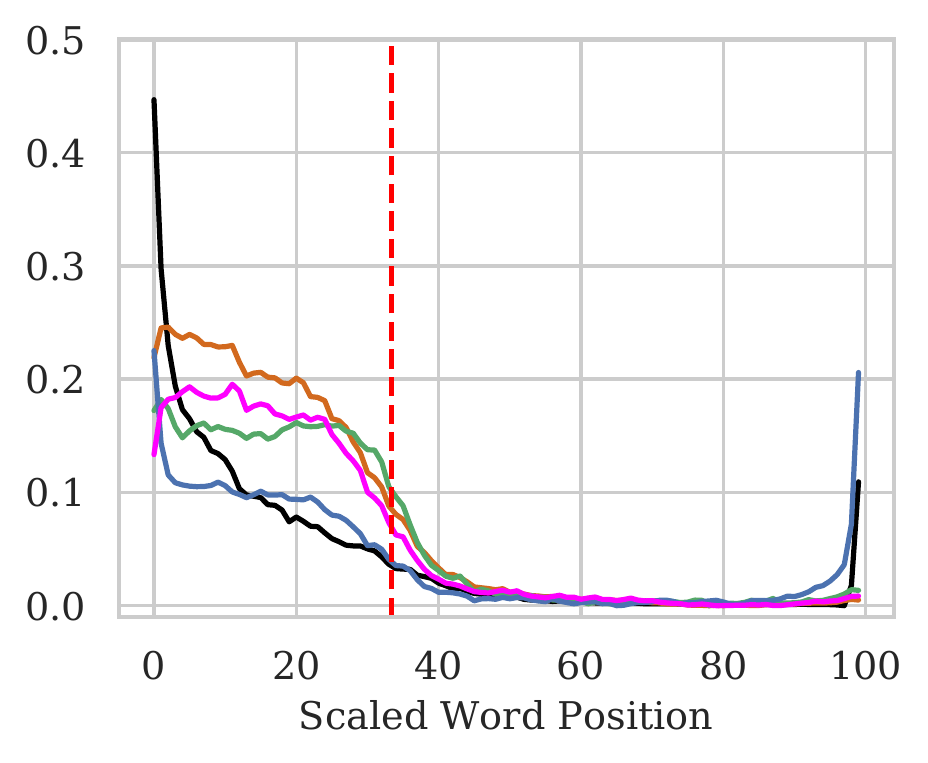}}
	\hfill
	\subcaptionbox{Mid \label{fig:YAHOO:NWI_mid}}{\includegraphics[width=0.24\textwidth]{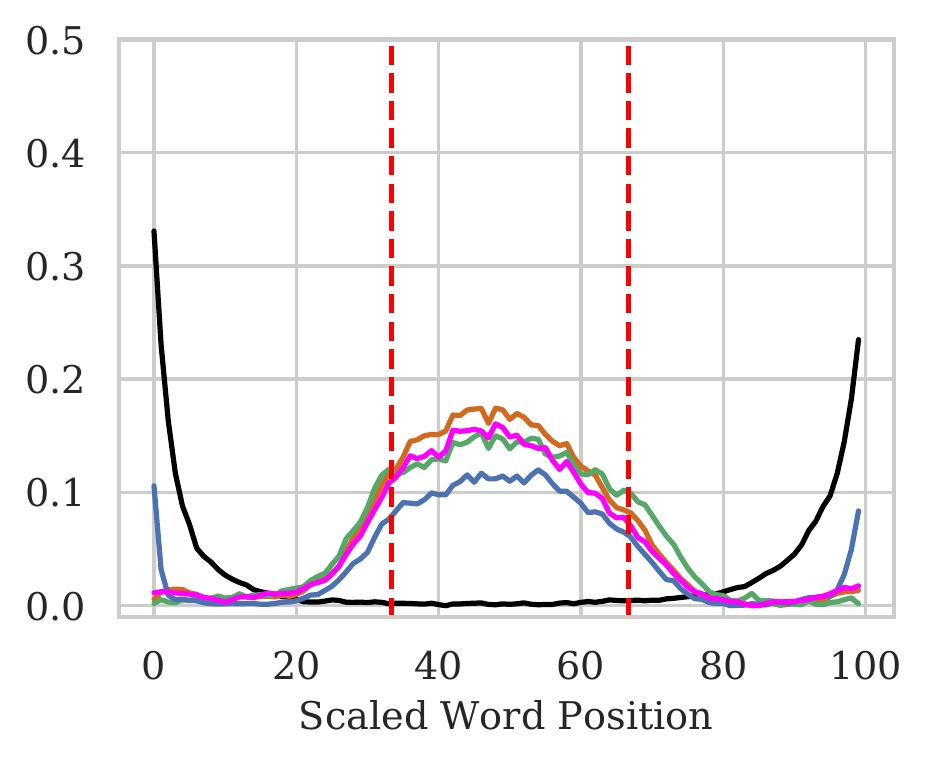}}
	\hfill
	\subcaptionbox{Mid-Short \label{fig:NWI_short_partial}}{\includegraphics[width=0.24\textwidth]{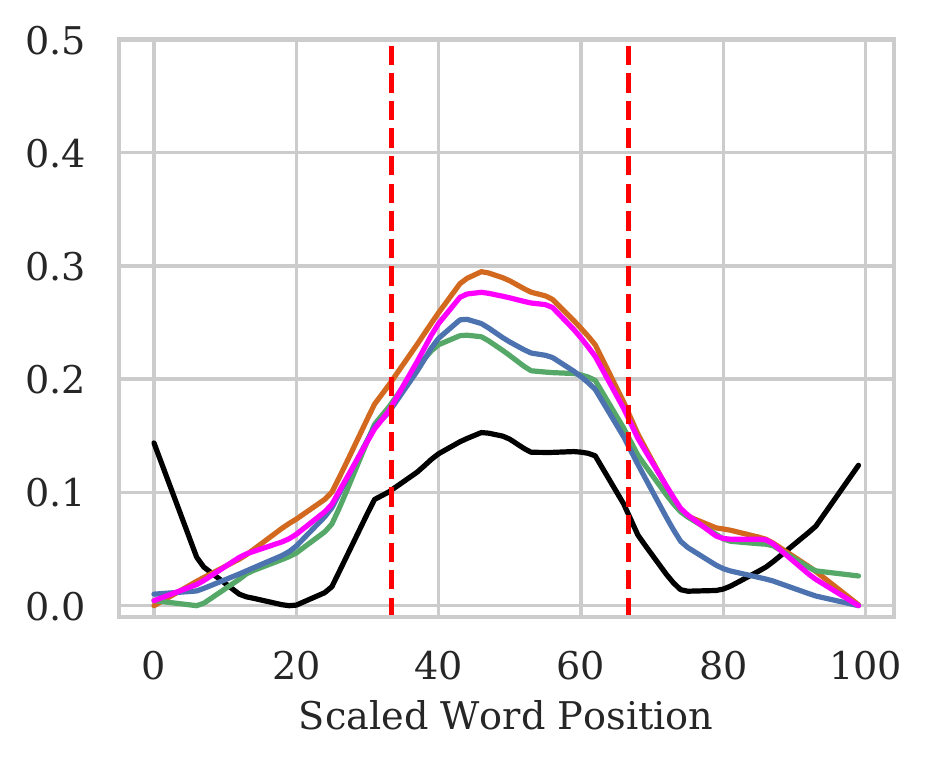}}
	\hfill
\caption{Normalized Word Importance w.r.t. word position averaged over examples of length between $400$-$500$ on the Yahoo ($25$K) dataset in (a,b,c) using $k = 5$; and NWI for examples of length between $50$-$60$ on the Yahoo Short (25K) dataset in (d) with $k. = 3$. Results shown for `standard', `left' \& `mid' training settings described in \S~\ref{subsec:coarse-grained}. 
The vertical red line represents a separator between relevant and irrelevant information (by construction). 
}
\label{fig:NWI:Yahoo-25K}
\end{figure*}
\paragraph{Experimental Setup:} We aim to achieve a fine-grained understanding of model biases w.r.t. each word position, as opposed to evaluating the same at a coarse level (between left, mid and right) as in the previous experiment (\S~\ref{subsec:coarse-grained}). 
To this end, we define Normalized Word Importance (NWI), a 
metric to determine the per-position importance of words as attributed by the model. 
It measures the importance of a particular word (or a set of words) on a model's prediction by calculating the change in the model's confidence in the prediction after replacing it with \texttt{UNK}.
(Figure~\ref{fig:NWI_explain}). The evaluation is further extended by removing a sequence of $k$ consecutive words to get a smoother metric. %
The metric is adapted from past efforts to assign word importance, with some differences~\cite{khandelwal&al18,verwimp&al18,jain2019attention}.\footnote{Unlike our metric, ~\citet{khandelwal&al18} remove all words beyond a certain context, and thus capture how important are \emph{all} the removed words, and not one particular word. ~\citet{jain2019attention}, in their leave-one-out approach, delete a given word rather than replacing it with \texttt{UNK}, thus shifting positions of words by one.} We provide a complete description of the algorithm to compute NWI in  Appendix~\ref{app:fine-grained-assessment}, along with further evaluation on IMDb and Amazon datasets.%
\paragraph{Results:} 
The results from this experiment are presented in Figure~\ref{fig:NWI:Yahoo-25K} (on the Yahoo dataset). The NWI for architectures with pooling indicate no bias w.r.t. word position, however for \lastf{} there exists a clear bias towards the extreme words on either ends (c.f. Figure~\ref{fig:YAHOO:NWI_none}). %
The word importance plots in Figure~\ref{fig:YAHOO:NWI_left} \& \ref{fig:YAHOO:NWI_mid} demonstrate how pooling is able to \emph{learn} to disambiguate between words that are important for sentence classification significantly better as opposed to \lastf{}. There is a clear peak in the middle in case of `mid' setting, and on the left in case of `left' setting for all the pooling architectures. \lastf{} is unable to respond to middle words in Figure~\ref{fig:YAHOO:NWI_mid}. However, they show reasonably higher importance to the left tokens in Figure~\ref{fig:YAHOO:NWI_left} which is justified by their good performance in the `left' experimental setting in Table~\ref{table:wiki-summary}. 
Results for NWI evaluation on all datasets and modified settings (left, mid and right) are available in Appendix~\ref{app:fine-grained-assessment}, and are consistent with the representative graphs in Figure~\ref{fig:NWI:Yahoo-25K}. We also perform such an analysis on models that are trained on datasets with shorter sentences. 
Interestingly, the NWI analysis for the Yahoo short dataset in
Figure~\ref{fig:NWI_short_partial}
shows that while \lastf{} can better respond to middle words for shorter sentences, it still remains heavily biased towards the ends. We detail these findings 
in Appendix~\ref{app:para-NWI-short}

%% file: sections/discussion.tex
Through detailed analysis we identify \emph{why} and \emph{when} pooling representations are beneficial in RNNs.
While some of the results pertaining to gradient propagation in pooling-based RNNs may be obvious in hindsight, we note that this is the first work to systematically and explicitly analyze the phenomenon.
\begin{enumerate}
\item
We attribute the performance benefits of pooling techniques to their learning ability (\emph{pooling mitigates the problem of vanishing gradients}), and positional invariance (\emph{pooling eliminates positional biases}). 
Our findings suggest that pooling offers large gains when training examples are few and long, or when salient words lie in the middle of the sequence. 
    \item In \S\ref{sec:gradients}, we observe that gradients in \last{} vanish only in initial iterations and recover slowly upon further training. We link the observation to training saturation 
    and provide insights as to why BiLSTMs fail in low-resource setups but pooled architectures do not. 
    \item We show that BiLSTMs suffer from positional biases even when sentence lengths are as short as 30 words (Figure~\ref{fig:NWI_short_partial}). 
     
    \item We note that pooling makes models significantly more robust to insertions of random words on either end of the input \emph{regardless} of the amount of training data (Figures~\ref{fig:IMDB:wiki-attack-5K}, \ref{fig:Amazon:wiki-attack-5K}, \ref{fig:IMDb_last:wiki-attack}).
    \item Lastly, we introduce a novel pooling technique (max-attention) that combines the benefits of max-pooling and attention and achieves superior performance on 80\% of our tasks.
\end{enumerate}

\noindent Most of our insights are derived for sequence classification tasks using RNNs.
While our proposed pooling method and analyses are broadly applicable, it remains a part of the future work to evaluate its impact on other tasks and architectures.

%% file: sections/appendix.tex
\section{Equations for Recurrent Networks}
In this section, we provide a mathematical formulation of the equations governing 
LSTMs and basic RNNs.

\subsection{Basic RNN}
Recurrent Neural Networks use a series of input sequence 
$x_{t}$ and pass it sequentially over a network of hidden states 
where each each hidden state leads to the next. Mathematically, this is given by: 

\begin{equation*}
	\begin{split}		
		h_{t} &= \sigma(Ux_{t} + Wh_{t-1} + b) \\
		y_{t} &= \text{softmax}(Vh_{t} + c)
	\end{split}
\end{equation*}
where $x_{t}$ refers to the input sequence at time step t, and $W$, $U$, $V$ are weights for the RNN cell, and $\sigma$ is a non-linearity of choice.

\subsection{LSTM}
\label{app:LSTM_eqns}

The forward propagation of information in a basic LSTM are governed by the following equations:

\begin{equation*}
	\begin{split}	
		i_{t} &= \sigma(W_{ii}x_{t} +b_{ii} +W_{hi}h_{t-1} +b_{hi} )\\
		f_{t} &= \sigma(W_{if}x_{t} +b_{if} +W_{hf}h_{t-1} +b_{hf} )\\
		g_{t} &= \tanh(W_{ig}x_{t} +b_{ig} +W_{hg}h_{t-1} +b_{hg} )\\
		o_{t} &=  \sigma(W_{io}x_{t} +b_{io} +W_{ho}h_{t-1} +b_{ho} )\\
		c_{t} &= f_{t}*c_{t-1}  +i_{t}*g_{t} \\
		h_{t} &= o_{t}*\tanh(c_{t})
	\end{split}
\end{equation*}
\\
where at time $t$, $h_{t}$ is the hidden state, $c_{t}$ is the cell state, $x_{t}$ is the input, and $i_{t}$, $f_{t}$, $g_{t}$, $o_{t}$ are the input, forget, cell, and output gates, respectively. $\sigma$ is the sigmoid function, and * is the Hadamard product.

\subsection{\meanout{}}
\label{app:Mean_eqns}
For a mean-pooled LSTM, while the forward propagation remains the same as \last{}, the output embedding is given by: \\
\vspace{-0.3cm}
$$s_{emb}^{i} = \frac{\sum_{t \in (1,n)}h_{t}^{i}}{n} $$
where $h_{t}^{i}$ represents the $i^{th}$ dimension of the hidden state at time step = $t$, and $s_{emb}$ represents the final output embedding returned by the recurrent structure. This implies that during backpropagation we find a direct influence of the $t^{th}$ hidden state as:
\vspace{-0.3cm}
$$
\frac{\partial s_{emb}^{i}}{\partial h_{t}^{i}}=
\frac{\sum_{k \in (1,n)}\frac{\partial h_{k}^{i}}{\partial h_{t}^{i}}}{n}
$$
\begin{figure*}[h!]
        \begin{center}
            \includegraphics[width=0.8\textwidth]{Figures/Gradients/ratio_legend.png}\\    
        \end{center}
        \subcaptionbox{\att\label{fig:attvec_ratios}}{\includegraphics[width=0.33\textwidth]{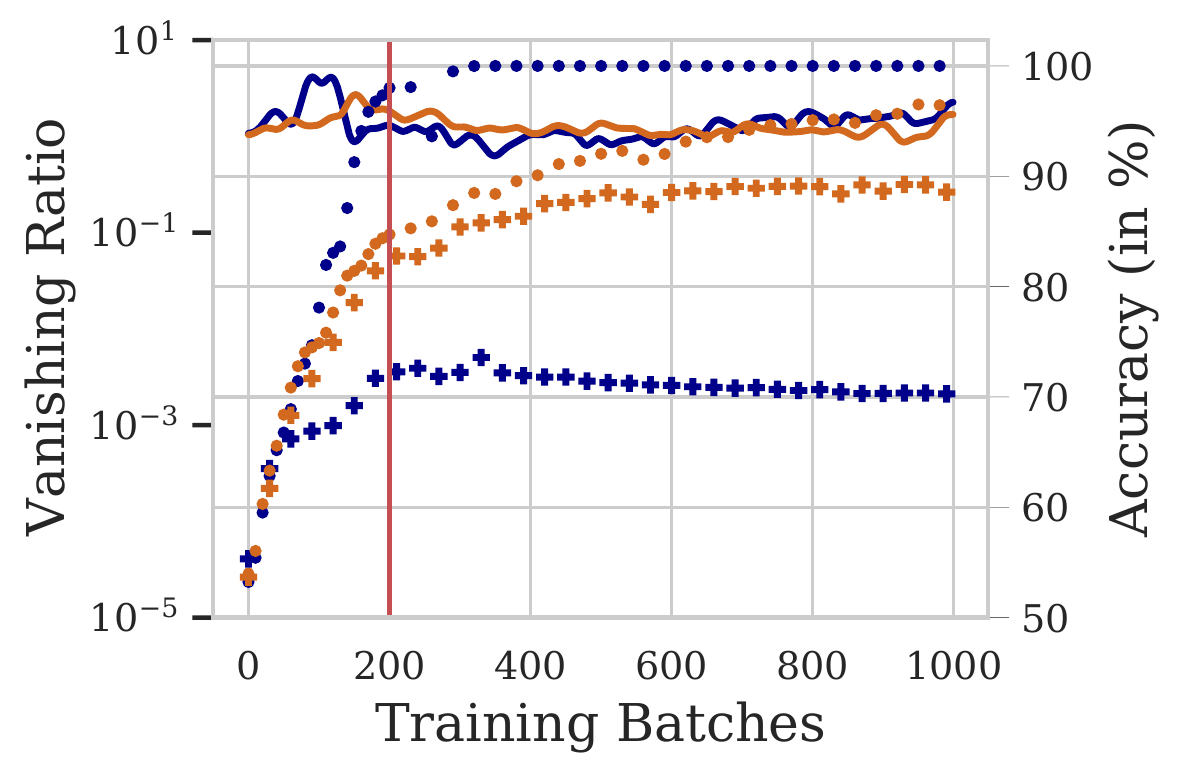}}
        \subcaptionbox{\maxout\label{fig:max_ratios}}{\includegraphics[width=0.33\textwidth]{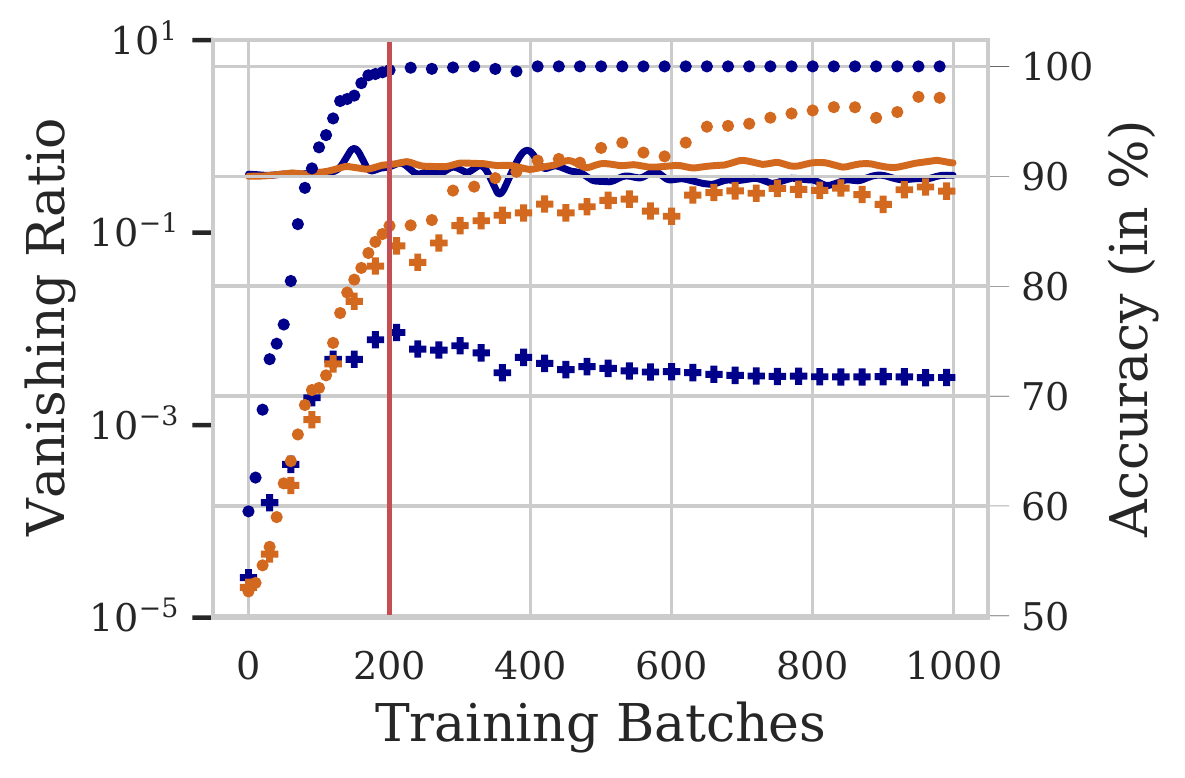}}
        \subcaptionbox{\meanout\label{fig:mean_ratios}}{\includegraphics[width=0.33\textwidth]{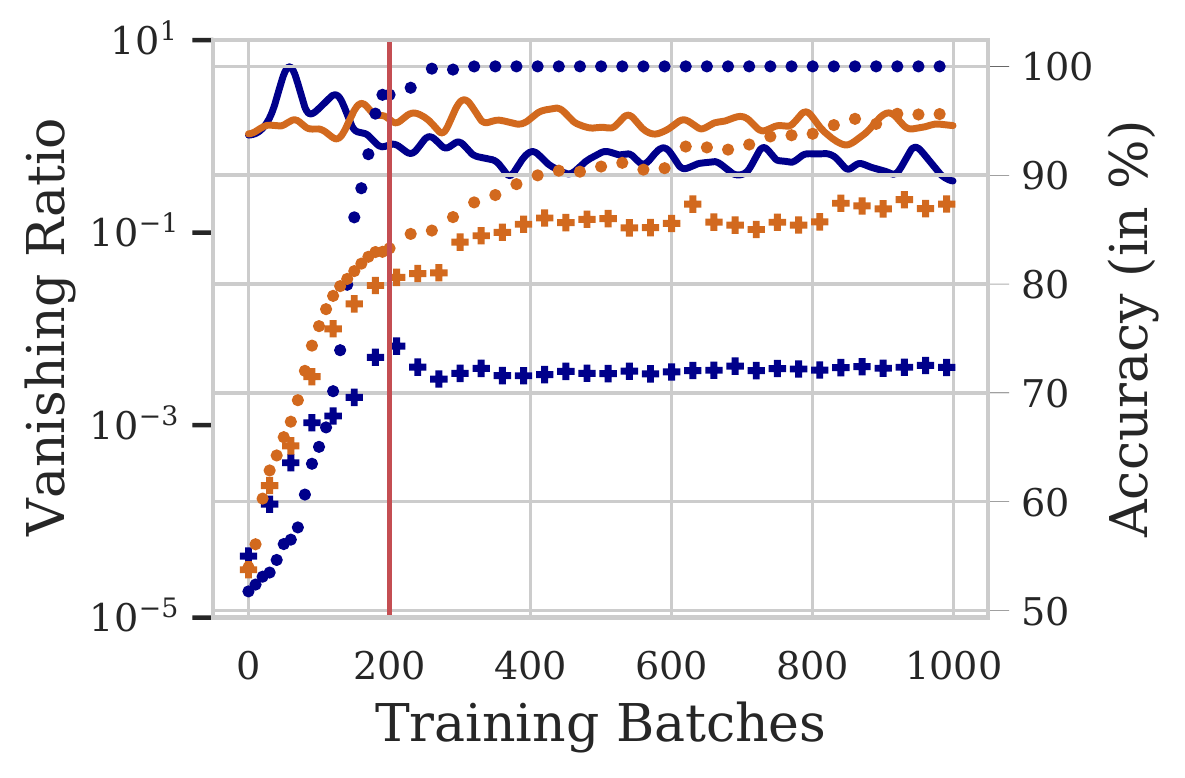}}

        \vspace{-0.4cm}
        \caption{The vanishing ratio $(\|\frac{\partial L}{\partial h_{\text{end}}}\|$$/$$\|\frac{\partial L}{\partial h_{\text{mid}}}\|)$ over training steps for \att{}, \maxout{}, \meanout{} using $1$K, $20$K training examples from the IMDb dataset. The respective training and validation accuracies are also depicted.} 
        \label{fig:app_vanishing_ratios}
        
\end{figure*}
\section{Datasets and Experimental Settings}
\subsection{Dataset Extraction}
\label{app:datasets}
\paragraph{Amazon Reviews} The Amazon Reviews Dataset \cite{ni-etal-2019-justifying} includes reviews (ratings, text, helpfulness votes) and product metadata (descriptions, category etc.) pertaining to products on the Amazon website. We extract the product category and review text corresponding to 2500 reviews from to each of the following 20 classes:
\begin{itemize}
\setlength\itemsep{-0.3em}
\item Automotive
\item Books
\item Clothing Shoes and Jewelry
\item Electronics
\item Movies and TV
\item Arts Crafts and Sewing
\item Toys and Games
\item Pet Supplies
\item Sports and Outdoors
\item Grocery and Gourmet Food
\item CDs and Vinyl
\item Tools and Home Improvement 
\item Software
\item  Office Products
\item  Patio Lawn and Garden
\item  Home and Kitchen
\item  Industrial and Scientific
\item  Luxury Beauty
\item  Musical Instruments
\item  Kindle Store
\end{itemize}
In the standard setting, we ensure that all reviews have lengths between 100 and 500 words.%

\paragraph{IMDb} The IMDb Movie Reviews Dataset \cite{IMDB} is a popular binary sentiment classification task. We take a subset of 20000 reviews that have length greater than 100 words for the purposes of experimentation in this paper. 
\paragraph{Yahoo} Yahoo! Answers \cite{Zhang&al15} has over 1,400,000 question and answer pairs spread across 10 classes. We do not use information such as question, title, date and location for the purpose of classification. As in the case of Amazon reviews, in the standard setting, we ensure that all answers have lengths between 100 and 1000 words, while in the short sentence setting, the maximum answer length in the filtered dataset is 100 words.

\paragraph{Yelp Reviews}: Yelp Reviews \cite{Zhang&al15} is a sentiment analysis task with 5 matching classes. For the purposes of experimentation, we create a subset which is filtered to contain sentences in the range 100 to 1000 tokens. Further, all reviews with a score of 4 or 5 are maked positive, while those with a score of 1 or 2 are marked negative for the binary classification task.

\subsection{Reproducibility}
\label{app:reproducibility}
\paragraph{Computing Infrastructure}
For all the experiments described in the paper, we use a Tesla K40 GPUs supporting a maximum of 10GB of GPU memory. All experiments can be performed on a single GPU. The brief experimentation done on transformer models was done using Tesla V100s that support 32 GB of GPU memory.
\paragraph{Run Time}
The average run-time for each epoch varies linearly with the amount of training data and average sentence length. For the mode with 25K training data in standard setting (sentences with greater than 100 words, and no wikipedia words) the average training time for 1 epoch is under 2 minutes. Further, across all pooling techniques, the run time varies only marginally.
\paragraph{Number of Parameters}
The number of parameters in the model varies with the vocabulary size. We cap the maximum vocabulary size to 25,000 words. However, in the 1K training data setting, the actual vocabulary size is lesser (depending on the training data). The majority of the parameters of the model are accounted for in the model's embedding matrix $=$ (vocabulary size)$\times$(embedding size). The number of parameters for the main LSTM model are around 70,000, with the \att{} model having a few more parameters than other methods due to a learnable query vector.
\paragraph{Validation Scores}
We provide validation results in Table~\ref{table:vanish_table} for the standard setting. However, in interest of brevity, we only detail the test scores in all subsequent tables. Note that we always select the model based on the best validation accuracy during the training process (among all the epochs).
\paragraph{Evaluation Metric}
The evaluation metric used is the model's accuracy on the test set and is reported as an average over 5 different seeds. All the classes are nearly balanced in the datasets chosen, hence standard accuracy metric serves as an accurate indicator.
\paragraph{Hyperparameters search}
An explicit hyperparameter search is not performed for each model in each training setting over all seeds, since the purpose of the paper is not to beat the state of art, but rather to analyze the effect of pooling in recurrent architectures. We do note that, in the manual search performed on the learning rates of $\{1\times10^{-3}, 2\times10^{-3}, 5\times10^{-3} \}$ on the IMDb and Yahoo datasets, we find that for all the pooling and non-pooling methods discussed, models trained on learning rate equal to $2\times10^{-3}$ showed the best validation accuracy. Thus, we use that for all the following results. However, we do perform a hyperparameter search for the best regularization parameters as described in Appendix~\ref{app:regularization}. We keep the embedding dimension and hidden dimension fixed for all experiments.
\begin{figure*}[htb]
	\centering
		\includegraphics[width=0.6\textwidth]{Figures/Wiki_Attack/legend.pdf}\\
		\subcaptionbox{Left \label{fig:AMAZON:Wiki_L}}{\includegraphics[width=0.34\textwidth]{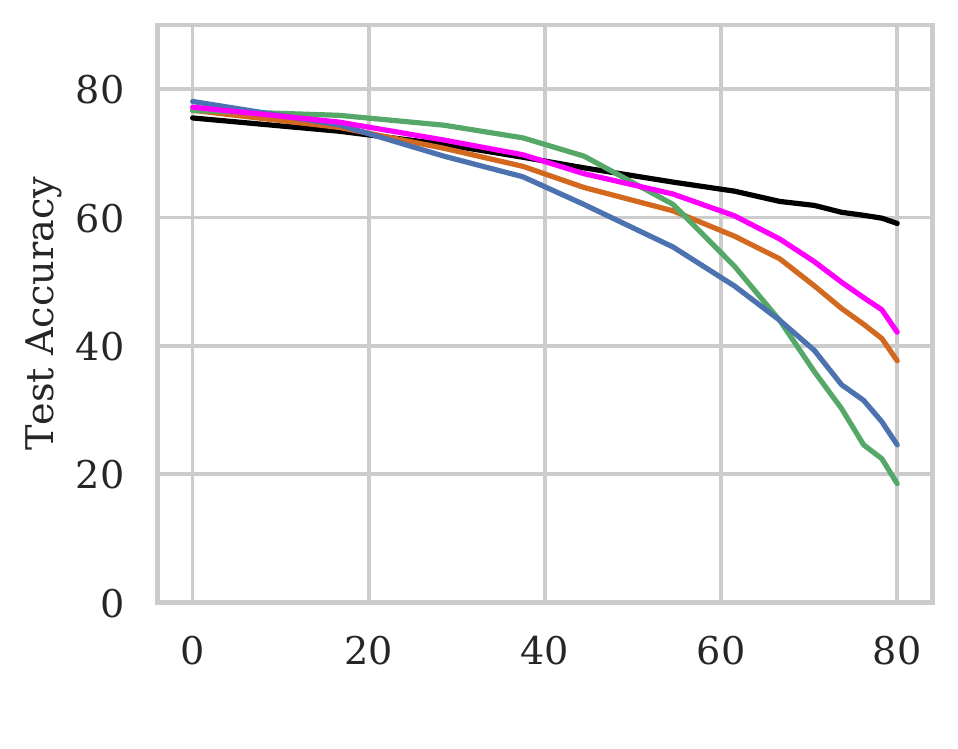}}
		\hfill
		\subcaptionbox{Mid \label{fig:AMAZON:Wiki_M}}{\includegraphics[width=0.32\textwidth]{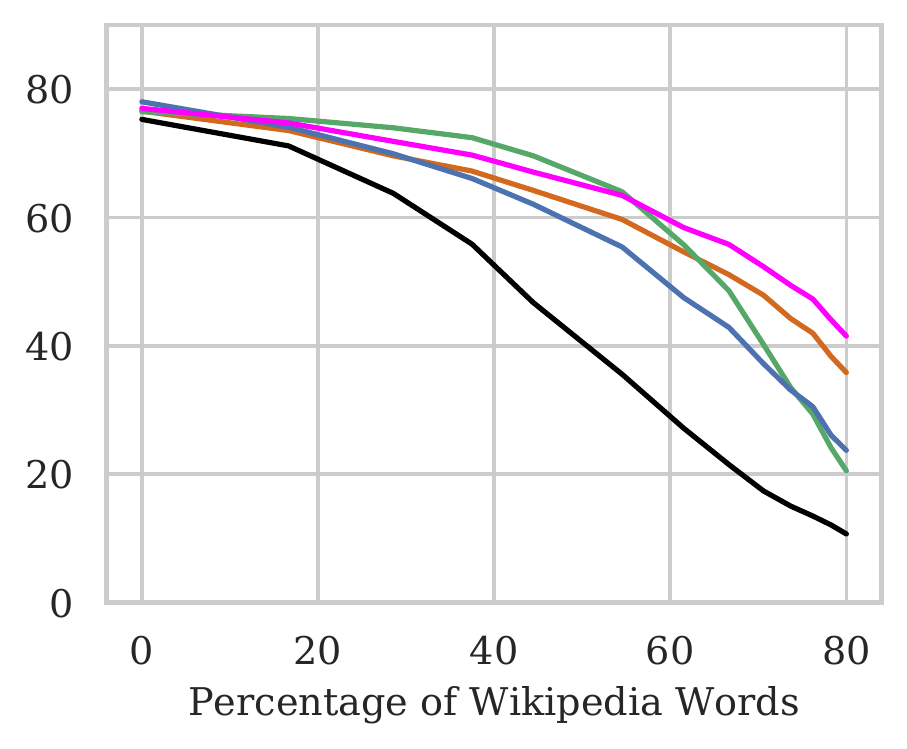}}
		\hfill
		\subcaptionbox{Right \label{fig:AMAZON:Wiki_r}}{\includegraphics[width=0.32\textwidth]{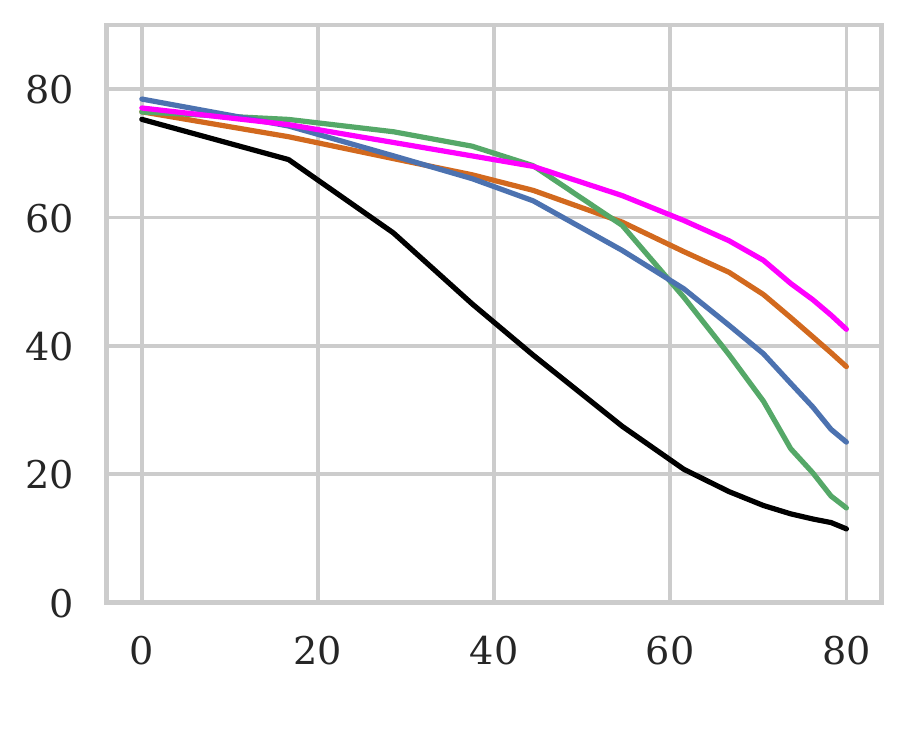}}
		\hfill
		
	\caption{Amazon Dataset (10K setting): Random Wikipedia sentences are appended to the original input paragraphs. Original input is preserved on the (a) left, (b) middle, and (c) right of the new input. Test accuracies are reported by varying the percentage of total Wikipedia words in the new input.}\label{fig:Amazon:wiki-attack-5K}
	
	\end{figure*}
\begin{figure*}[htb]
	\centering
		\includegraphics[width=0.35\textwidth]{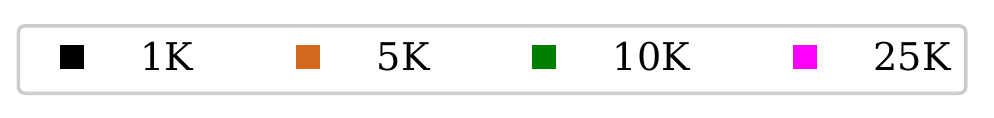}\\
		\subcaptionbox{Left \label{fig:IMDb_last:Wiki_L}}{\includegraphics[width=0.34\textwidth]{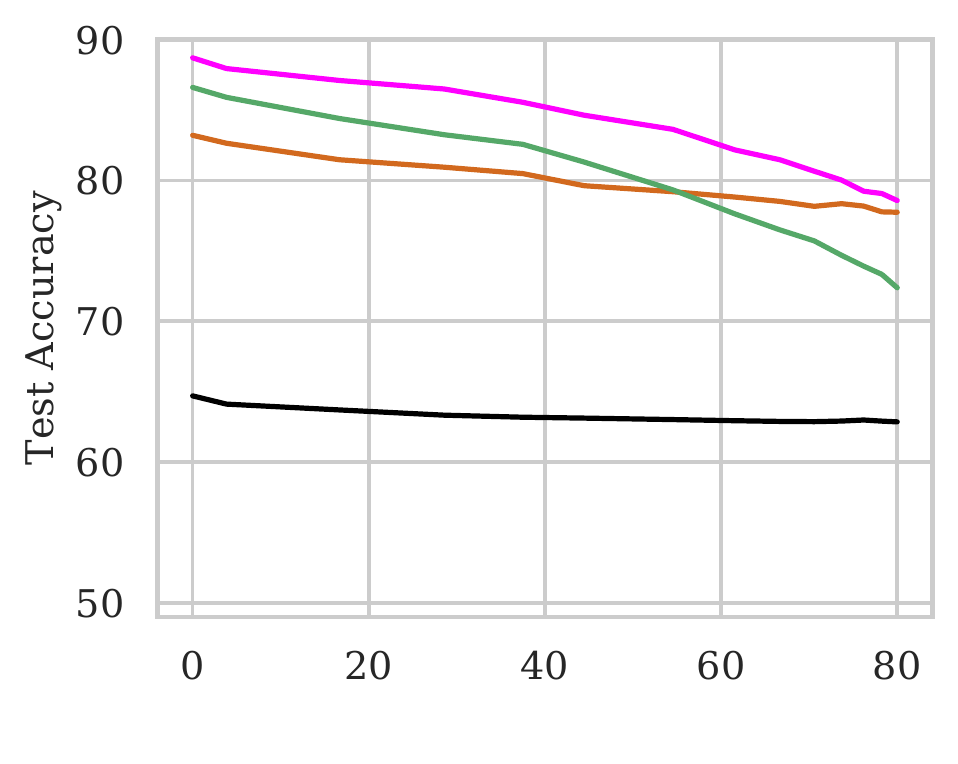}}
		\hfill
		\subcaptionbox{Mid \label{fig:IMDb_last:Wiki_M}}{\includegraphics[width=0.32\textwidth]{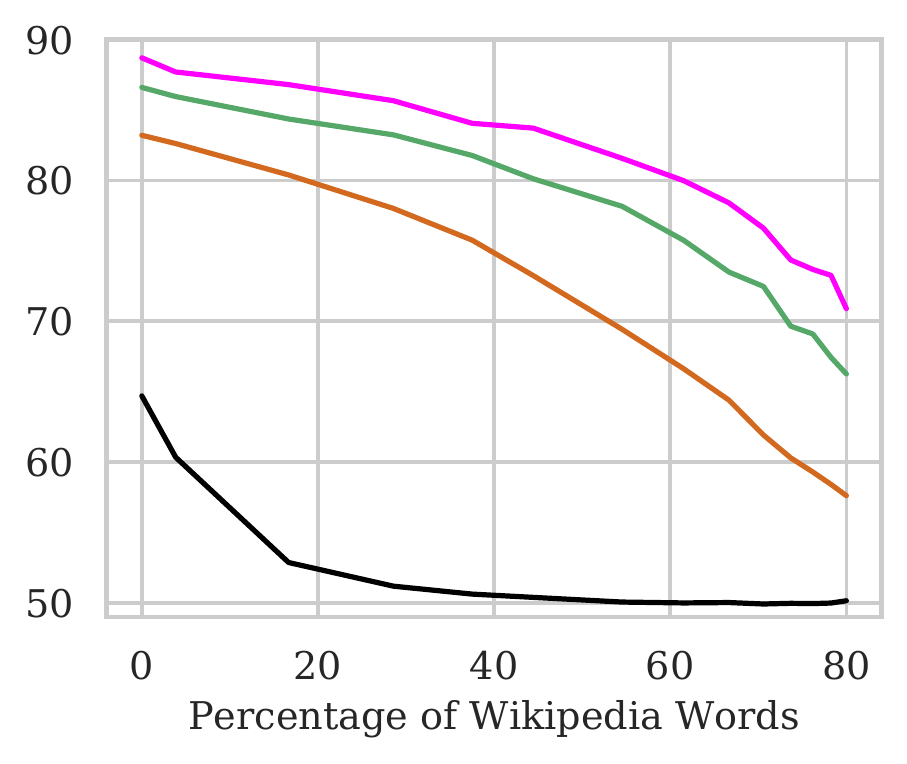}}
		\hfill
		\subcaptionbox{Right \label{fig:IMDb_last:Wiki_r}}{\includegraphics[width=0.32\textwidth]{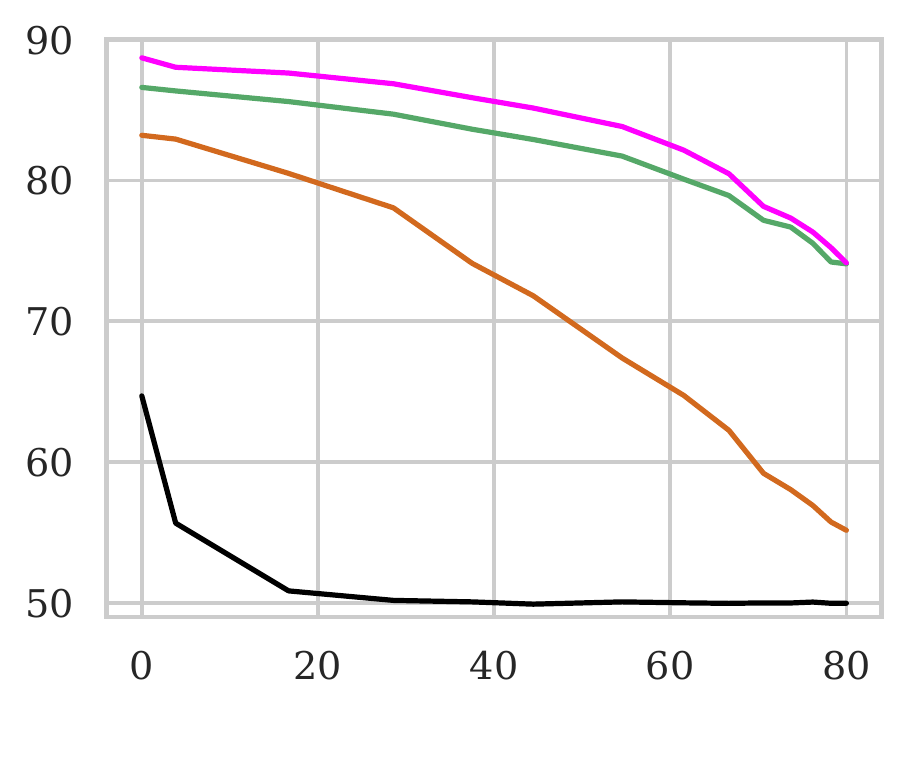}}
		\hfill
		
	\caption{IMDb Dataset (BiLSTM): Random Wikipedia sentences are appended to the original input paragraphs for the standard BiLSTM models trained on 1K, 5K, 10K and 20K examples. Original input is preserved on the (a) left, (b) middle, and (c) right of the new input. Test accuracies are reported by varying the percentage of total Wikipedia words in the new input. BiLSTM is unrepsonsive to any appended tokens as long as the `left' text is preserved in the 1K and 5K setting. But this bias dilutes with more training samples. Given sufficient data (more than 10K unique examples) the effect of appending random words on both ends is more detrimental than that on appending at only one end.}\label{fig:IMDb_last:wiki-attack}
	
\end{figure*}
\section{Gradient Propagation}
\label{app:vanishing}
The plots of the change in vanishing ratios for \att{}, \maxout{} and \meanout{} are shown in Figure~\ref{fig:app_vanishing_ratios}. This completes the representative analysis for \last{} and \attmax{} shown in Figure~\ref{fig:vanishing_ratios}. It can be seen that for all the different pooling types discussed in this paper, the vanishing ratios are small right from the beginning of training. This motivates future research to further formally analyze and discover other learning advantages (apart from vanishing ratios) that distinguish the performance of one pooling technique from the other.

\input{tables/table_yahoo_amazon.tex}
\input{tables/table_IMDB_Yelp.tex}
\section{Evaluating Natural Positional Biases}
\label{app:wiki_attack:full}

In line with our results in \S~\ref{subsec:effect_rand_noise}, we further evaluate 
models trained on the Amazon dataset in the same settings to re-validate our results.
The effect of appending random Wikipedia sentences to input examples on models 
trained on the Amazon dataset can be found in Figure~\ref{fig:Amazon:wiki-attack-5K}. We use 
the model trained on 10K examples to perform this experiment. The graphs show 
similar findings as in Figure~\ref{fig:IMDB:wiki-attack-5K}, and further supports 
the hypothesis that \lastf{} gives a strong emphasis on extreme words when trained 
on standard datasets, which is why its performance significantly deteriorates when 
random Wikipedia sentences are appended on both ends. 

\paragraph{Effect of Amount of Training Data:} Figure~\ref{fig:IMDB:wiki-attack-5K} suggests that \lastf{} is equally responsive to the effect of appending random words to the left or right. However, in case of the Amazon Reviews dataset (Figure~\ref{fig:Amazon:wiki-attack-5K}), we notice that the BiLSTM is more resilient when the text to the left is preserved. This indicates a learning bias, where the BiLSTM pays greater emphasis to outputs of one chain of the bidirectional LSTM. It is interesting to note that on reducing the training data, this bias increases significantly in the case of IMDb dataset as well. 

We hypothesize that such a phenomenon may have resulted due to an artifact of the training process itself, that is, the model is able to find `easily identifiable' important sentiment at the beginning of the reviews during training (speculatively due to the added effects of padding to the right). 
Therefore, given less training data, BiLSTMs prematurely learn to use features from only one of the two LSTM chains and (in this case) the left $\,\to\,$ right chain of the dominates the final prediction.
We confirm from Figure~\ref{fig:IMDb_last:wiki-attack} that with a decrease in training data (such as in the 1K IMDb data setting), the bias towards one end substantially increases, that is,  \lastf{} is extremely insensitive to random sentence addition, as long as the left end is preserved.

\paragraph{Practical Implications} We observe that \meanout{} and \lastf{} can be susceptible to \emph{changes in test-time data distribution}. This questions the use of such models in real word settings. We speculate that paying equal importance to all hidden states handicaps \meanout{} from being able to distil out important information effectively, while the preceding discussion on the effect of size of training data highlights the possible cause of this occurrence in \lastf{}. We observe that other pooling methods like \attmax{} are able to circumvent this issue as they are only mildly affected by the added Wikipedia sentences.
\section{Training to Skip Unimportant Words}
\label{app:coarse-grained}
We demonstrate in \S~\ref{subsec:coarse-grained} that the ability of \lastf{}, and its different pooling variants, to learn to skip unrelated words can be greatly diminished in challenging datasets especially given less amount of input data. 
In this section, we aim to (a) provide a complete evaluation on all positions of data modification and dataset size settings (including those which were skipped in the main paper for brevity); 
(b) evaluate the same experiment in a setting where input examples are shorter in length.

\subsection{Full Evaluation}
\label{app:wiki-full_eval}
For completeness, we perform the evaluation in \S~\ref{subsec:coarse-grained} on each of \{1K, 2K, 5K, 10K, 25K\} dataset size settings, and also report the results when Wikipedia words are appended on the right, preserving the original input to the left. We report results for the Yahoo and Amazon datasets in Table~\ref{table:wiki-long-yahoo-amazon} and the IMDb and Yelp Reviews datasets in Table~\ref{table:wiki-long-IMDb-Yelp}. It can be noted that the advantages of \attmax{} over other pooling and non-pooling techniques significantly increase in the three Wikipedia settings in each of the tables. This suggests that \attmax{} performs better in more challenging scenarios where the important signals are hidden in the input data.  Further, the performance advantages of \attmax{} are more when amount of training data is less.

\subsection{Short Sentences}
\label{app:wiki-short_sent}
\begin{table}[H]
    \small \centering
    \begin{tabular}{@{}lccccc@{}}
    \toprule
    Dataset & Classes & \makecell{Avg. \\ Length} & \makecell{Max \\ Length} & \makecell{Train \\ Size} & \makecell{Test \\ Size} \\ 
    \midrule
    Yahoo! Answers  & 10      & 30.1       & 95        & 25K        & 25K \\
    Amazon Reviews  & 20       & 29.1       & 100       & 25K        & 12.5K \\
    \bottomrule
    \end{tabular}
    \caption{Corpus statistics for classification tasks (short datasets).}
    \label{table:corpus_statistics_short}
\end{table}

\input{tables/table_short.tex}
For shorter sequences, we reuse two of our text classification tasks: 
(1) \textbf{Yahoo! Answers}; and  
(2) \textbf{Amazon} Reviews. Similar to the setting with long sentences in the main paper, we use only the text and labels, ignoring any auxiliary information (like title or location). 
We select subsets of the datasets with sequences having a length (number of space separated words) less than  $100$. 
A summary of statistics with respect to sentence length and corpus size is given in Table~\ref{table:corpus_statistics_short}.

The results for the performance of the trained models can be found in Table~\ref{table:wiki-short-summary}. 
In the `Mid' setting, we observe that \lastf{} performs significantly better on shorter sequences as opposed to the long sequences. For instance, in case of Amazon Dataset (Mid), under the 25K data setting, the classification accuracy increases from 7.8\% in Table~\ref{table:wiki-long-yahoo-amazon} to 51.5\% in Table~\ref{table:wiki-short-summary}, which is a significant improvement from only doing as well as majority guessing in the former. 
We note that most of the learning issues of \lastf{} in long sentence setting are largely absent 
when sentence lengths are short, with \lastf{} also emerging as the best-performing model in a few cases. This corroborates the effect of gradients vanishing with longer time steps.

\subsection{On using regularization}
\label{app:regularization}
For the experiments in the work, we do not regularize trained LSTMs. This has two analytical advantages (1) we can examine the benefits of pooling without having to account for the the effect of regularization; and (2) training to 100\% accuracy acts as an indicator of training the models \emph{adequately}. However, for validation, we also performed our experiments on the IMDb dataset with 2 different types of regularization schemes, following best practices used in previous works \cite{merity2017regularizing}. We use DropConnect \cite{DropConnect-wan13} \footnote{grid search over mask rate: \{0.1,0.3,0.5\}} and Weight Decay \footnote{grid search over decay value: \{$10^{-3},10^{-4},10^{-6},10^{-8}$\}} for regularization of all the models. We observe that the effect of regularization consistently improves the final accuracies by 1-2\% across the board. However, even after sustained training (up to 50 epochs), \lastf{} still 
suffers from the learning issues outlined in the paper.
The goal of this paper is not to study the effect of various regularization schemes, but to merely understand the effect pooling in improving the performance of \lastf{}. 

\begin{figure*}[!ht]
\vspace{3pt}
\centering
    \includegraphics[width=0.6\textwidth]{Figures/Wiki_Attack/legend.pdf}
    \\
    \subcaptionbox{Standard - IMDb \label{fig:app:IMDB:NWI_none}}{\includegraphics[width=0.34\textwidth]{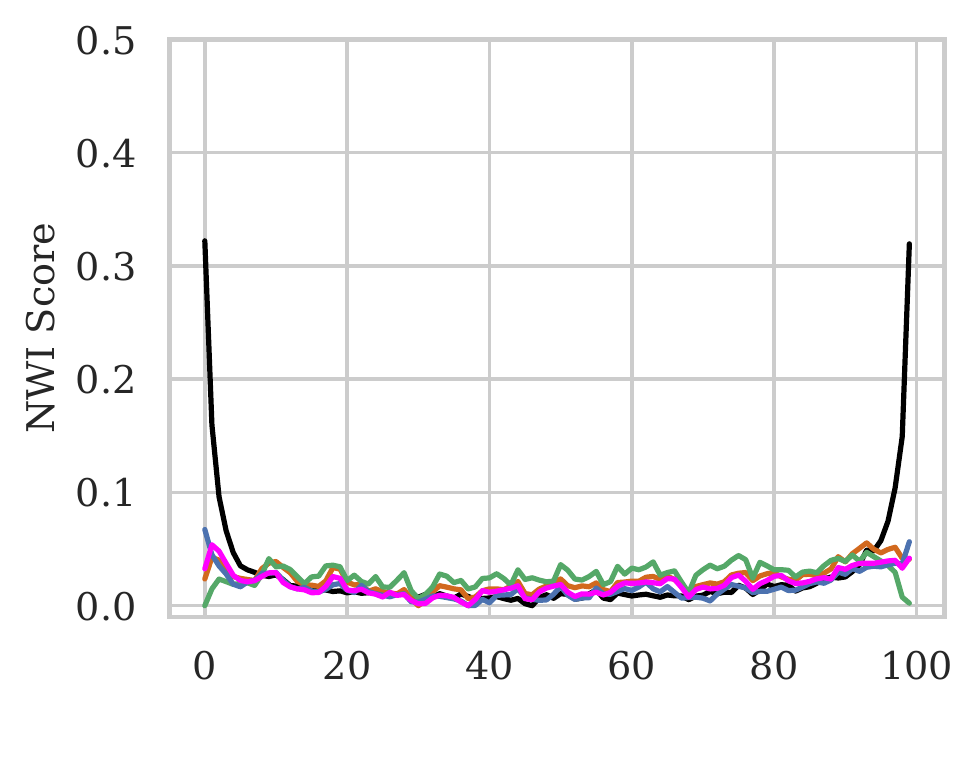}}
	\hfill
	\subcaptionbox{Standard - Yahoo \label{fig:app:Yahoo:NWI_none}}{\includegraphics[width=0.32\textwidth]{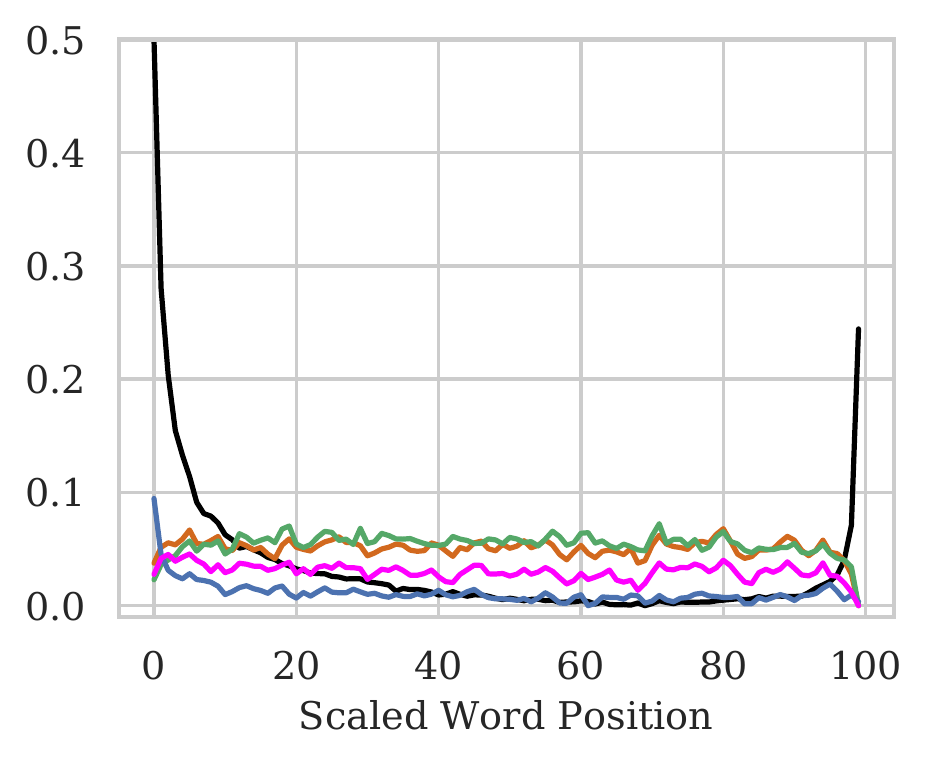}}
	\hfill
	\subcaptionbox{Standard - Amazon \label{fig:app:Amazon:NWI_none}}{\includegraphics[width=0.32\textwidth]{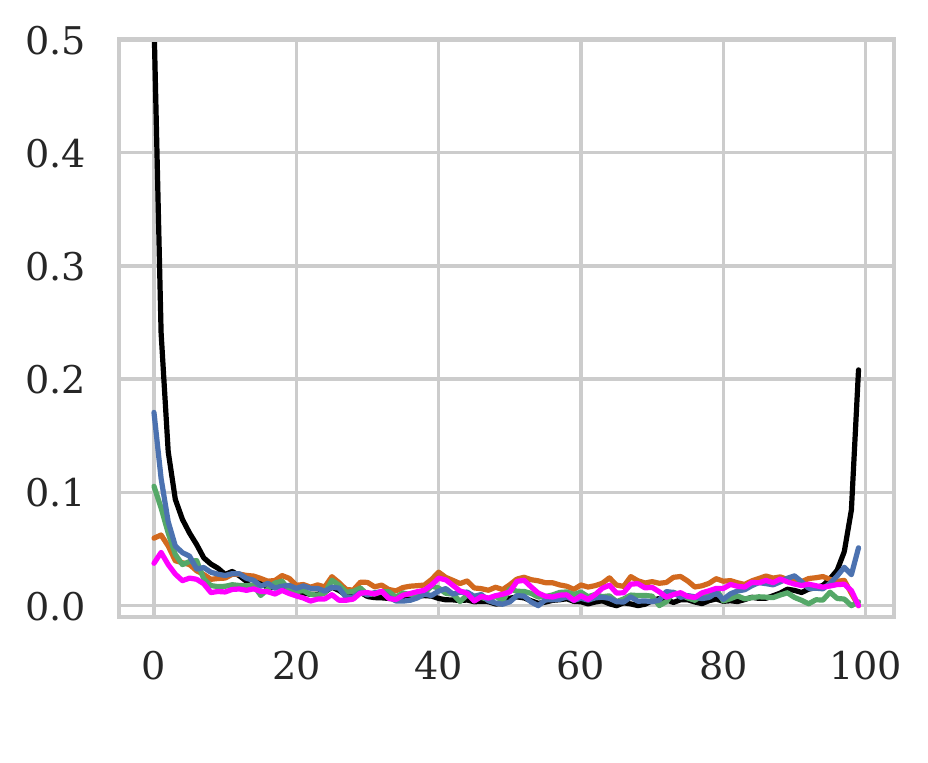}}
	\hfill
	\\
	
	\subcaptionbox{Left - IMDb \label{fig:app:IMDB:NWI_left}}{\includegraphics[width=0.34\textwidth]{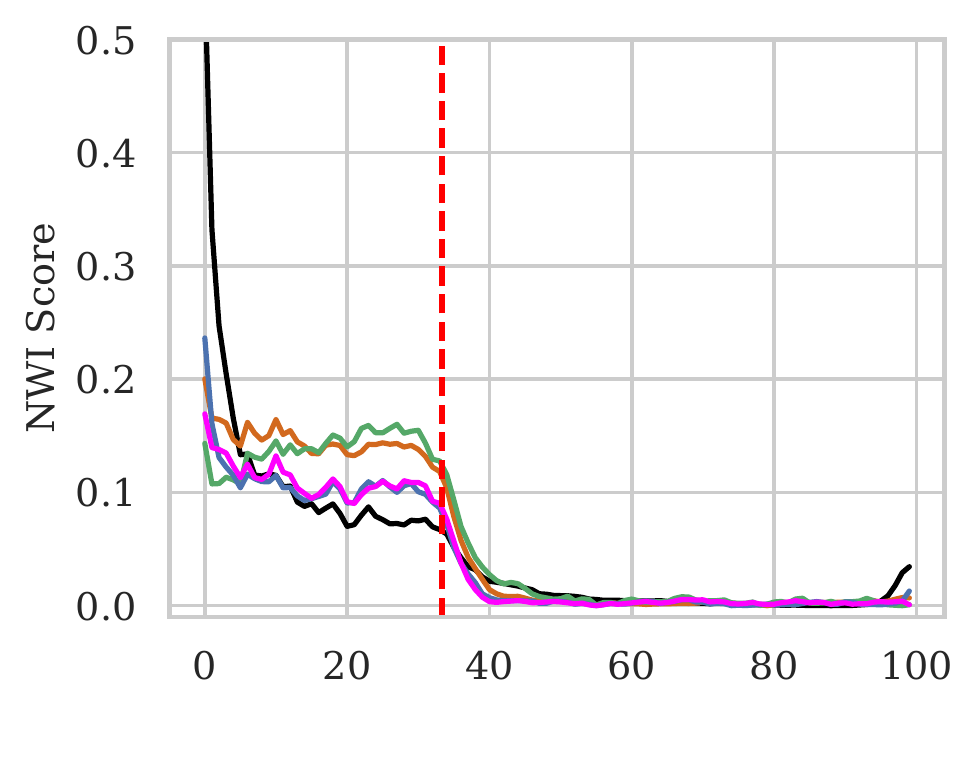}}
	\hfill
	\subcaptionbox{Left - Yahoo \label{fig:app:Yahoo:NWI_left}}{\includegraphics[width=0.32\textwidth]{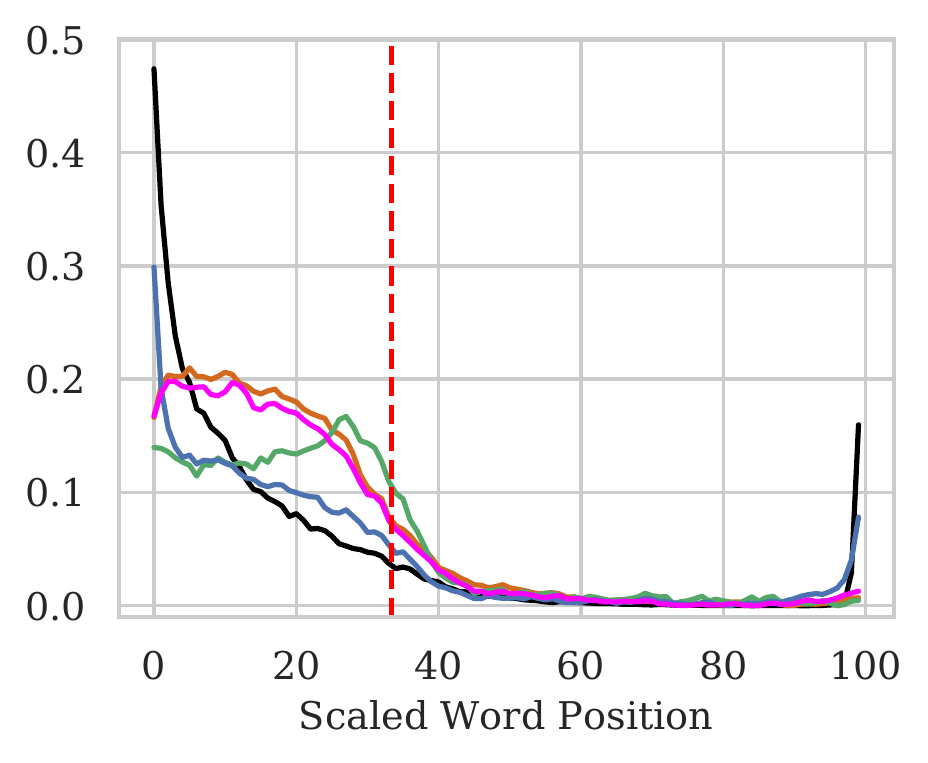}}
	\hfill
	\subcaptionbox{Left - Amazon \label{fig:app:Amazon:NWI_left}}{\includegraphics[width=0.32\textwidth]{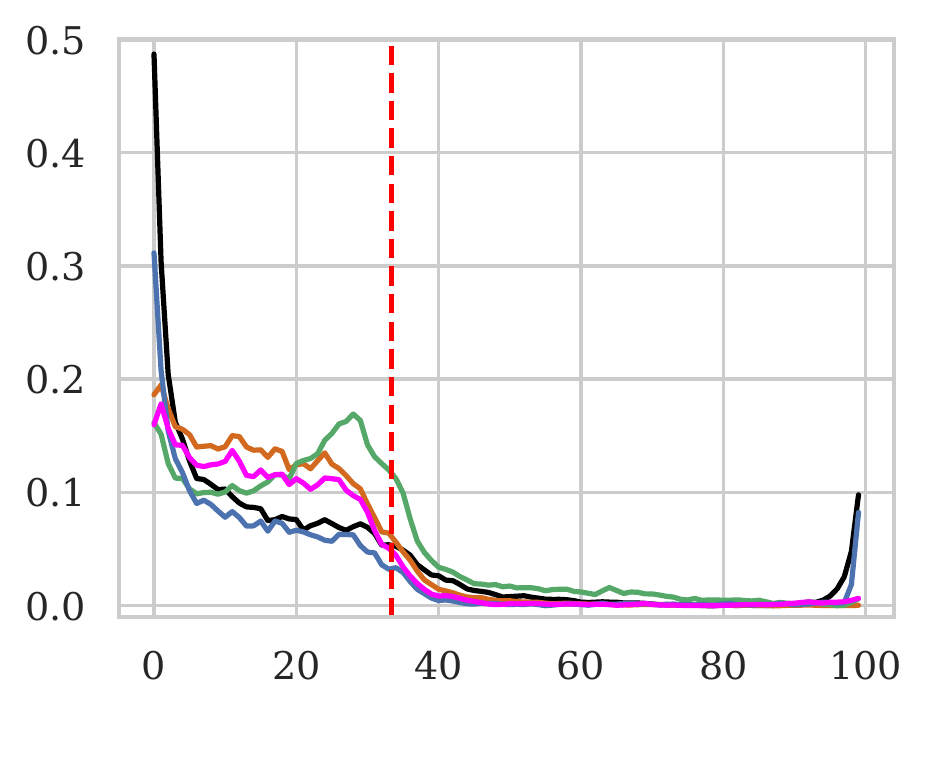}}
	\hfill
	\\
	
	\subcaptionbox{Mid - IMDb \label{fig:app:IMDB:NWI_mid}}{\includegraphics[width=0.34\textwidth]{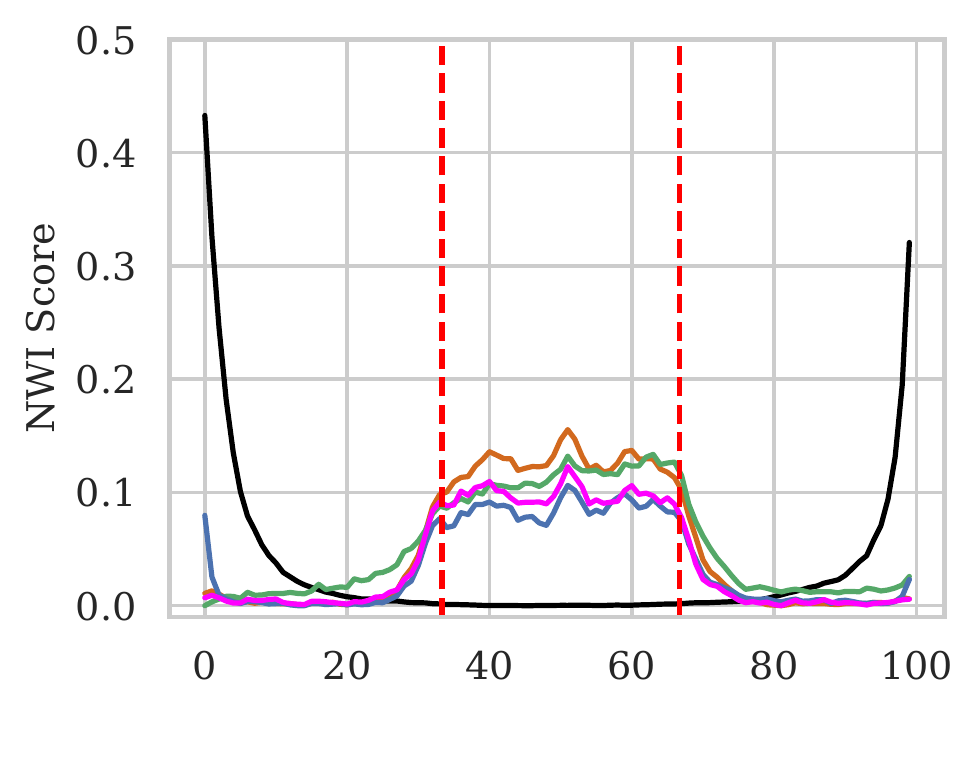}}
	\hfill
	\subcaptionbox{Mid - Yahoo \label{fig:app:Yahoo:NWI_mid}}{\includegraphics[width=0.32\textwidth]{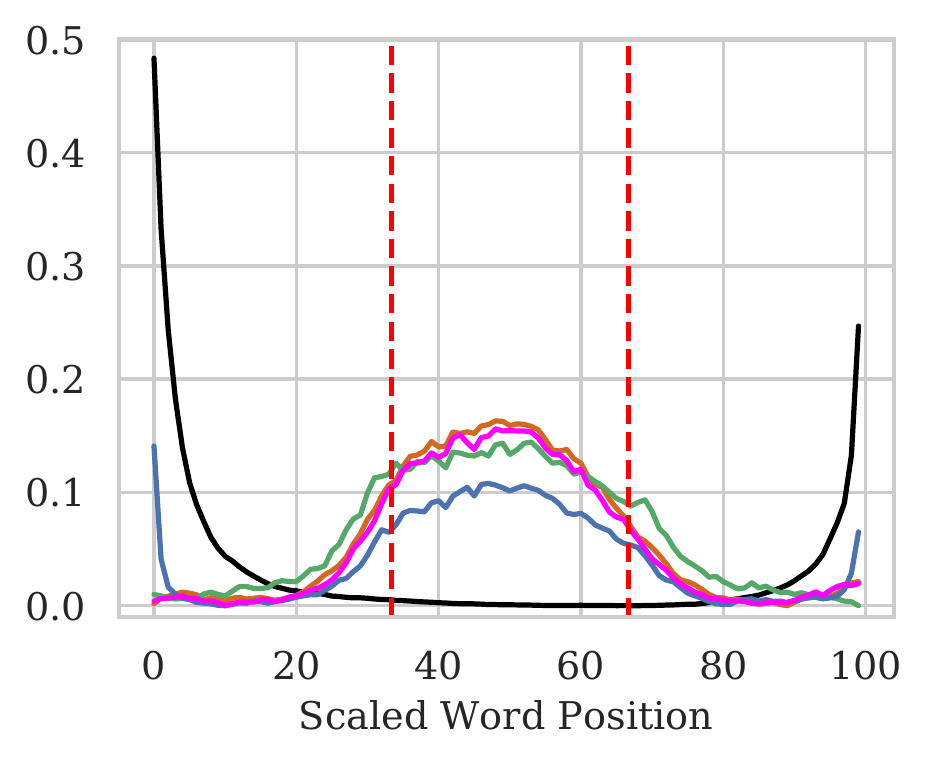}}
	\hfill
	\subcaptionbox{Mid - Amazon \label{fig:app:Amazon:NWI_mid}}{\includegraphics[width=0.32\textwidth]{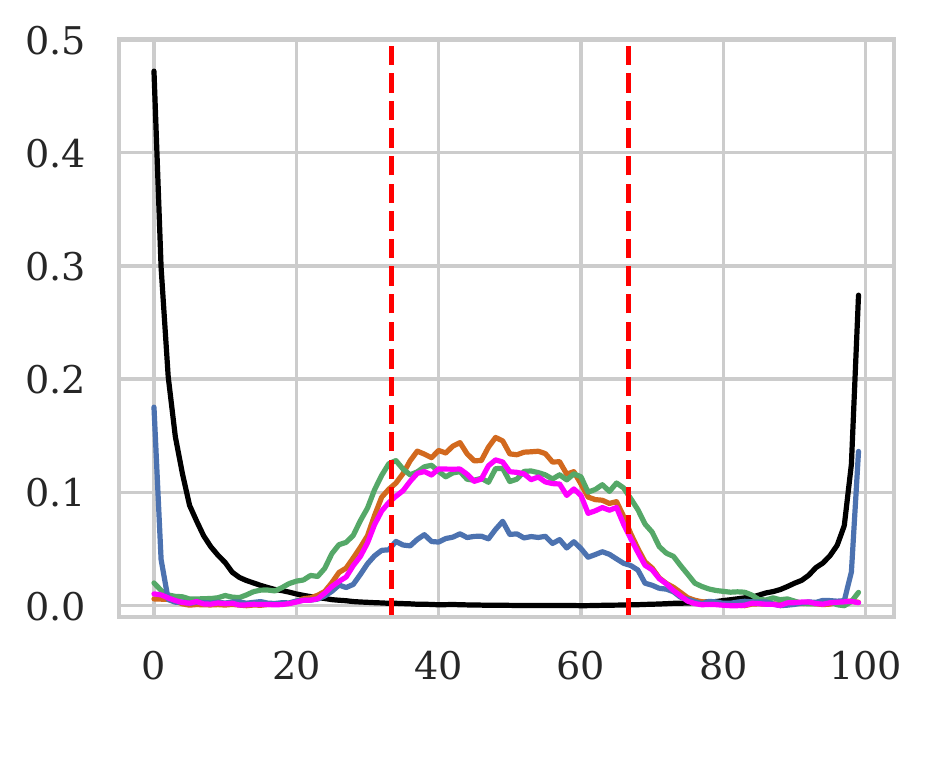}}
	\hfill
	\\
	
	\subcaptionbox{Right - IMDb \label{fig:app:IMDB:NWI_right}}{\includegraphics[width=0.34\textwidth]{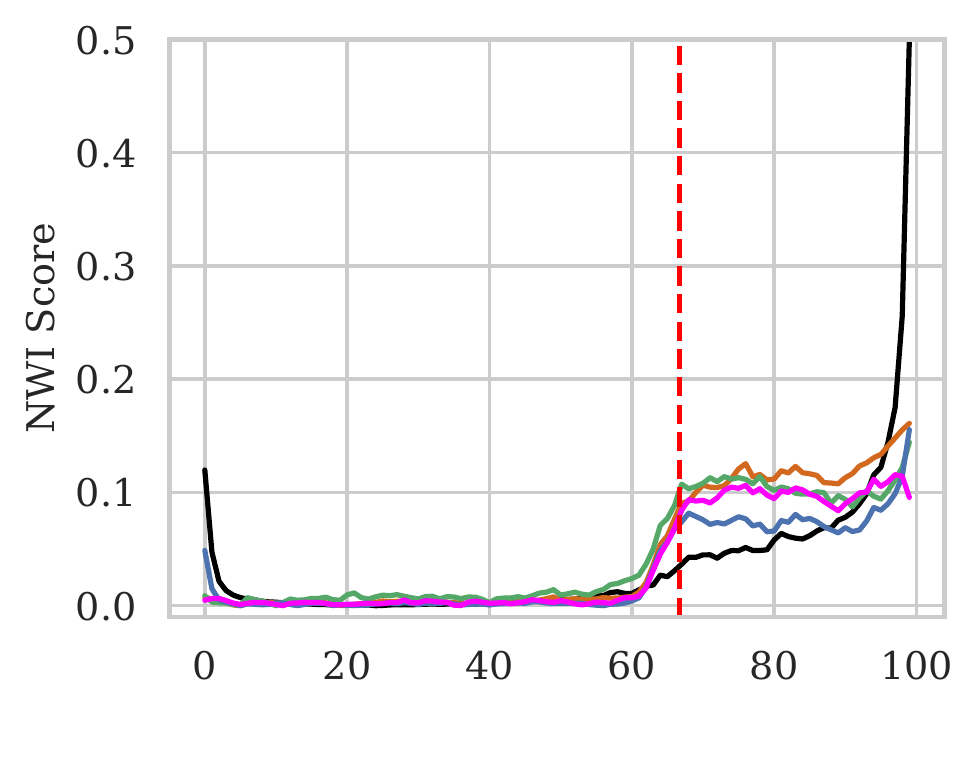}}
	\hfill
	\subcaptionbox{Right - Yahoo \label{fig:app:Yahoo:NWI_right}}{\includegraphics[width=0.32\textwidth]{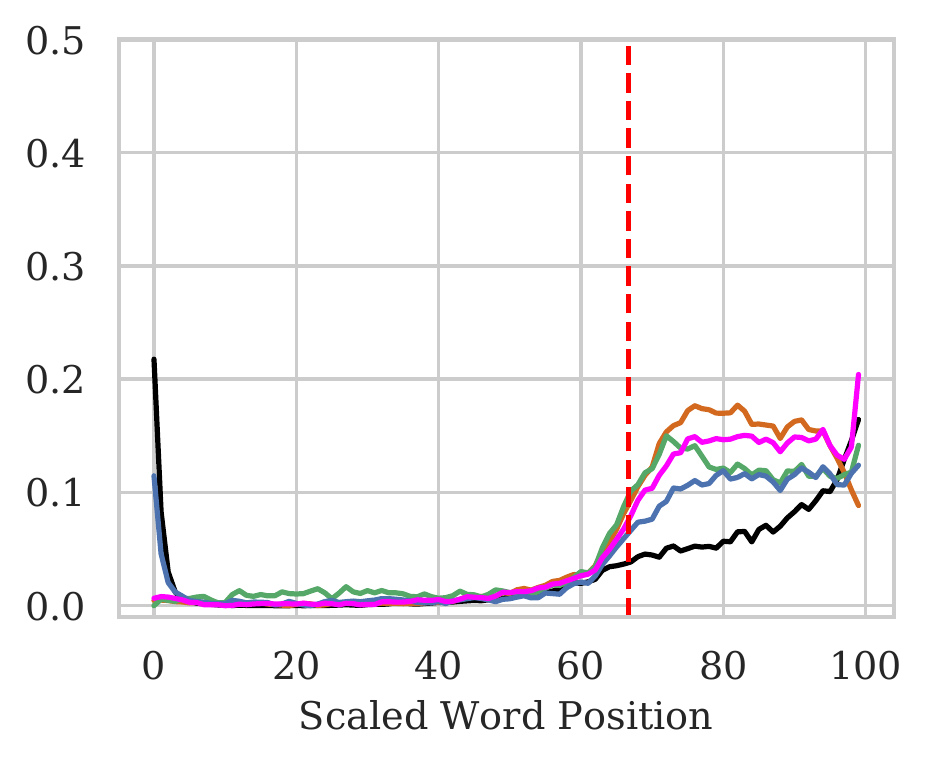}}
	\hfill
	\subcaptionbox{Right - Amazon \label{fig:app:Amazon:NWI_right}}{\includegraphics[width=0.32\textwidth]{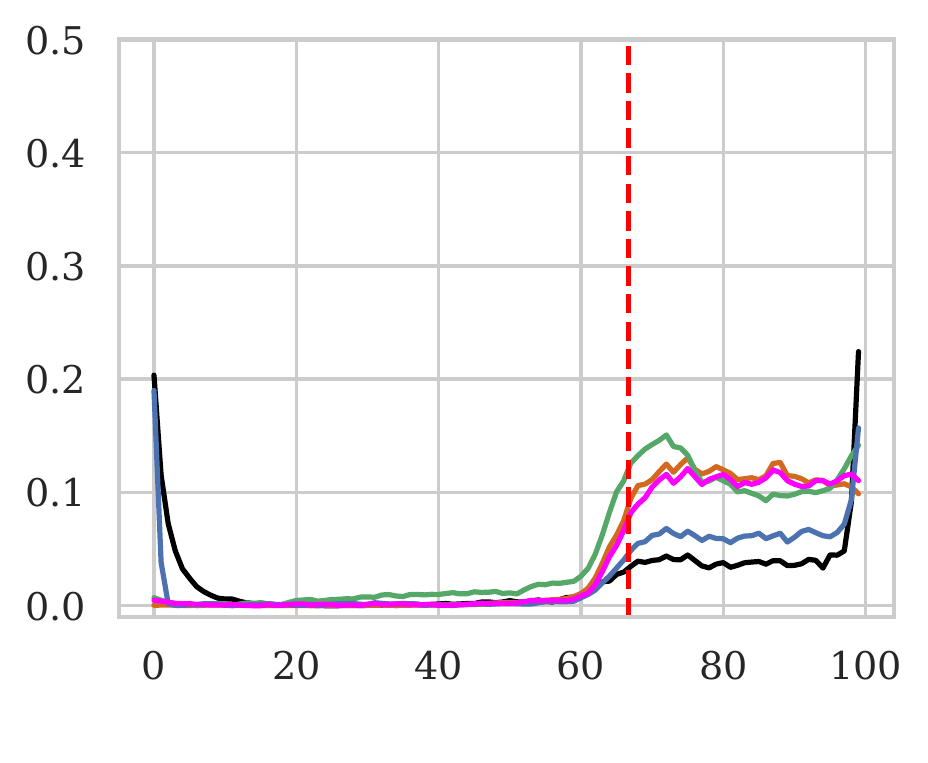}}
	\hfill
	\\
	
\caption{Normalized Word Importance w.r.t. word position for k = 5; averaged over sentences of length between 400-500 on the IMDb,  Yahoo, Amazon (10K) Datasets. Results shown for the `standard', `left', `mid' and `right' training settings described in \S~\ref{subsec:coarse-grained}. The vertical red line represents an approximate separator between relevant and irrelevant information (by construction). For instance, The word positions to the `left' of the vertical line in graphs in the second row of the Figure contain data from true input examples, while those to the right contain Wikipedia sentences.}
\label{fig:app:NWI:full}
\vspace{3pt}
\end{figure*}
\begin{figure*}[!ht]
\centering
    \includegraphics[width=0.6\textwidth]{Figures/Wiki_Attack/legend.pdf}\\
    \subcaptionbox{Standard \label{fig:Yahoo-short:NWI_none}}{\includegraphics[width=0.32\textwidth]{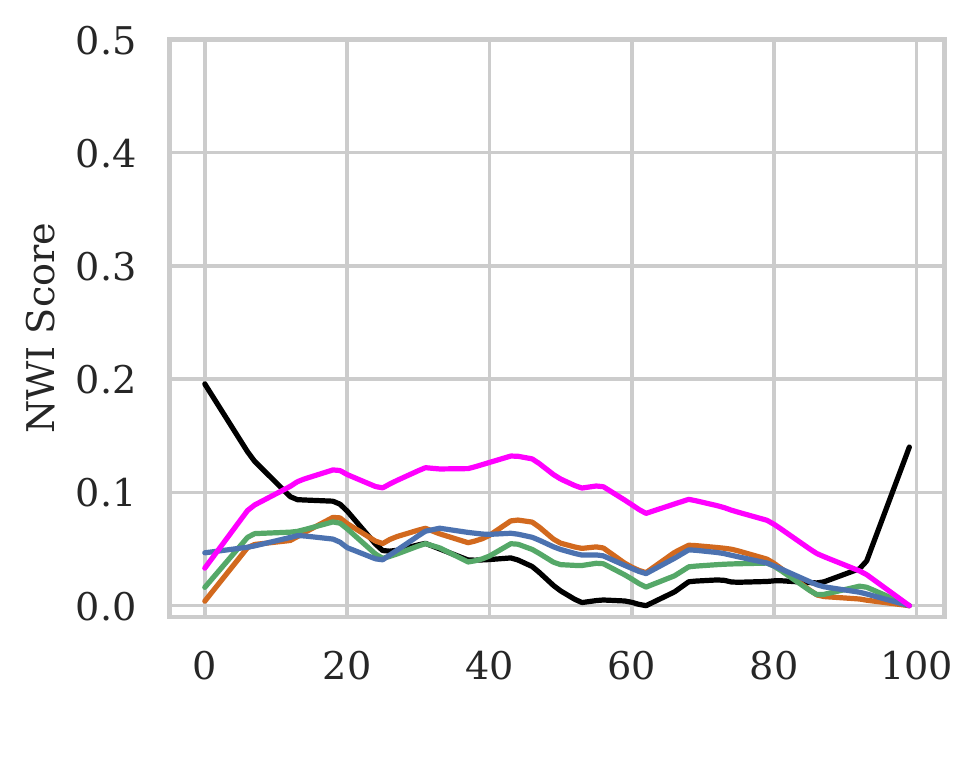}}
	\hfill
	\subcaptionbox{Left \label{fig:Yahoo-short:NWI_left}}{\includegraphics[width=0.32\textwidth]{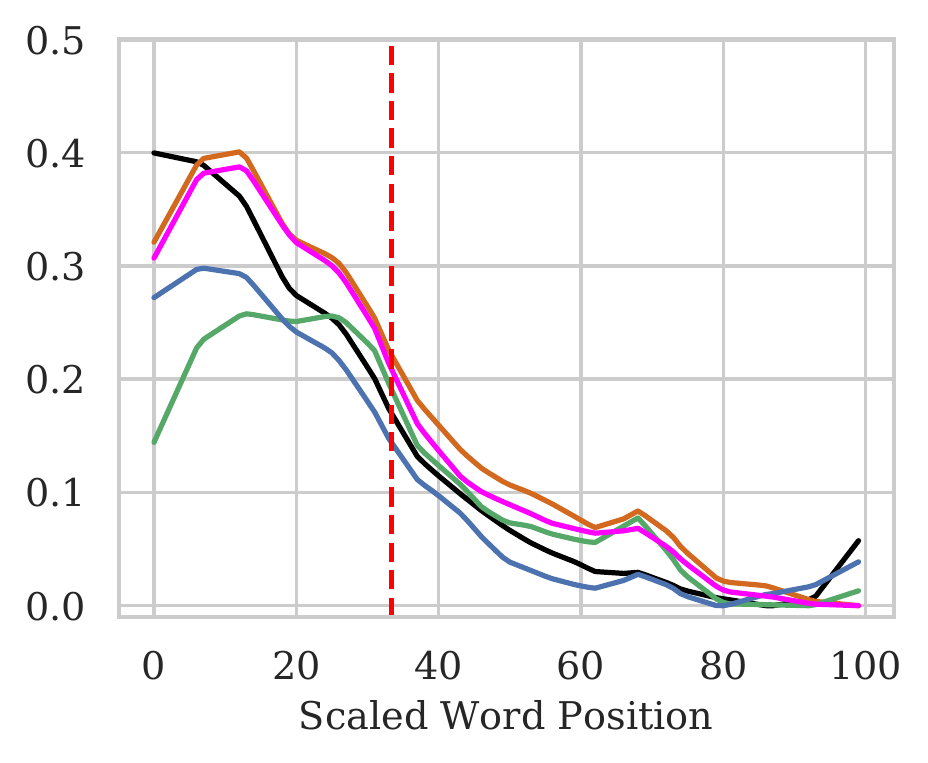}}
	\hfill
	\subcaptionbox{Mid \label{fig:Yahoo-short:NWI_mid}}{\includegraphics[width=0.32\textwidth]{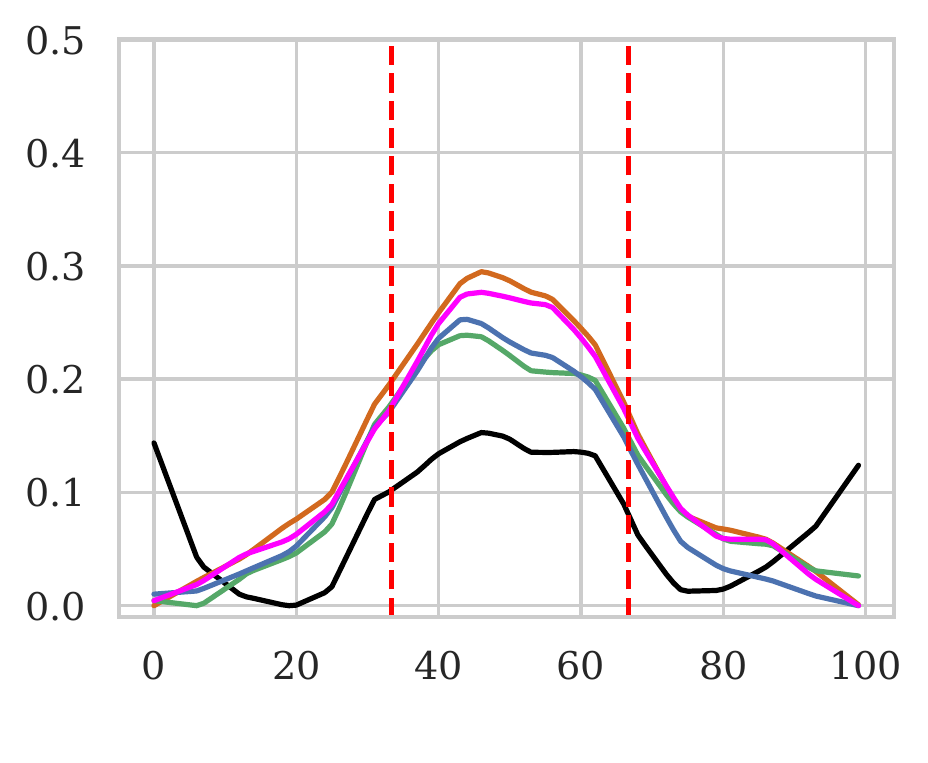}}
	\hfill
	\\
	\subcaptionbox{Standard \label{fig:Amazon-short:NWI_none}}{\includegraphics[width=0.32\textwidth]{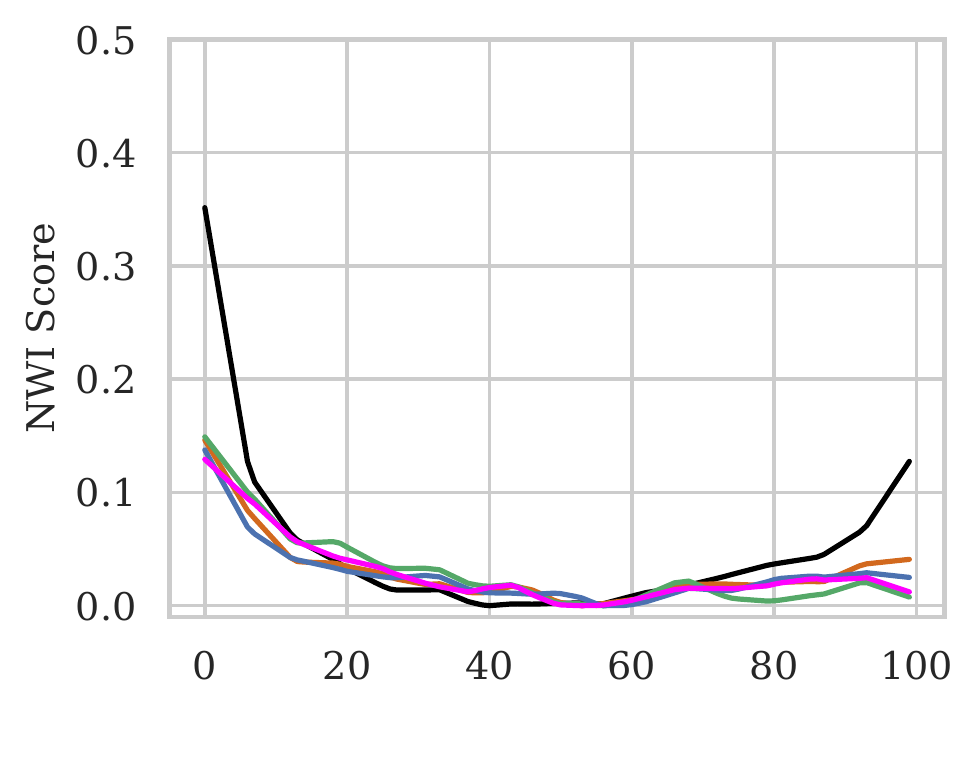}}
	\hfill
	\subcaptionbox{Left \label{fig:Amazon-short:NWI_left}}{\includegraphics[width=0.32\textwidth]{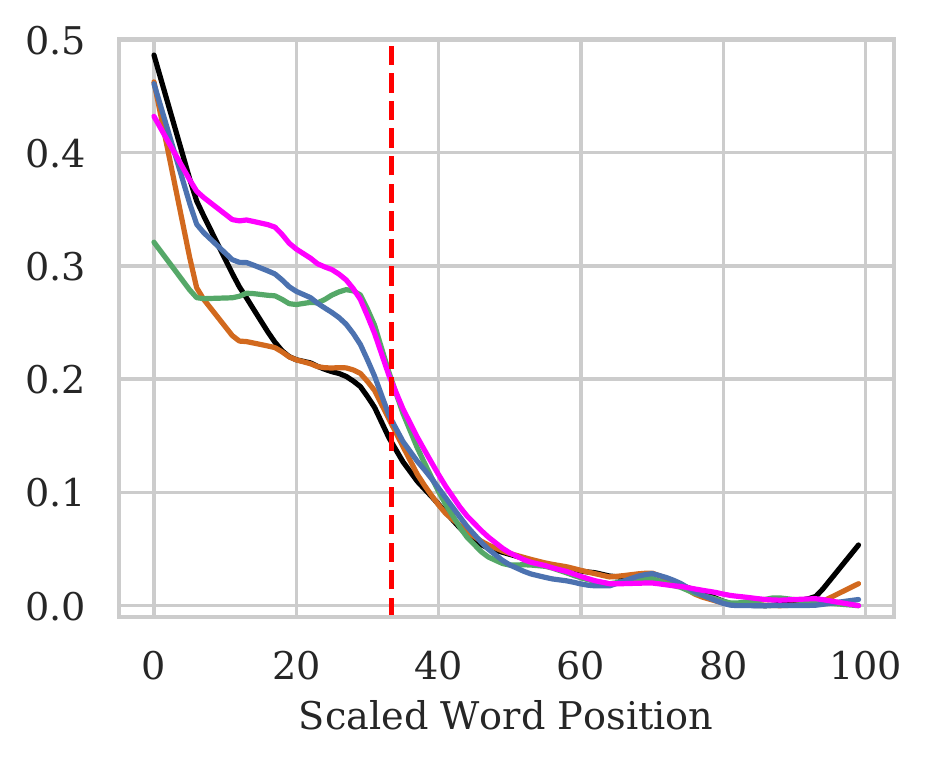}}
	\hfill
	\subcaptionbox{Mid \label{fig:Amazon-short:NWI_mid}}{\includegraphics[width=0.32\textwidth]{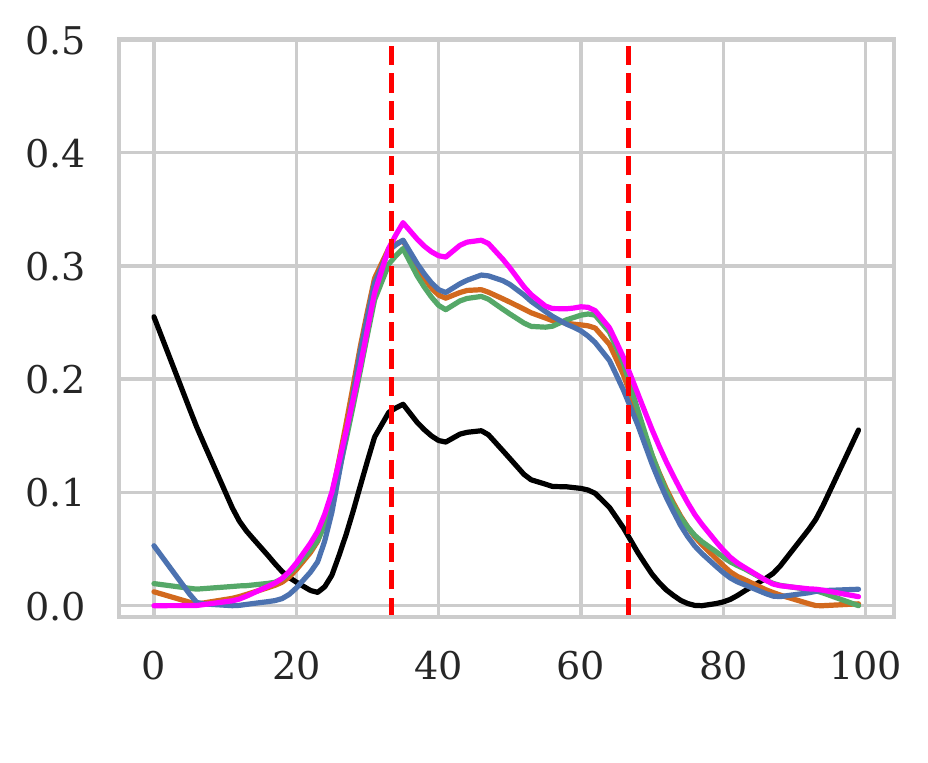}}
	\hfill
	
\caption{Normalized Word Importance w.r.t. word position for k = 3; averaged over sentences of length between 50-60 on the Yahoo, Amazon (10K) Datasets. Results shown for the `standard', `left' and `mid' training settings described in Appendix~\ref{app:wiki-short_sent}. The vertical red line represents an approximate separator between relevant and irrelevant information (by construction). For instance, The word positions to the `left' of the vertical line in (b), (e) contain data from true input examples, while those to the right contain Wikipedia sentences.}
\label{fig:NWI_short_full}
\end{figure*}
\section{Fine-grained Positional Biases}
\label{app:fine-grained-assessment}

We detail the method for calculating the Normalized Word Importance (NWI) score in Algorithm~\ref{alg:NWI}. 
\begin{algorithm}[H]
\vspace{4pt}
  \caption{\textit{NWI} evaluation}    
    \begin{algorithmic}
  \STATE {\bfseries Input:} softmax classifier $P_\theta$,  test set $D$
  \STATE{\bfseries Parameters:} $k$ 
  \FOR {$s_j = \{x_j^1,\dots,x_j^n\}, y_j$ in $D$}
      \STATE $p_j = \log \{P_{\theta}(y_j|s_j) \}$
      \FOR{$t=0\dots \frac{n}{k}$}
        \STATE $s_j^t = \{x_j^1, \dots, x_j^{k.t},
        \underbrace{
        \texttt{UNK}, \dots, \texttt{UNK}}_\text{k words}
        ,\dots,x_j^n\}$
        \STATE $p_j^t = \log \{P_{\theta}(y_j|s_j^t) \}$
        \STATE $\delta_j^{t} = |p_j^t - p_j|$
      \ENDFOR
        \STATE $\text{nwi}_j = \frac{\delta_j}{\max_{t \in (1,\frac{n}{k})}\delta_j^{t}}$ 
        \STATE $\text{nwi}_j =  \text{nwi}_j - \min_{t \in (1,\frac{n}{k})}\delta_j^{t}$
        \STATE 
        $\text{nwi}_j =
        \texttt{LinInterp}(\text{nwi}_j,\frac{n}{k},100)$
    \ENDFOR
  \STATE \textbf{return} $\frac{1}{|D|}\sum_{j = 1}^{|D|} \text{nwi}_j$
  \STATE *\texttt{LinInterp}$(x,n,l)$ linearly interpolates input distribution $x$ of $n$ discrete steps to $l$ steps.
\end{algorithmic}
\label{alg:NWI}
\vspace{4pt}
\end{algorithm}
The parameter $k$ can be adjusted according to the average sentence length. 
For a sentence of length 100, setting an extremely low value of $k$ (say 1) may have very little impact of the model's prediction $\log \{P_{\theta}(y_j|s_j^t) \}$ for all positions $t$. On the other hand, setting an extremely high value of $k$ (say 20) may provide only few data points, and also change the model prediction drastically at all values of $t$.

Complete graphs for the positional importance (as perceived by the model) of words are detailed in this section.
The trends observed for the remaining datasets are similar to the representative graphs shown in the main paper.  We show graphs for the IMDb, Yahoo and Amazon datasets in Figure~\ref{fig:app:NWI:full}.

\paragraph{Practical Implications} Our findings suggest that adversaries can easily replace the middle portion of texts with racist or abusive sentences, and still stay undetected by \lastf{} based detection systems. This is because \lastf{} attributes little or no importance to words in the middle of the input. Pooling based models are able to circumvent this issue by being able to attribute importance to words irrespective of their position. %

\subsection{NWI for Short sentences }
\label{app:para-NWI-short}
We repeat our experiments of NWI evaluation on the datasets with short sentences ($<$100 words) as described in Appendix~\ref{app:wiki-short_sent}. 
It is interesting to observe the graphs on the Yahoo and Amazon short datasets in Figure \ref{fig:NWI_short_full}, where due to the short sentence length, even \lastf{} is able to show the desired importance characteristic in case of mid setting. 
This supports the fact that the test time accuracies in the mid setting are no longer as bad as a majority class predictor. 
Interestingly, in case of short sentences in the mid setting (Figures~\ref{fig:Yahoo-short:NWI_mid},\ref{fig:Amazon-short:NWI_mid}), we observe three peaks in the NWI graph. The one in the middle is expected given the data distribution. However, the two peaks in NWI at the extreme ends help establish that while \lastf{} is able to propagate gradients to the middle given the short sentences, it is still not able to forego the extreme bias towards the end tokens.

%% file: tables/table_yahoo_amazon.tex
\begin{table*}[htb]
\centering
\small
\begin{tabular}{@{}lrrrrrr|rrrrrr@{}}

\toprule
\multicolumn{1}{c}{} & \multicolumn{12}{c}{Datasets with Long Sentences}\\ 

\toprule
\multicolumn{1}{c}{} & \multicolumn{6}{c}{Yahoo Dataset}  & \multicolumn{6}{c}{Amazon Dataset}\\ 
\midrule
\multicolumn{1}{c}{} & 
\multicolumn{1}{c}{1K} & \multicolumn{1}{c}{2K} & \multicolumn{1}{c}{5K} & \multicolumn{1}{c}{10K} & \multicolumn{1}{c}{25K} & 
\multicolumn{1}{c}{} & \multicolumn{1}{c}{} & 
\multicolumn{1}{c}{1K} & \multicolumn{1}{c}{2K} & \multicolumn{1}{c}{5K} & \multicolumn{1}{c}{10K} & \multicolumn{1}{c}{25K}    
\\
\cmidrule(r){2-7} \cmidrule(r){9-13} 

\lastf{}   & 38.3 \tiny$\pm$ 4.8 & 51.4 \tiny$\pm$ 2.1 & 57.4 \tiny$\pm$ 0.6 & 63.5 \tiny$\pm$ 0.6 & 67.5 \tiny$\pm$ 0.8 &  &
						  	& 38.5 \tiny$\pm$ 4.2 & 52.7 \tiny$\pm$ 7.7 & 70.0 \tiny$\pm$ 0.9  & 76.2 \tiny$\pm$ 0.7 & 81.8 \tiny$\pm$ 0.3 \\
\meanout{} & 48.2 \tiny$\pm$ 2.3 & 56.6 \tiny$\pm$ 0.5 & 60.8 \tiny$\pm$ 0.5 & 64.7 \tiny$\pm$ 0.6 & 68.7 \tiny$\pm$ 0.6 &  &
						  	& 44.8 \tiny$\pm$ 9.8 & 55.6 \tiny$\pm$ 6.4 & 71.2 \tiny$\pm$ 0.9  & 76.9 \tiny$\pm$ 0.4 & 82.0 \tiny$\pm$ 0.3 \\
\maxout{}  & 50.2 \tiny$\pm$ 2.1 & 56.3 \tiny$\pm$ 1.8 & 61.3 \tiny$\pm$ 0.9 & 63.9 \tiny$\pm$ 1.1 & 67.0 \tiny$\pm$ 1.1 &  &
						  	& 49.6 \tiny$\pm$ 3.9 & 61.6 \tiny$\pm$ 2.6 & 73.9 \tiny$\pm$ 0.2  & 79.1 \tiny$\pm$ 0.4 & 84.2 \tiny$\pm$ 0.2 \\
\att{}     & 47.3 \tiny$\pm$ 2.2 & 54.2 \tiny$\pm$ 1.1 & 61.0 \tiny$\pm$ 0.5 & 65.1 \tiny$\pm$ 1.5 & 68.2 \tiny$\pm$ 0.7 &  &
						  	& 54.1 \tiny$\pm$ 5.2 & 61.2 \tiny$\pm$ 2.9 & 72.0 \tiny$\pm$ 0.2  & 77.0 \tiny$\pm$ 0.3 & 82.6 \tiny$\pm$ 0.1 \\
\attmax{}  & 51.8 \tiny$\pm$ 1.1 & 57.0 \tiny$\pm$ 1.1 & 63.2 \tiny$\pm$ 0.4 & 65.1 \tiny$\pm$ 1.1 & 68.4 \tiny$\pm$ 0.6 &  &
						  	& 58.2 \tiny$\pm$ 3.8 & 65.6 \tiny$\pm$ 0.9 & 72.8 \tiny$\pm$ 0.5  & 77.3 \tiny$\pm$ 0.2 & 82.4 \tiny$\pm$ 0.2 \\

\toprule
\multicolumn{1}{c}{} & \multicolumn{6}{c}{Yahoo (left) + Wiki}  & \multicolumn{6}{c}{Amazon (left) + Wiki}\\ 
\midrule
\multicolumn{1}{c}{} & 
\multicolumn{1}{c}{1K} & \multicolumn{1}{c}{2K} & \multicolumn{1}{c}{5K} & \multicolumn{1}{c}{10K} & \multicolumn{1}{c}{25K} & 
\multicolumn{1}{c}{} & \multicolumn{1}{c}{} & 
\multicolumn{1}{c}{1K} & \multicolumn{1}{c}{2K} & \multicolumn{1}{c}{5K} & \multicolumn{1}{c}{10K} & \multicolumn{1}{c}{25K}    
\\
\cmidrule(r){2-7} \cmidrule(r){9-13} 

\lastf{}   	& 41.4 \tiny$\pm$ 2.9 & 51.0 \tiny$\pm$ 0.5 & 56.2 \tiny$\pm$ 1.2 & 60.9 \tiny$\pm$ 0.7 & 64.6 \tiny$\pm$ 2.0 &  &
		  	& 44.9 \tiny$\pm$ 0.7 & 57.0 \tiny$\pm$ 0.8 & 68.3 \tiny$\pm$ 0.9 & 73.5 \tiny$\pm$ 0.4 & 79.6 \tiny$\pm$ 0.2 \\ 
\meanout{} 	& 31.9 \tiny$\pm$ 1.5 & 43.3 \tiny$\pm$ 1.7 & 51.4 \tiny$\pm$ 0.9 & 58.8 \tiny$\pm$ 0.7 & 65.1 \tiny$\pm$ 0.3 &  &
		  	& 31.0 \tiny$\pm$ 2.1 & 48.1 \tiny$\pm$ 1.4 & 65.0 \tiny$\pm$ 1.4 & 70.8 \tiny$\pm$ 1.2 & 79.1 \tiny$\pm$ 0.8 \\
\maxout{}  	& 33.6 \tiny$\pm$ 0.9 & 42.3 \tiny$\pm$ 1.4 & 52.7 \tiny$\pm$ 2.0 & 60.7 \tiny$\pm$ 0.9 & 66.0 \tiny$\pm$ 1.0 &  &
		  	& 19.2 \tiny$\pm$ 1.9 & 42.5 \tiny$\pm$ 3.5 & 68.5 \tiny$\pm$ 2.6 & 76.8 \tiny$\pm$ 0.6 & 82.1 \tiny$\pm$ 0.5 \\
\att{}     	& 37.3 \tiny$\pm$ 0.5 & 47.2 \tiny$\pm$ 2.2 & 57.6 \tiny$\pm$ 1.6 & 62.5 \tiny$\pm$ 1.0 & 67.6 \tiny$\pm$ 0.3 &  &
		  	& 47.6 \tiny$\pm$ 2.0 & 59.3 \tiny$\pm$ 1.1 & 70.8 \tiny$\pm$ 0.9 & 75.6 \tiny$\pm$ 0.3 & 81.3 \tiny$\pm$ 0.3 \\
\attmax{}  	& 40.0 \tiny$\pm$ 0.6 & 48.7 \tiny$\pm$ 0.5 & 59.6 \tiny$\pm$ 1.4 & 63.0 \tiny$\pm$ 1.4 & 67.2 \tiny$\pm$ 0.9 &  &
		  	& 56.1 \tiny$\pm$ 1.3 & 63.8 \tiny$\pm$ 1.3 & 70.3 \tiny$\pm$ 0.3 & 75.6 \tiny$\pm$ 0.2 & 80.7 \tiny$\pm$ 0.5 \\ 

\toprule
\multicolumn{1}{c}{} & \multicolumn{6}{c}{Yahoo (mid) + Wiki}  & \multicolumn{6}{c}{Amazon (mid) + Wiki}\\ 
\midrule
\multicolumn{1}{c}{} & 
\multicolumn{1}{c}{1K} & \multicolumn{1}{c}{2K} & \multicolumn{1}{c}{5K} & \multicolumn{1}{c}{10K} & \multicolumn{1}{c}{25K} & 
\multicolumn{1}{c}{} & \multicolumn{1}{c}{} & 
\multicolumn{1}{c}{1K} & \multicolumn{1}{c}{2K} & \multicolumn{1}{c}{5K} & \multicolumn{1}{c}{10K} & \multicolumn{1}{c}{25K}    
\\
\cmidrule(r){2-7} \cmidrule(r){9-13} 

\lastf{}   & 12.7 \tiny$\pm$ 1.1 & 12.7 \tiny$\pm$ 1.1 & 12.0 \tiny$\pm$ 0.9 & 11.4 \tiny$\pm$ 0.8 & 13.2 \tiny$\pm$ 2.2 &  &
						  	& 5.3 \tiny$\pm$ 0.3  & 5.4 \tiny$\pm$ 0.3  & 5.0 \tiny$\pm$ 0.1   & 5.1 \tiny$\pm$ 0.4  & 7.8 \tiny$\pm$ 5.2  \\
\meanout{} & 31.9 \tiny$\pm$ 2.3 & 43.1 \tiny$\pm$ 2.0 & 50.1 \tiny$\pm$ 1.6 & 58.5 \tiny$\pm$ 0.6 & 64.9 \tiny$\pm$ 0.7 &  &
						  	& 34.4 \tiny$\pm$ 3.5 & 52.7 \tiny$\pm$ 3.5 & 63.4 \tiny$\pm$ 2.0  & 70.3 \tiny$\pm$ 1.7 & 79.0 \tiny$\pm$ 0.6 \\
\maxout{}  & 33.0 \tiny$\pm$ 1.0 & 40.1 \tiny$\pm$ 1.4 & 51.0 \tiny$\pm$ 1.2 & 58.4 \tiny$\pm$ 1.2 & 65.5 \tiny$\pm$ 0.7 &  &
						  	& 17.0 \tiny$\pm$ 0.7 & 34.5 \tiny$\pm$ 2.0 & 58.8 \tiny$\pm$ 0.4  & 72.8 \tiny$\pm$ 0.6 & 80.4 \tiny$\pm$ 0.3 \\
\att{}     & 39.4 \tiny$\pm$ 0.5 & 45.1 \tiny$\pm$ 1.8 & 57.0 \tiny$\pm$ 2.0 & 61.5 \tiny$\pm$ 1.7 & 66.5 \tiny$\pm$ 0.6 &  &
						  	& 48.0 \tiny$\pm$ 1.7 & 59.1 \tiny$\pm$ 1.8 & 69.5 \tiny$\pm$ 0.6  & 75.3 \tiny$\pm$ 0.5 & 81.1 \tiny$\pm$ 0.2 \\
\attmax{}  & 39.6 \tiny$\pm$ 0.9 & 48.5 \tiny$\pm$ 0.6 & 58.7 \tiny$\pm$ 1.5 & 62.2 \tiny$\pm$ 1.6 & 66.5 \tiny$\pm$ 0.7 &  &
						  	& 57.7 \tiny$\pm$ 0.5 & 63.0 \tiny$\pm$ 0.8 & 69.8 \tiny$\pm$ 0.6  & 74.8 \tiny$\pm$ 0.5 & 80.3 \tiny$\pm$ 0.4 \\

\toprule
\multicolumn{1}{c}{} & \multicolumn{6}{c}{Yahoo (right) + Wiki}  & \multicolumn{6}{c}{Amazon (right) + Wiki}\\ 
\midrule
\multicolumn{1}{c}{} & 
\multicolumn{1}{c}{1K} & \multicolumn{1}{c}{2K} & \multicolumn{1}{c}{5K} & \multicolumn{1}{c}{10K} & \multicolumn{1}{c}{25K} & 
\multicolumn{1}{c}{} & \multicolumn{1}{c}{} & 
\multicolumn{1}{c}{1K} & \multicolumn{1}{c}{2K} & \multicolumn{1}{c}{5K} & \multicolumn{1}{c}{10K} & \multicolumn{1}{c}{25K}    
\\
\cmidrule(r){2-7} \cmidrule(r){9-13}

\lastf{}   & 18.8 \tiny$\pm$ 2.5 & 37.3 \tiny$\pm$ 0.9 & 52.9 \tiny$\pm$ 2.1 & 60.1 \tiny$\pm$ 1.5 & 65.4 \tiny$\pm$ 0.6 &  &
						  	& 7.9 \tiny$\pm$ 0.6  & 27.9 \tiny$\pm$ 9.9 & 45.8 \tiny$\pm$ 16.2 & 70.8 \tiny$\pm$ 1.5 & 78.7 \tiny$\pm$ 0.8 \\
\meanout{} & 33.9 \tiny$\pm$ 2.1 & 43.2 \tiny$\pm$ 1.0 & 50.6 \tiny$\pm$ 0.8 & 58.6 \tiny$\pm$ 0.4 & 64.6 \tiny$\pm$ 0.5 &  &
						  	& 33.3 \tiny$\pm$ 1.0 & 48.2 \tiny$\pm$ 3.4 & 64.1 \tiny$\pm$ 0.7  & 71.9 \tiny$\pm$ 0.8 & 78.8 \tiny$\pm$ 0.2 \\
\maxout{}  & 33.1 \tiny$\pm$ 2.5 & 41.2 \tiny$\pm$ 0.9 & 53.0 \tiny$\pm$ 3.6 & 60.9 \tiny$\pm$ 1.0 & 66.0 \tiny$\pm$ 0.7 &  &
						  	& 17.0 \tiny$\pm$ 1.7 & 36.5 \tiny$\pm$ 3.0 & 64.3 \tiny$\pm$ 1.5  & 72.4 \tiny$\pm$ 0.3 & 80.2 \tiny$\pm$ 0.9 \\
\att{}     & 37.9 \tiny$\pm$ 1.4 & 47.6 \tiny$\pm$ 2.3 & 58.1 \tiny$\pm$ 1.4 & 62.2 \tiny$\pm$ 0.9 & 67.0 \tiny$\pm$ 0.3 &  &
						  	& 48.9 \tiny$\pm$ 1.5 & 58.9 \tiny$\pm$ 1.3 & 69.7 \tiny$\pm$ 0.6  & 75.7 \tiny$\pm$ 0.3 & 81.1 \tiny$\pm$ 0.3 \\
\attmax{}  & 40.3 \tiny$\pm$ 1.5 & 50.1 \tiny$\pm$ 1.6 & 59.3 \tiny$\pm$ 1.2 & 63.1 \tiny$\pm$ 0.7 & 66.8 \tiny$\pm$ 0.3 &  &
						  	& 57.8 \tiny$\pm$ 0.8 & 63.7 \tiny$\pm$ 0.8 & 71.1 \tiny$\pm$ 0.6  & 75.3 \tiny$\pm$ 0.3 & 80.7 \tiny$\pm$ 0.5 \\

\bottomrule
\end{tabular}
\caption{Mean test accuracy ($\pm$ std) (in \%) on different manipulated settings across 5 random seeds on the Yahoo, Amazon datasets with long sentences (greater than 100 words).}
\label{table:wiki-long-yahoo-amazon}
\end{table*}

%% file: tables/table_IMDB_Yelp.tex
\begin{table*}[htb]
\centering
\small
\begin{tabular}{@{}lllllll|llllll@{}}
\toprule
\multicolumn{1}{c}{} & \multicolumn{12}{c}{Datasets with Long Sentences}\\ 

\toprule
\multicolumn{1}{c}{} & \multicolumn{6}{c}{IMDb Dataset}  & \multicolumn{6}{c}{Yelp Dataset}\\ 
\midrule
\multicolumn{1}{c}{} & 
\multicolumn{1}{c}{1K} & \multicolumn{1}{c}{2K} & \multicolumn{1}{c}{5K} & \multicolumn{1}{c}{10K} & \multicolumn{1}{c}{20K} & 
\multicolumn{1}{c}{} & \multicolumn{1}{c}{} & 
\multicolumn{1}{c}{1K} & \multicolumn{1}{c}{2K} & \multicolumn{1}{c}{5K} & \multicolumn{1}{c}{10K} & \multicolumn{1}{c}{25K}    
\\
\cmidrule(r){2-7} \cmidrule(r){9-13} 

\lastf{}                    & 64.7 \tiny$\pm$ 2.3 & 75.0 \tiny$\pm$ 0.4 & 83.2 \tiny$\pm$ 0.4 & 86.6 \tiny$\pm$ 0.8 & 88.7 \tiny$\pm$ 0.6 &  &
                            & 80.7 \tiny$\pm$ 4.1 & 84.9 \tiny$\pm$ 8.0 & 92.2 \tiny$\pm$ 0.3 & 93.1 \tiny$\pm$ 0.1 & 94.1 \tiny$\pm$ 0.3 \\
\meanout{}                  & 73.0 \tiny$\pm$ 3.0 & 81.7 \tiny$\pm$ 0.7 & 85.4 \tiny$\pm$ 0.1 & 87.1 \tiny$\pm$ 0.6 & 88.6 \tiny$\pm$ 0.3 &  &
                            & 87.1 \tiny$\pm$ 1.2 & 87.9 \tiny$\pm$ 1.7 & 92.2 \tiny$\pm$ 0.4 & 93.4 \tiny$\pm$ 0.3 & 94.4 \tiny$\pm$ 0.1 \\
\maxout{}                   & 69.0 \tiny$\pm$ 3.9 & 80.1 \tiny$\pm$ 0.5 & 85.7 \tiny$\pm$ 0.2 & 87.8 \tiny$\pm$ 0.6 & 89.9 \tiny$\pm$ 0.3 &  &
                            & 84.4 \tiny$\pm$ 2.0 & 86.4 \tiny$\pm$ 5.1 & 92.2 \tiny$\pm$ 0.3 & 93.4 \tiny$\pm$ 0.2 & 94.7 \tiny$\pm$ 0.2 \\
\att{}                      & 75.7 \tiny$\pm$ 2.6 & 82.8 \tiny$\pm$ 0.8 & 86.9 \tiny$\pm$ 0.7 & 89.0 \tiny$\pm$ 0.3 & 90.3 \tiny$\pm$ 0.2 &  &
                            & 82.5 \tiny$\pm$ 3.7 & 85.6 \tiny$\pm$ 6.5 & 92.6 \tiny$\pm$ 0.4 & 93.7 \tiny$\pm$ 0.2 & 94.9 \tiny$\pm$ 0.1 \\
\attmax{}                   & 75.9 \tiny$\pm$ 2.2 & 82.5 \tiny$\pm$ 0.4 & 86.1 \tiny$\pm$ 0.8 & 88.5 \tiny$\pm$ 0.5 & 89.9 \tiny$\pm$ 0.2 &  &
                            & 81.3 \tiny$\pm$ 5.1 & 86.0 \tiny$\pm$ 6.3 & 92.6 \tiny$\pm$ 0.2 & 93.7 \tiny$\pm$ 0.3 & 94.8 \tiny$\pm$ 0.1 \\

\toprule
\multicolumn{1}{c}{} & \multicolumn{6}{c}{IMDb (left) + Wiki}  & \multicolumn{6}{c}{Yelp (left) + Wiki}\\ 
\midrule
\multicolumn{1}{c}{} & 
\multicolumn{1}{c}{1K} & \multicolumn{1}{c}{2K} & \multicolumn{1}{c}{5K} & \multicolumn{1}{c}{10K} & \multicolumn{1}{c}{20K} & 
\multicolumn{1}{c}{} & \multicolumn{1}{c}{} & 
\multicolumn{1}{c}{1K} & \multicolumn{1}{c}{2K} & \multicolumn{1}{c}{5K} & \multicolumn{1}{c}{10K} & \multicolumn{1}{c}{25K}     
\\
\cmidrule(r){2-7} \cmidrule(r){9-13} 

\lastf{}                    & 67.6 \tiny$\pm$ 1.1 & 74.7 \tiny$\pm$ 1.2 & 80.6 \tiny$\pm$ 0.3 & 84.5 \tiny$\pm$ 0.4 & 87.2 \tiny$\pm$ 0.4 &  &
                            & 81.7 \tiny$\pm$ 0.5 & 87.5 \tiny$\pm$ 0.5 & 90.7 \tiny$\pm$ 0.5 & 92.0 \tiny$\pm$ 0.3 & 93.8 \tiny$\pm$ 0.2 \\
\meanout{}                  & 69.7 \tiny$\pm$ 3.4 & 76.6 \tiny$\pm$ 0.9 & 81.7 \tiny$\pm$ 0.7 & 83.6 \tiny$\pm$ 1.0 & 86.5 \tiny$\pm$ 0.8 &  &
                            & 78.1 \tiny$\pm$ 1.3 & 87.0 \tiny$\pm$ 1.1 & 90.9 \tiny$\pm$ 0.3 & 92.5 \tiny$\pm$ 0.1 & 93.8 \tiny$\pm$ 0.2 \\
\maxout{}                   & 68.8 \tiny$\pm$ 1.2 & 76.8 \tiny$\pm$ 1.7 & 82.2 \tiny$\pm$ 0.8 & 86.9 \tiny$\pm$ 0.9 & 88.4 \tiny$\pm$ 0.5 &  &
                            & 80.2 \tiny$\pm$ 1.5 & 87.5 \tiny$\pm$ 1.0 & 91.4 \tiny$\pm$ 0.2 & 93.0 \tiny$\pm$ 0.4 & 94.3 \tiny$\pm$ 0.1 \\
\att{}                      & 76.5 \tiny$\pm$ 1.5 & 79.6 \tiny$\pm$ 1.1 & 82.6 \tiny$\pm$ 0.6 & 86.9 \tiny$\pm$ 0.8 & 88.9 \tiny$\pm$ 0.5 &  &
                            & 84.7 \tiny$\pm$ 1.6 & 89.5 \tiny$\pm$ 0.7 & 92.0 \tiny$\pm$ 0.2 & 92.9 \tiny$\pm$ 0.4 & 94.4 \tiny$\pm$ 0.2 \\
\attmax{}                   & 75.8 \tiny$\pm$ 1.5 & 80.6 \tiny$\pm$ 1.0 & 84.1 \tiny$\pm$ 1.5 & 87.1 \tiny$\pm$ 0.6 & 89.1 \tiny$\pm$ 0.2 &  &
                            & 84.7 \tiny$\pm$ 1.3 & 89.7 \tiny$\pm$ 0.6 & 92.1 \tiny$\pm$ 0.1 & 93.1 \tiny$\pm$ 0.4 & 94.2 \tiny$\pm$ 0.4 \\

\toprule
\multicolumn{1}{c}{} & \multicolumn{6}{c}{IMDb (mid) + Wiki}  & \multicolumn{6}{c}{Yelp (mid) + Wiki}\\ 
\midrule
\multicolumn{1}{c}{} & 
\multicolumn{1}{c}{1K} & \multicolumn{1}{c}{2K} & \multicolumn{1}{c}{5K} & \multicolumn{1}{c}{10K} & \multicolumn{1}{c}{20K} & 
\multicolumn{1}{c}{} & \multicolumn{1}{c}{} & 
\multicolumn{1}{c}{1K} & \multicolumn{1}{c}{2K} & \multicolumn{1}{c}{5K} & \multicolumn{1}{c}{10K} & \multicolumn{1}{c}{25K}    
\\
\cmidrule(r){2-7} \cmidrule(r){9-13} 

\lastf{}                    & 49.6 \tiny$\pm$ 0.7 & 49.9 \tiny$\pm$ 0.5 & 50.2 \tiny$\pm$ 0.3 & 50.3 \tiny$\pm$ 0.3 & 50.1 \tiny$\pm$ 0.3 &  &
                            & 50.2 \tiny$\pm$ 0.4 & 51.1 \tiny$\pm$ 0.9 & 51.2 \tiny$\pm$ 0.8 & 51.4 \tiny$\pm$ 0.7 & 51.5 \tiny$\pm$ 0.5 \\
\meanout{}                  & 69.8 \tiny$\pm$ 2.1 & 76.2 \tiny$\pm$ 1.0 & 82.2 \tiny$\pm$ 0.7 & 84.1 \tiny$\pm$ 0.7 & 86.5 \tiny$\pm$ 0.8 &  &
                            & 79.2 \tiny$\pm$ 1.1 & 86.7 \tiny$\pm$ 1.0 & 90.7 \tiny$\pm$ 0.3 & 92.7 \tiny$\pm$ 0.2 & 94.0 \tiny$\pm$ 0.1 \\
\maxout{}                   & 64.5 \tiny$\pm$ 1.8 & 77.2 \tiny$\pm$ 2.0 & 82.9 \tiny$\pm$ 1.2 & 86.0 \tiny$\pm$ 0.8 & 88.4 \tiny$\pm$ 0.6 &  &
                            & 81.1 \tiny$\pm$ 1.5 & 85.6 \tiny$\pm$ 0.6 & 90.7 \tiny$\pm$ 0.4 & 92.5 \tiny$\pm$ 0.4 & 94.1 \tiny$\pm$ 0.2 \\
\att{}                      & 75.0 \tiny$\pm$ 0.8 & 79.4 \tiny$\pm$ 0.8 & 83.4 \tiny$\pm$ 1.0 & 86.7 \tiny$\pm$ 1.4 & 88.8 \tiny$\pm$ 0.2 &  &
                            & 84.4 \tiny$\pm$ 1.0 & 89.3 \tiny$\pm$ 1.0 & 91.8 \tiny$\pm$ 0.6 & 92.5 \tiny$\pm$ 0.6 & 94.4 \tiny$\pm$ 0.2 \\
\attmax{}                   & 75.4 \tiny$\pm$ 2.4 & 80.9 \tiny$\pm$ 1.8 & 84.7 \tiny$\pm$ 1.3 & 86.8 \tiny$\pm$ 0.5 & 88.7 \tiny$\pm$ 0.4 &  &
                            & 85.1 \tiny$\pm$ 0.8 & 89.4 \tiny$\pm$ 0.5 & 91.7 \tiny$\pm$ 0.7 & 92.9 \tiny$\pm$ 0.3 & 94.3 \tiny$\pm$ 0.2 \\

\toprule
\multicolumn{1}{c}{} & \multicolumn{6}{c}{IMDb (right) + Wiki}  & \multicolumn{6}{c}{Yelp (right) + Wiki}\\ 
\midrule
\multicolumn{1}{c}{} & 
\multicolumn{1}{c}{1K} & \multicolumn{1}{c}{2K} & \multicolumn{1}{c}{5K} & \multicolumn{1}{c}{10K} & \multicolumn{1}{c}{20K} & 
\multicolumn{1}{c}{} & \multicolumn{1}{c}{} & 
\multicolumn{1}{c}{1K} & \multicolumn{1}{c}{2K} & \multicolumn{1}{c}{5K} & \multicolumn{1}{c}{10K} & \multicolumn{1}{c}{25K}    
\\
\cmidrule(r){2-7} \cmidrule(r){9-13}

\lastf{}                    & 53.5 \tiny$\pm$ 2.5 & 64.7 \tiny$\pm$ 2.8 & 79.7 \tiny$\pm$ 4.3 & 85.9 \tiny$\pm$ 0.5 & 88.5 \tiny$\pm$ 0.2 &  &
                            & 59.4 \tiny$\pm$ 3.7 & 79.6 \tiny$\pm$ 6.2 & 91.7 \tiny$\pm$ 0.3 & 92.7 \tiny$\pm$ 0.4 & 93.7 \tiny$\pm$ 0.4 \\
\meanout{}                  & 70.0 \tiny$\pm$ 1.1 & 76.8 \tiny$\pm$ 1.0 & 81.8 \tiny$\pm$ 0.5 & 84.8 \tiny$\pm$ 0.9 & 87.1 \tiny$\pm$ 0.3 &  &
                            & 79.4 \tiny$\pm$ 0.9 & 87.1 \tiny$\pm$ 0.6 & 90.9 \tiny$\pm$ 0.7 & 92.3 \tiny$\pm$ 0.4 & 93.8 \tiny$\pm$ 0.3 \\
\maxout{}                   & 65.9 \tiny$\pm$ 4.6 & 77.8 \tiny$\pm$ 0.9 & 84.9 \tiny$\pm$ 0.8 & 87.2 \tiny$\pm$ 0.6 & 89.3 \tiny$\pm$ 0.3 &  &
                            & 80.6 \tiny$\pm$ 0.8 & 86.7 \tiny$\pm$ 0.9 & 91.9 \tiny$\pm$ 0.5 & 93.2 \tiny$\pm$ 0.2 & 94.5 \tiny$\pm$ 0.3 \\
\att{}                      & 74.7 \tiny$\pm$ 1.4 & 80.2 \tiny$\pm$ 1.8 & 84.7 \tiny$\pm$ 1.1 & 87.1 \tiny$\pm$ 1.0 & 89.4 \tiny$\pm$ 0.3 &  &
                            & 84.8 \tiny$\pm$ 0.7 & 89.1 \tiny$\pm$ 0.9 & 92.0 \tiny$\pm$ 0.4 & 92.8 \tiny$\pm$ 0.4 & 94.3 \tiny$\pm$ 0.1 \\
\attmax{}                   & 77.9 \tiny$\pm$ 0.9 & 81.9 \tiny$\pm$ 0.5 & 85.2 \tiny$\pm$ 0.8 & 87.2 \tiny$\pm$ 0.5 & 89.4 \tiny$\pm$ 0.3 &  &
                            & 84.1 \tiny$\pm$ 2.5 & 89.5 \tiny$\pm$ 0.7 & 91.7 \tiny$\pm$ 0.9 & 93.0 \tiny$\pm$ 0.4 & 94.3 \tiny$\pm$ 0.1 \\

\bottomrule
\end{tabular}
\caption{Mean test accuracy ($\pm$ std) (in \%) on different manipulated settings across 5 random seeds on the IMDb, Yelp Reviews datasets with long sentences (less than 100 words).}
\label{table:wiki-long-IMDb-Yelp}
\end{table*}

%% file: tables/table_short.tex
\begin{table*}[htb]
\centering
\small
\begin{tabular}{@{}lrrrrrr|rrrrrr@{}}

\toprule
\multicolumn{1}{c}{} & \multicolumn{12}{c}{Datasets with Short Sentences}\\ 

\toprule
\multicolumn{1}{c}{} & \multicolumn{6}{c}{Yahoo Dataset}  & \multicolumn{6}{c}{Amazon Dataset}\\ 
\midrule
\multicolumn{1}{c}{} & 
\multicolumn{1}{c}{1K} & \multicolumn{1}{c}{2K} & \multicolumn{1}{c}{5K} & \multicolumn{1}{c}{10K} & \multicolumn{1}{c}{25K} & 
\multicolumn{1}{c}{} & \multicolumn{1}{c}{} & 
\multicolumn{1}{c}{1K} & \multicolumn{1}{c}{2K} & \multicolumn{1}{c}{5K} & \multicolumn{1}{c}{10K} & \multicolumn{1}{c}{25K}    
\\
\cmidrule(r){2-7} \cmidrule(r){9-13} 

\lastf{}   			& 20.5 \tiny$\pm$ 2.9 & 25.8 \tiny$\pm$ 3.7 & 33.1 \tiny$\pm$ 2.4 & 42.4 \tiny$\pm$ 0.2 & 46.0 \tiny$\pm$ 0.4 &  &
	  				& 26.6 \tiny$\pm$ 4.4 & 37.7 \tiny$\pm$ 3.4 & 48.6 \tiny$\pm$ 2.2 & 54.0 \tiny$\pm$ 2.6 & 61.7 \tiny$\pm$ 0.2 \\
\meanout{} 			& 23.1 \tiny$\pm$ 1.8 & 28.4 \tiny$\pm$ 1.5 & 35.3 \tiny$\pm$ 1.8 & 43.0 \tiny$\pm$ 0.3 & 46.5 \tiny$\pm$ 0.4 &  &
	  				& 29.4 \tiny$\pm$ 4.0 & 38.2 \tiny$\pm$ 3.4 & 49.0 \tiny$\pm$ 1.8 & 54.4 \tiny$\pm$ 2.6 & 62.0 \tiny$\pm$ 0.2 \\
\maxout{}  			& 23.0 \tiny$\pm$ 2.8 & 31.2 \tiny$\pm$ 1.4 & 37.3 \tiny$\pm$ 1.9 & 43.3 \tiny$\pm$ 0.4 & 46.8 \tiny$\pm$ 0.8 &  &
	  				& 33.5 \tiny$\pm$ 4.5 & 41.4 \tiny$\pm$ 3.3 & 50.8 \tiny$\pm$ 1.7 & 55.9 \tiny$\pm$ 2.0 & 62.8 \tiny$\pm$ 0.1 \\
\att{}     			& 24.3 \tiny$\pm$ 1.1 & 30.7 \tiny$\pm$ 2.5 & 36.3 \tiny$\pm$ 2.0 & 43.1 \tiny$\pm$ 0.2 & 46.4 \tiny$\pm$ 0.6 &  &
	  				& 36.4 \tiny$\pm$ 3.7 & 43.3 \tiny$\pm$ 1.7 & 50.9 \tiny$\pm$ 0.6 & 55.6 \tiny$\pm$ 0.6 & 61.9 \tiny$\pm$ 0.2 \\
\attmax{}  			& 25.1 \tiny$\pm$ 2.2 & 30.8 \tiny$\pm$ 2.6 & 37.9 \tiny$\pm$ 1.1 & 43.3 \tiny$\pm$ 0.3 & 46.8 \tiny$\pm$ 0.7 &  &
	  				& 37.4 \tiny$\pm$ 3.8 & 44.6 \tiny$\pm$ 1.2 & 51.6 \tiny$\pm$ 0.8 & 56.2 \tiny$\pm$ 0.8 & 62.4 \tiny$\pm$ 0.4 \\

\toprule
\multicolumn{1}{c}{} & \multicolumn{6}{c}{Yahoo (left) + Wiki}  & \multicolumn{6}{c}{Amazon (left) + Wiki}\\ 
\midrule
\multicolumn{1}{c}{} & 
\multicolumn{1}{c}{1K} & \multicolumn{1}{c}{2K} & \multicolumn{1}{c}{5K} & \multicolumn{1}{c}{10K} & \multicolumn{1}{c}{25K} & 
\multicolumn{1}{c}{} & \multicolumn{1}{c}{} & 
\multicolumn{1}{c}{1K} & \multicolumn{1}{c}{2K} & \multicolumn{1}{c}{5K} & \multicolumn{1}{c}{10K} & \multicolumn{1}{c}{25K}    
\\
\cmidrule(r){2-7} \cmidrule(r){9-13} 

\lastf{}   			& 19.6 \tiny$\pm$ 1.7 & 28.5 \tiny$\pm$ 0.8 & 34.5 \tiny$\pm$ 0.3 & 38.2 \tiny$\pm$ 0.7 & 43.0 \tiny$\pm$ 0.4 &  &
	  				& 20.0 \tiny$\pm$ 1.8 & 30.7 \tiny$\pm$ 1.9 & 43.4 \tiny$\pm$ 0.4 & 49.9 \tiny$\pm$ 0.4 & 56.1 \tiny$\pm$ 0.3 \\
\meanout{} 			& 17.0 \tiny$\pm$ 2.9 & 20.3 \tiny$\pm$ 0.3 & 29.1 \tiny$\pm$ 1.1 & 34.8 \tiny$\pm$ 1.1 & 42.0 \tiny$\pm$ 0.3 &  &
	  				& 10.4 \tiny$\pm$ 2.1 & 18.0 \tiny$\pm$ 2.4 & 34.3 \tiny$\pm$ 2.4 & 46.2 \tiny$\pm$ 1.1 & 55.0 \tiny$\pm$ 0.5 \\
\maxout{}  			& 15.7 \tiny$\pm$ 0.8 & 24.0 \tiny$\pm$ 1.2 & 33.5 \tiny$\pm$ 0.4 & 37.5 \tiny$\pm$ 1.0 & 43.7 \tiny$\pm$ 0.1 &  &
	  				& 12.4 \tiny$\pm$ 1.9 & 26.0 \tiny$\pm$ 0.6 & 44.5 \tiny$\pm$ 1.0 & 51.4 \tiny$\pm$ 0.3 & 57.5 \tiny$\pm$ 0.2 \\
\att{}     			& 19.8 \tiny$\pm$ 3.1 & 26.0 \tiny$\pm$ 0.5 & 35.5 \tiny$\pm$ 0.9 & 40.1 \tiny$\pm$ 0.5 & 43.8 \tiny$\pm$ 0.2 &  &
	  				& 21.3 \tiny$\pm$ 4.3 & 37.1 \tiny$\pm$ 0.7 & 46.1 \tiny$\pm$ 0.6 & 51.3 \tiny$\pm$ 0.8 & 57.2 \tiny$\pm$ 0.2 \\
\attmax{}  			& 19.7 \tiny$\pm$ 3.5 & 27.0 \tiny$\pm$ 0.9 & 36.2 \tiny$\pm$ 1.3 & 40.0 \tiny$\pm$ 0.6 & 43.7 \tiny$\pm$ 0.3 &  &
	  				& 22.1 \tiny$\pm$ 5.9 & 36.7 \tiny$\pm$ 1.3 & 46.7 \tiny$\pm$ 0.1 & 52.2 \tiny$\pm$ 0.1 & 57.5 \tiny$\pm$ 0.2 \\

\toprule
\multicolumn{1}{c}{} & \multicolumn{6}{c}{Yahoo (mid) + Wiki}  & \multicolumn{6}{c}{Amazon (mid) + Wiki}\\ 
\midrule
\multicolumn{1}{c}{} & 
\multicolumn{1}{c}{1K} & \multicolumn{1}{c}{2K} & \multicolumn{1}{c}{5K} & \multicolumn{1}{c}{10K} & \multicolumn{1}{c}{25K} & 
\multicolumn{1}{c}{} & \multicolumn{1}{c}{} & 
\multicolumn{1}{c}{1K} & \multicolumn{1}{c}{2K} & \multicolumn{1}{c}{5K} & \multicolumn{1}{c}{10K} & \multicolumn{1}{c}{25K}    
\\
\cmidrule(r){2-7} \cmidrule(r){9-13} 

\lastf{}   			& 9.9 \tiny$\pm$ 0.7  & 12.3 \tiny$\pm$ 0.8 & 17.4 \tiny$\pm$ 1.1 & 24.2 \tiny$\pm$ 0.9 & 36.3 \tiny$\pm$ 0.5 &  &
	  				& 5.6 \tiny$\pm$ 0.4  & 6.9 \tiny$\pm$ 0.5  & 20.3 \tiny$\pm$ 1.0 & 37.9 \tiny$\pm$ 0.9 & 51.5 \tiny$\pm$ 1.0 \\
\meanout{} 			& 14.9 \tiny$\pm$ 2.2 & 22.1 \tiny$\pm$ 1.3 & 28.3 \tiny$\pm$ 0.4 & 32.8 \tiny$\pm$ 0.8 & 39.2 \tiny$\pm$ 0.4 &  &
	  				& 10.8 \tiny$\pm$ 1.9 & 20.8 \tiny$\pm$ 1.3 & 39.0 \tiny$\pm$ 0.6 & 46.5 \tiny$\pm$ 0.5 & 54.8 \tiny$\pm$ 0.1 \\
\maxout{}  			& 14.1 \tiny$\pm$ 2.6 & 22.6 \tiny$\pm$ 0.3 & 28.6 \tiny$\pm$ 0.5 & 33.8 \tiny$\pm$ 1.2 & 40.1 \tiny$\pm$ 0.5 &  &
	  				& 10.6 \tiny$\pm$ 1.8 & 21.3 \tiny$\pm$ 1.7 & 37.1 \tiny$\pm$ 1.4 & 47.0 \tiny$\pm$ 0.9 & 55.3 \tiny$\pm$ 0.4 \\
\att{}     			& 16.9 \tiny$\pm$ 3.0 & 24.8 \tiny$\pm$ 1.1 & 31.4 \tiny$\pm$ 0.9 & 37.6 \tiny$\pm$ 0.5 & 42.1 \tiny$\pm$ 0.4 &  &
	  				& 17.4 \tiny$\pm$ 3.2 & 33.2 \tiny$\pm$ 1.0 & 43.9 \tiny$\pm$ 0.5 & 49.7 \tiny$\pm$ 0.3 & 55.4 \tiny$\pm$ 0.1 \\
\attmax{}  			& 18.2 \tiny$\pm$ 2.4 & 25.7 \tiny$\pm$ 0.5 & 32.6 \tiny$\pm$ 0.6 & 37.8 \tiny$\pm$ 0.8 & 42.1 \tiny$\pm$ 0.4 &  &
	  				& 17.8 \tiny$\pm$ 4.6 & 35.0 \tiny$\pm$ 1.2 & 44.7 \tiny$\pm$ 0.3 & 49.7 \tiny$\pm$ 0.5 & 55.8 \tiny$\pm$ 0.4 \\

\toprule
\multicolumn{1}{c}{} & \multicolumn{6}{c}{Yahoo (right) + Wiki}  & \multicolumn{6}{c}{Amazon (right) + Wiki}\\ 
\midrule
\multicolumn{1}{c}{} & 
\multicolumn{1}{c}{1K} & \multicolumn{1}{c}{2K} & \multicolumn{1}{c}{5K} & \multicolumn{1}{c}{10K} & \multicolumn{1}{c}{25K} & 
\multicolumn{1}{c}{} & \multicolumn{1}{c}{} & 
\multicolumn{1}{c}{1K} & \multicolumn{1}{c}{2K} & \multicolumn{1}{c}{5K} & \multicolumn{1}{c}{10K} & \multicolumn{1}{c}{25K}    
\\
\cmidrule(r){2-7} \cmidrule(r){9-13}

\lastf{}   			& 12.3 \tiny$\pm$ 0.5 & 23.8 \tiny$\pm$ 1.2 & 33.4 \tiny$\pm$ 0.6 & 38.2 \tiny$\pm$ 0.2 & 43.8 \tiny$\pm$ 0.3 &  &
	  				& 7.4 \tiny$\pm$ 0.8  & 15.3 \tiny$\pm$ 3.2 & 40.8 \tiny$\pm$ 0.5 & 50.4 \tiny$\pm$ 0.7 & 58.4 \tiny$\pm$ 0.4 \\
\meanout{} 			& 15.7 \tiny$\pm$ 1.9 & 22.7 \tiny$\pm$ 0.4 & 27.7 \tiny$\pm$ 0.9 & 34.2 \tiny$\pm$ 0.6 & 41.3 \tiny$\pm$ 0.1 &  &
	  				& 14.8 \tiny$\pm$ 2.0 & 20.4 \tiny$\pm$ 3.3 & 40.1 \tiny$\pm$ 1.2 & 48.6 \tiny$\pm$ 0.5 & 56.9 \tiny$\pm$ 0.3 \\
\maxout{}  			& 14.7 \tiny$\pm$ 0.6 & 22.5 \tiny$\pm$ 1.5 & 33.6 \tiny$\pm$ 0.5 & 38.5 \tiny$\pm$ 0.4 & 43.4 \tiny$\pm$ 0.5 &  &
	  				& 11.1 \tiny$\pm$ 2.3 & 24.0 \tiny$\pm$ 1.9 & 45.6 \tiny$\pm$ 0.5 & 52.0 \tiny$\pm$ 0.4 & 58.4 \tiny$\pm$ 0.3 \\
\att{}     			& 19.7 \tiny$\pm$ 0.2 & 27.4 \tiny$\pm$ 1.5 & 35.9 \tiny$\pm$ 0.2 & 40.0 \tiny$\pm$ 0.4 & 43.8 \tiny$\pm$ 0.7 &  &
	  				& 22.4 \tiny$\pm$ 5.7 & 36.6 \tiny$\pm$ 1.3 & 46.7 \tiny$\pm$ 0.4 & 52.5 \tiny$\pm$ 0.4 & 59.1 \tiny$\pm$ 0.3 \\
\attmax{}  			& 20.3 \tiny$\pm$ 1.3 & 28.1 \tiny$\pm$ 0.9 & 35.4 \tiny$\pm$ 0.8 & 40.3 \tiny$\pm$ 0.4 & 43.8 \tiny$\pm$ 0.4 &  &
	  				& 20.8 \tiny$\pm$ 6.8 & 37.3 \tiny$\pm$ 0.9 & 47.8 \tiny$\pm$ 0.4 & 53.1 \tiny$\pm$ 0.3 & 59.0 \tiny$\pm$ 0.2\\
\bottomrule
\end{tabular}
  \caption{Mean test accuracy ($\pm$ std) (in \%) on different manipulated settings across 5 random seeds on the Yahoo, Amazon datasets with short sentences (less than 100 words).}
  \label{table:wiki-short-summary}
\end{table*}